\documentclass[Afour,sageh,times]{sagej}

\def\volumeyear{2025}
\graphicspath{{./images/}}

\usepackage{amsmath,amsfonts,bm}

\def\eqref#1{equation~\ref{#1}}

\def\1{\bm{1}}

\def\vc{{\bm{c}}}

\def\vf{{\bm{f}}}

\def\vh{{\bm{h}}}

\def\vk{{\bm{k}}}
\def\vl{{\bm{l}}}

\def\vp{{\bm{p}}}
\def\vq{{\bm{q}}}
\def\vr{{\bm{r}}}

\def\vx{{\bm{x}}}
\def\vy{{\bm{y}}}

\def\vsA{{\mathcal{A}}}

\def\vsF{{\mathcal{F}}}

\def\vsH{{\mathcal{H}}}

\def\vsS{{\mathcal{S}}}

\def\vsX{{\mathcal{X}}}
\def\vsY{{\mathcal{Y}}}
\def\vsZ{{\mathcal{Z}}}

\def\mA{{\bm{A}}}
\def\mB{{\bm{B}}}

\def\mI{{\bm{I}}}

\def\mK{{\bm{K}}}

\def\mO{{\bm{O}}}

\def\mR{{\bm{R}}}

\def\mT{{\bm{T}}}

\def\mW{{\bm{W}}}
\def\mX{{\bm{X}}}

\DeclareMathAlphabet{\mathsfit}{\encodingdefault}{\sfdefault}{m}{sl}
\SetMathAlphabet{\mathsfit}{bold}{\encodingdefault}{\sfdefault}{bx}{n}

\def\sD{{\mathbb{D}}}

\def\sI{{\mathbb{I}}}

\def\sS{{\mathbb{S}}}

\newcommand{\R}{\mathbb{R}}

\usepackage{mathtools}
\usepackage{xparse}
\usepackage{amssymb}
\usepackage{bm}
\usepackage{xfrac}
\usepackage{tensor}
\usepackage{eqparbox}
\usepackage{xcolor}
\usepackage{pifont}
\usepackage[nice]{nicefrac}
\usepackage[all,cmtip]{xy}
\DeclarePairedDelimiterX{\norm}[1]{\lVert}{\rVert}{#1}
\newcommand{\st}{\;|\;}
\newcommand{\stforall}{ \st \forall \; }

\newcommand{\C}{\mathbb{C}} 
\newcommand{\N}{\mathbb{N}} 
\newcommand{\transpose}{\intercal}

\newcommand{\functionSpace}{\mathcal{F}}

\newcommand{\G}[1][]{\mathbb{G}_{\scalebox{0.6}{$#1$}}}       
\newcommand{\KleinFourGroup}{\mathbb{K}_{4}}
\newcommand{\CyclicGroup}[1][]{\mathbb{C}_{#1}}        
\newcommand{\DihedralGroup}[1][]{\mathbb{D}_{#1}}      
\newcommand{\OG}[1][n]{\mathbb{O}_{#1}}              

\newcommand{\GLGroup}{\mathbb{GL}}          
\newcommand{\SO}[1][\dimEvolution]{\mathbb{SO}_{#1}}                 
\newcommand{\SE}[1][\dimEvolution]{\mathbb{SE}_{#1}}                 
\newcommand{\se}{\mathfrak{se}}                 
\newcommand{\so}{\mathfrak{so}}                 
\newcommand{\EG}[1][\dimEvolution]{\mathbb{E}_{#1}}
\newcommand{\mSO}{\mR}                 
\newcommand{\mSE}{\mX}                 
\newcommand{\mEG}{\mX}                 
\newcommand{\mso}[1][]{[{\angvel[#1]}]_\times}                 

\newcommand{\g}{g}       
\newcommand{\ginv}[1][]{g_{#1}^{\scalebox{0.7}{$-1$}}}       
\newcommand{\Glact}{\mathrel{\scalebox{0.8}{$\triangleright$}}}             
\newcommand{\Gconj}{\mathrel{\scalebox{0.8}{$\diamond$}}}                   
\newcommand{\Gcomp}{\mathrel{\scalebox{0.8}{$\circ$}}}                      

\newcommand{\Oplus}{\ensuremath{\vcenter{\hbox{\scalebox{1.1}{$\bm{\oplus}$}}}}}
\newcommand{\irrepMultiplicity}[1][]{m_{#1}}

\newcommand{\rep}[2][]{  
	{\rho_{
				{{\hbox{\scalebox{0.6}{$#1$}}}}
			}
			\def\temp{#2}\ifx\temp\empty
			\else
				(#2)%
			\fi
		}
}
\newcommand{\irrep}[2][]{
{\bar{\rho}_{\scriptscriptstyle{#1}}
\def\temp{#2}\ifx\temp\empty
\else
	(#2)%
\fi
}
}

\newcommand{\homomorphismDiag}[5]{
	\xymatrix{
		#1 \ar@{-}[r]^{#3}    \ar[d]^{#5}    & #1 \ar[d]^{#5} \\
		#2 \ar@{-}[r]^{#4}                   & #2
	}
}

\newcommand{\isomorphismDiag}[5]{
	\xymatrix{
		#1 \ar@{-}[r]^{#3}    \ar@{-}[d]^{#5} & #1 \ar@{-}[d]^{#5} \\
		#2 \ar@{-}[r]^{#4}                   & #2
	}
}

\newcommand{\isoCompNum}{{n_{\text{iso}}}}
\newcommand{\isoCompIdx}{{k}}
\newcommand{\isoCompIrrepIdx}{{p}}

\newcommand{\iso}{\text{iso}}

\newcommand{\identity}[1][]{\bm{1}}

\newcommand{\Time}{\mathbb{T}}

\newcommand{\measure}[1][]{\mu_{#1}}     

\newcommand{\kinE}{T}
\newcommand{\potE}{V}
\newcommand{\workFn}{U}
\newcommand{\genForces}{\bm{\tau}}
\newcommand{\genForcesJs}{\bm{\tau}_{\js}}
\newcommand{\dimEvolution}{d}

\newcommand{\confSpace}{\mathcal{Q}}
\newcommand{\confBundle}{\mathcal{T}\confSpace}
\newcommand{\confSpaceJS}{\mathcal{M}}
\newcommand{\tangConfSpace}[1][\q]{\mathcal{T}_{#1}\confSpace}

\newcommand{\tangConfSpaceJS}{\mathcal{T}_{\q}\confSpaceJS}

\newcommand{\nj}{{n_{j}}}   
\newcommand{\nb}{{n_{b}}}    
\newcommand{\nch}{{n_{rep}}}   
\newcommand{\nchdof}{{n_{dof}}}   
\newcommand{\nuch}{{n_{k}}}   
\newcommand{\setKinStructLabels}{\sS}

\newcommand{\baseSE}{\mX_{\base}}
\newcommand{\basese}{\dot{\mX}_{\base}}

\newcommand{\momentum}[1][]{\vh_{#1}}       
\newcommand{\angMomentum}[1][]{\vk_{#1}}    
\newcommand{\linMomentum}[1][]{\vl_{#1}}    
\newcommand{\CMM}{\mA_{C}}
\newcommand{\pos}{\vr}           
\newcommand{\vel}{\dot{\pos}}     
\newcommand{\acc}{\ddot{\pos}}     
\newcommand{\angvel}[1][]{\boldsymbol{\mathit{w}}_{#1}}  

\newcommand{\Jacob}[1][]{J_{{#1}}}
\newcommand{\PosJacob}[1][]{J_{t_{#1}}}
\newcommand{\OriJacob}[1][]{J_{R_{#1}}}

\newcommand{\Mass}{\bm{M}}
\newcommand{\mass}[1][]{m_{#1}}
\newcommand{\Inertia}[1][]{\mI_{#1}}

\newcommand{\base}{{\scalebox{0.6}{$B$}}}
\newcommand{\smallbase}{\scalebox{0.6}{$B$}}

\newcommand{\state}{(\q,\dq)}           

\newcommand{\isometryState}[1][\g]{(#1 \Glact \q, #1 \Glact \dq)}           
\newcommand{\morphState}[1][\g]{(#1 \morphOp \q, #1 \morphOp \dq)}           

\newcommand{\morphOp}{\mathrel{\scalebox{0.8}{\raisebox{-2pt}{\stackon[0.5pt]{$\triangleright$}{$\diamond$}}}}}

\newcommand{\q}[1][]{\vq_{#1}}                    
\newcommand{\dq}[1][]{\dot{\vq}_{#1}}                   
\newcommand{\ddq}[1][]{\ddot{\vq}_{#1}}            

\newcommand{\gq}[1][]{\g \Glact \q[#1]}                    
\newcommand{\gdq}[1][]{\g \Glact \dq[#1]}                    

\newcommand{\js}{\text{js}}
\newcommand{\qj}[1][\js]{\q[#1]}                      
\newcommand{\dqj}[1][\js]{\dq[#1]}                    
\newcommand{\ddqj}[1][\js]{\ddq[#1]}
\newcommand{\tauj}[1][\js]{\bm{\tau}_{#1}}                 
\newcommand{\workj}[1][\js]{W_{#1}}                 
\newcommand{\momentumj}[1][\js]{\vp_{#1}}                 

\newcommand{\qjiso}[1][\text{iso}]{\qj^{\scalebox{0.6}{$#1$}}}
\newcommand{\dqjiso}[1][\text{iso}]{\dqj^{\scalebox{0.6}{$#1$}}}
\newcommand{\ddqjiso}[1][\text{iso}]{\ddqj^{\scalebox{0.6}{$#1$}}}
\newcommand{\taujiso}[1][\text{iso}]{\tauj^{\scalebox{0.6}{$#1$}}}
\newcommand{\workjiso}[1][\text{iso}]{\workj^{\scalebox{0.6}{$#1$}}}
\newcommand{\momentumjiso}[1][\text{iso}]{\momentumj^{\scalebox{0.6}{$#1$}}}

\newcommand{\gqj}[1][\js]{\g \Glact \qj[#1]}                      
\newcommand{\gdqj}[1][\js]{\g \Glact \dqj[#1]}                    

\newcommand{\nq}{{n_q}}   

\newcommand{\Lagrangian}{L}

\newcommand{\equivObsSpaceDual}[1][_{\measure[t]}^{*\G}]
{
	\tensor*[]{\scalebox{0.8}{$\mathcal{X}$}}{#1}
}

\newcommand{\nnParams}{\boldsymbol{\phi}}
\newcommand{\nn}[1][\nnParams]{f_{#1}}
\newcommand{\nnIn}[1][]{\vx_{#1}}
\newcommand{\nnOut}[1][]{\vy_{#1}}

\newtheorem{theorem}{Theorem}

\newtheorem{lemma}{Lemma}

\newtheorem{definition}{Definition}

\newtheorem{proposition}{Proposition}  
\usepackage{tcolorbox}
\usepackage{xcolor}
\usepackage{physics}
\usepackage{amsmath}
\usepackage{float}
\usepackage{tikz}
\usepackage{mathdots}
\usepackage{yhmath}
\usepackage{cancel}
\usepackage{color}
\usepackage{siunitx}
\usepackage{array}
\usepackage{multirow}
\usepackage{amssymb}
\usepackage{gensymb}
\usepackage{extarrows}
\usepackage{booktabs}

\usepackage{hyperref}
\usepackage[nameinlink]{cleveref}   

\usepackage{graphicx}   
\usepackage{ulem}       
\usepackage{mathtools}  
\usepackage{multicol}   
\usepackage{adjustbox}  
\usepackage{wrapfig}    

\usepackage{enumitem}       
\usepackage{pifont}         

\setlist{leftmargin=5.5mm}  
\usetikzlibrary{fadings}
\usetikzlibrary{patterns}
\usetikzlibrary{shadows.blur}
\usetikzlibrary{shapes}

\definecolor{gray}{rgb}{0.6, 0.7, 0.7}
\definecolor{awesomeblue}{rgb}{0.054, 0.415, 0.505}
\definecolor{awesomeorange}{rgb}{0.570, 0.458, 0.0912}

\newcommand{\highlight}[1]{{\textit{#1}}}

\newtcolorbox{highlightBox}[3][]
{
	colframe = #2!25,
	colback  = #2!15,
	coltitle = #2!20!black,
	title    = {\textbf{#3}},
	#1,
}

\newcommand{\ubcolor}[3][awesomeblue]{{
			\color{#1}{
				\underbrace{\color{black}{#2}}_{#3}
			}
		}}

\usepackage{stackengine} 
\usepackage{mdframed} 
\usepackage{algorithm}
\usepackage{algpseudocode}\usepackage{subcaption}
\usepackage{physics}
\usepackage{amsmath}
\usepackage{tikz}
\usepackage{stfloats}
\usepackage{orcidlink}
\usepackage{nccmath} 
\usepackage{nccmath} 
\usepackage{mathdots}
\usepackage{yhmath}
\usepackage{cancel}
\usepackage{color}
\usepackage{siunitx}
\usepackage{array}
\usepackage{multirow}
\usepackage{amssymb}
\usepackage{gensymb}
\usepackage{tabularx}
\usepackage{extarrows}
\usepackage{booktabs}
\usetikzlibrary{fadings}
\usetikzlibrary{patterns}
\usetikzlibrary{shadows.blur}
\usetikzlibrary{shapes}

\usepackage{wrapfig} 

\usepackage{hyperref}
\usepackage{moreverb,url}
\hypersetup{
    colorlinks=true,
    bookmarksopen=true,
    bookmarksnumbered=true,
    citecolor=blue,
    urlcolor=blue,
    linkcolor=blue, 
	pdftitle={Morphological Symmetries in Robotics},
    pdfauthor={{Daniel Ordoñez Apraez, Giulio Turrisi, Vladimir Kostic, Mario Martin, Antonio Agudo, Francesc Moreno-Noguer, Massimiliano Pontil, Claudio Semini, and Carlos Mastalli}},
    pdfsubject={Submitted to The International Journal of Robotics Research (IJRR)},
    pdfkeywords={morphological symmetries, equivariant neural networks, geometric learning, discrete symmetries, continuous symmetries, data-driven modelling, robotics},
    pdftoolbar=true
}

\newcommand{\source}{ArXiv}  
\usepackage{glossaries}
\newacronym[
    name={NCM}, 
    description={
        The term Normal Configuration Mode (NCM) of motion, introduced in this work, denotes a unique mode of symmetry-constrained motion of a robot configuration. The qualifier "Normal" underscores the orthogonality between different NCMs, drawing parallels with the concept of Normal Vibrational Modes in molecular dynamics \citep{dresselhaus2007group_theory_applications_to_physics_of_condensed_matter}
    }, firstplural={normal configuration modes (NCMs)}
    ]{ncm}{NCM}{normal configuration mode}
\newacronym[
    name={DHA}, 
    description={
        The term Dynamics' Harmonics Analysis, introduced in \citet{ordonez2024dynamics}, refers to the decomposition of the system's dynamics into a superposition of lower-dimensional, independent dynamics. This concept is inspired from the goal of abstract harmonic analysis, namely, to decompose functions (in this case, the system's set of differential equations of motion) into a superposition of simpler, symmetric functions (in this case the dynamics of each Normal Configuration Mode (NCM) of motion). This use of abstract harmonic analysis is also referred to the literature as a generalized Fourier transformation \citep{chirikjian2000engineering}
    },
    ]{dha}{DHA}{Dynamics' Harmonic Analysis}
\newacronym{eom}{EoM}{equations of motion}
\newacronym[
    description={
        The center of mass refers to the centroidal point of a distribution of mass in space. This is often referred to as the barycenter or balance point 
    }
    ]{com}{CoM}{center of mass}
\newacronym{dof}{DoF}{degrees of freedom}
\newacronym{cmm}{CMM}{centroidal momentum matrix}
\newacronym{mdp}{MDP}{Markov decision process}
\newacronym{mlp}{MLP}{multi layer perceptron}
\newacronym{cnn}{CNN}{convolutional neural network}
\newacronym{cnn-aug}{CNN\textsubscript{aug}}{CNN trained with data augmentation}
\newacronym{ecnn}{eCNN}{equivariant CNN}
\newacronym{mlp-aug}{MLP\textsubscript{aug}}{MLP with data augmentation}
\newacronym{emlp}{eMLP}{equivariant multi layer perceptron}

\makeglossaries

\crefname{definition}{definition}{definitions}
\crefname{proposition}{proposition}{propositions}
\crefname{theorem}{theorem}{theorems}
\crefname{figure}{figure}{figures}
\crefname{equation}{eq.}{eqs.}
\crefname{section}{section}{sections}
\crefname{subsection}{section}{sections}
\crefname{subsubsection}{section}{sections}
\crefname{algorithm}{algorithm}{algorithms}
\Crefname{definition}{Definition}{Definitions}
\Crefname{proposition}{Proposition}{Propositions}
\Crefname{theorem}{Theorem}{Theorems}
\Crefname{figure}{Figure}{Figures}
\Crefname{equation}{Equation}{Equations}
\Crefname{section}{Section}{Sections}
\Crefname{subsection}{Section}{Sections}
\Crefname{subsubsection}{Section}{Sections}
\Crefname{algorithm}{Algorithm}{Algorithms}

\begin{document}

\runninghead{Morphological symmetries in robotics}

\title{Morphological symmetries in robotics}

\author{Daniel Ordoñez Apraez\affilnum{1,2}\orcidlink{0000-0002-9793-2482}\,,
	Giulio Turrisi\affilnum{1}\orcidlink{0000-0003-3007-3553}\,,
	Vladimir Kostic\affilnum{2}\orcidlink{0000-0001-8341-1400}\,,
	Mario Martin\affilnum{3}\,,
	Antonio Agudo\affilnum{3}\orcidlink{0000-0001-6845-4998}\,,
	Francesc Moreno-Noguer\affilnum{3}\orcidlink{0000-0002-8640-684X}\,,
	Massimiliano Pontil\affilnum{2}\orcidlink{0000-0001-9415-098X}\,,
	Claudio Semini\affilnum{1}\orcidlink{0000-0002-3034-4686}\,,
	and Carlos Mastalli\affilnum{4,5}\orcidlink{0000-0002-0725-4279}
}

\affiliation{
	\affilnum{1}Dynamic Legged Systems -- Istituto Italiano di Tecnologia (IIT)\\
	\affilnum{2}Computational Statistics and Machine Learning -- IIT\\
	\affilnum{3}Institut de Robòtica i Informàtica Industrial, CSIC-UPC\\
	\affilnum{4}Robot Motor Intelligence -- Heriot-Watt University\\
	\affilnum{5}IHMC Robotics -- Florida Institute for Human \& Machine Cognition
	\vspace{-1em}
}

\corrauth{Daniel Ordoñez Apraez.
	Dynamic Legged Systems \& Computational Statistics and Machine Learning, Istituto Italiano di Tecnologia (IIT), Via Morego 30, 16163 Genova, Italy.
	\\
	Email: \href{mailto:daniel.ordonez@iit.it}{daniel.ordonez@iit.it}
}


\begin{abstract}
	We present a comprehensive framework for studying and leveraging morphological symmetries in robotic systems. These are intrinsic properties of the robot's morphology, frequently observed in animal biology and robotics, which stem from the replication of kinematic structures and the symmetrical distribution of mass. We illustrate how these symmetries extend to the robot’s state space and both proprioceptive and exteroceptive sensor measurements, resulting in the equivariance of the robot’s equations of motion and optimal control policies. Thus, we recognize morphological symmetries as a relevant and previously unexplored physics-informed geometric prior, with significant implications for both data-driven and analytical methods used in modeling, control, estimation and design in robotics. For data-driven methods, we demonstrate that morphological symmetries can enhance the sample efficiency and generalization of machine learning models through data augmentation, or by applying equivariant/invariant constraints on the model's architecture. In the context of analytical methods, we employ abstract harmonic analysis to decompose the robot's dynamics into a superposition of lower-dimensional, independent dynamics. We substantiate our claims with both synthetic and real-world experiments conducted on bipedal and quadrupedal robots. Lastly, we introduce the repository \textsc{\href{https://github.com/Danfoa/MorphoSymm}{MorphoSymm}} to facilitate the practical use of the theory and applications outlined in this work.
\end{abstract}


\renewcommand{\journalname}{International Journal of Robotics Research}
\maketitle

\section{Introduction} \label{sec:introduction}
%
\begin{figure*}[t!]
	\centering
	\includegraphics[width=\textwidth]{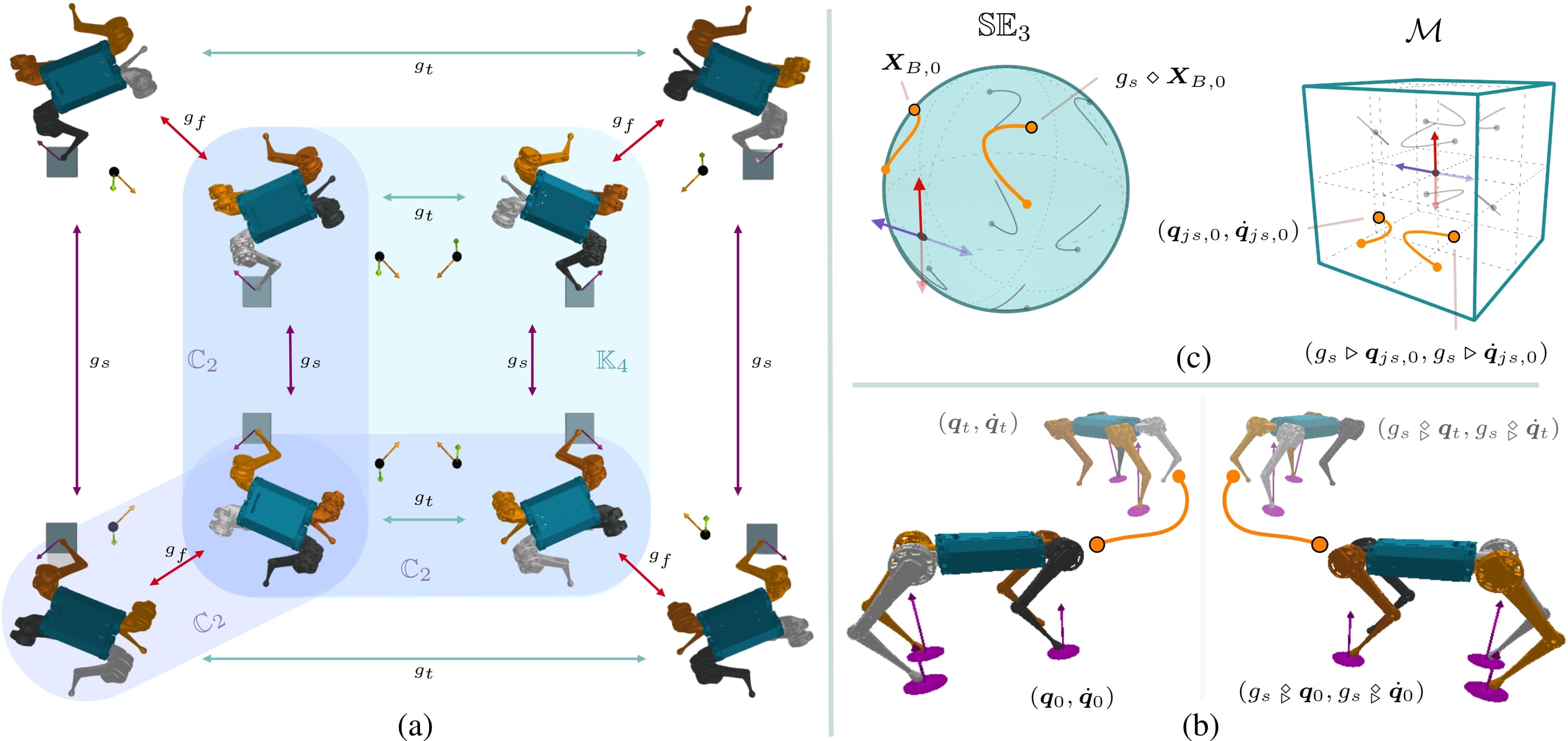}
	\vspace*{-5mm}
	\caption[Morphological symmetries of the Mini Cheetah robot]{
		(a) \href{https://en.wikipedia.org/wiki/Cayley_graph}{Caley diagram} of the morphological symmetries of the Mini Cheetah quadruped robot \citep{katz2019mini} (see
		\href{https://danfoa.github.io/MorphoSymm/static/animations/mini_cheetah-C2xC2xC2-symmetries_anim_static.gif?utm_source=\source}{animation}\endnote{
			\noindent
			Morphological symmetry group of the Mini Cheetah robot
			$\G = \CyclicGroup[2] \times \CyclicGroup[2] \times \CyclicGroup[2]$:
			See animation at
			\href{https://danfoa.github.io/MorphoSymm/static/animations/mini_cheetah-C2xC2xC2-symmetries_anim_static.gif?utm_source=\source}{https://bit.ly/MiniCheetah-MorphoSymm}
			\label{animation:mini_cheetah-C2xC2xC2-symmetries_anim_static}
		}). The robot's symmetries relate any system state with eight distinct symmetric states, each exhibiting identical or equivalent dynamics. These symmetries stem from the robot's ability to reconfigure its state, emulating three orthogonal reflections in $3D$ space: reflection with respect to the $yz$-plane ($\g_s$), $xz$-plane ($\g_t$), and $xy$-plane ($\g_f$). The robot's symmetry group is composed of the three reflections and any composition of these transformations (e.g., $\g_s \Gcomp \g_t$), resulting in the group $\G = \KleinFourGroup \times \CyclicGroup[2] = \CyclicGroup[2] \times \CyclicGroup[2] \times \CyclicGroup[2] = \{e, \g_s\} \times \{e, \g_t\} \times \{e, \g_f\}$
		(see \cref{fig:G_C2xC2xC2}).
		(b) Morphological symmetries manifest in the robot's dynamics, control policies, and both proprioceptive and exteroceptive sensor readings (e.g., contact points/forces, RGBD images). Consequently, any controlled trajectory of motion $[(\q[t],\dq[t])]_{t=0}^{T}$, along with the trajectory of sensor readings, can be augmented to feasible controlled trajectories and sensor readings for each of the $8$ symmetric states $[(\g \morphOp \q[t],\g \morphOp \dq[t])]_{t=0}^{T}$ for all $\g \in \G$ (see
		\href{https://danfoa.github.io/MorphoSymm/static/dynamic_animations/mini-cheetah_animation_C2xC2xC2.gif}{animation}\endnote{
			\noindent
			Equivariant temporal evolution of symmetric states of the Mini Cheetah robot
			See animation at
			\href{https://danfoa.github.io/MorphoSymm/static/dynamic_animations/mini-cheetah_animation_C2xC2xC2.gif}{https://bit.ly/mini-cheetah-equivariant-eom}
			\label{animation:mini_cheetah-dynamic_symmetric_temporal_evolution}
		}).
		(c) These trajectories, conceptualized as point trajectories evolving within the robot's configuration space $\confSpace := \SE[3] \times \confSpaceJS$, are decomposed into the $6$-dimensional manifold of the robot's base configurations (the special Euclidean group $\SE[3]$) and the $12$-dimensional manifold of joint space configurations $\confSpaceJS$. Morphological symmetries, imposing geometric constraints on $\SE[3]$ and $\confSpaceJS$, serve as a physics-informed geometric prior useful in robotics methods (refer to \cref{sec:applications}).
	}
	\vspace*{-0.4cm}
	\label{fig:teaser}
\end{figure*}

The field of robotics has recently witnessed a surge in the adoption of data-driven methods for modeling, estimation, and control. This trend is primarily fueled by several factors: (i) the ability of these methods to process complex data measurements, such as depth images, tactile sensing, and time-series data; (ii) their capacity to handle dynamic phenomena that are difficult to model, like friction and backlash; and (iii) their potential to bypass restrictive modeling/mathematical assumptions, such as the Markov property or ideal actuator dynamics. However, these methods often require high quantity and quality training data, which can be challenging, risky, or even impossible to obtain in robotics, especially for agile behaviors.

To mitigate this data dependency, contemporary machine learning methods aim to incorporate geometric priors, including symmetries and physics-informed inductive biases, into the learning process \citep{bronstein2021geometric,brehmer2023geometric2,weiler2023EquivariantAndCoordinateIndependentCNNs,de2023euclidean}. These priors have been instrumental in the success and interpretability of traditional computational methods for modeling, estimation, and control. In this work, we characterize the morphological symmetries of robot as a type of physics-informed geometric prior. When effectively leveraged in data-driven methods, these symmetries can enhance the sample efficiency and generalization of machine learning models for modeling, estimation, and control in robotics. Thus, this approach significantly mitigates the challenges of data collection in robotics.

Essentially, morphological symmetries are structural symmetries of a robot's body that allow the robot to reconfigure itself to mimic some spatial transformations (i.e., rotations, reflections and/or translations). The simplest and most common example of such symmetries is the bilateral/sagittal symmetry in humans and many animals, where each side of the body is a mirror image of the other. This symmetry allows individuals to easily mimic a mirrored body pose by simply permuting the roles of their arms and legs. While this bilateral symmetry is common in both humans and robotic systems, robots can exhibit a broader number of such symmetries. For instance, certain quadruped robots have the ability to mimic up to eight distinct spatial transformations, as illustrated in \cref{fig:teaser}.

The primary incentive for studying these symmetries in practice is that they serve as a significant geometric prior when modeling and controlling the temporal evolution of the robot. This is because morphological symmetries manifest in the robot's state vector space, equations of motion, control policies, and proprioceptive and exteroceptive sensor data measurements used to perceive the robot's environment. Intuitively, when a person encounters an obstacle with their right leg while walking, the optimal visuomotor control response, required to react and balance, is essentially a mirrored version of the reaction if the mirrored event happens with the left leg. This is because the dynamics of both mirrored body poses at the moment of contact are equivalent (up to a reflection), implying that the optimal control policy should also be symmetric and respond equivalently to these mirrored events. The same reasoning applies to the $8$ equivalent states of the Mini Cheetah  robot in \cref{fig:teaser}.

Despite the significance of this geometric prior, morphological symmetries are often overlooked in both data-driven and analytical methods for modeling, control, and estimation. This is primarily because analytical models of rigid body dynamics, commonly used in robotics, implicitly leverage these symmetries due to their accurate representation of physics. Therefore, incorporating morphological symmetries into data-driven methods can be seen as a way to transfer one of the key inductive biases in analytical models to the data-driven paradigm.

Furthermore, analytical models can significantly benefit from the explicit exploitation of these symmetries. As detailed in \cref{sec:joint_space_basis_harmonic_motions}, the symmetric structure of a robot state space can be leveraged, through abstract harmonic analysis, to identify a decomposition of the robot's state space into orthogonal subspaces. This decomposition allows us to model and control the robot's dynamics as a superposition of independent lower-dimensional dynamics.


\subsubsection*{Structural symmetries in physics} The symmetries we study in this paper in the context of robotics share similarities with structural symmetries in particle physics, often referred to as point-group symmetries \citep[8.3]{dresselhaus2007group_theory_applications_to_physics_of_condensed_matter}. In physics, these symmetries arise from the replication of identical particles or atoms within molecules, atoms, or crystal structures. The principles underlying these symmetries, along with the analytical and data-driven methods used to exploit them in both traditional and contemporary atomistic and molecular dynamics \citep{cornwell1997group,noe2020machine,jumper2021highly,klein2023equivariant}, can be transferred to robotics. However, the utilization and implications of these symmetries in robotics diverge significantly from those in physics.

Unlike physics, where the primary objective is to model dynamics, the focus in robotics is on controlling these dynamics. Moreover, the morphology of a robot is a stable and often controlled aspect of the system, unlike in physics where symmetries are subject to temporal changes with the gain or loss of particles. This stability in robotics renders these symmetries a consistent inductive bias, unaffected by time, specific motion tasks, or the operational environment. These properties, combined with the precision in the design and manufacturing of mechanical systems, and the numerical benefits associated with morphological symmetries, provide a design space of robotic systems for optimally leveraging this geometric prior.
\subsection*{Contributions}
This work introduces a comprehensive theoretical and practical framework for studying and leveraging morphological symmetries in robotics. These structural symmetries, inherent of a robot's body, extend to symmetries in the system's state space and dynamics. Our aim is to formally define these symmetries, characterize their conditions of existence, and elucidate their implications in both analytical and data-driven methods. While these symmetries are readily identifiable in simple systems with basic symmetry groups, such as bilateral symmetry, our theoretical framework paves the way for identifying and exploiting morphological symmetries in complex real-world robotic systems. This includes humanoids, quadrupeds, wheeled robots, drones, modular, soft, and continuum robots. Our contributions, which can be summarized as follows, involve:
\begin{itemize}[topsep=5pt,itemsep=5pt, partopsep=0pt, parsep=0pt]
	\item Characterizing how morphological symmetries manifest as symmetries in the system's state space, sensor readings,   the equivariance of the equations of motion and optimal control policies.
	\item Identifying the generalized mass matrix's equivariance as the defining property of symmetric robotic systems.
\end{itemize}
In the context of rigid body dynamics, this enable us to:
\begin{itemize}[topsep=5pt, itemsep=0pt, partopsep=0pt,parsep=0pt]
	\item[\ding{104}] Establish the conditions for morphological symmetry existence, defined as analytical constraints on the robot's kinematic and dynamic parameters (\cref{sec:constraints_on_the_floating,sec:constraints_on_the_kinematic_structure}). Using these, we introduce an algorithm for the systematic identification of these symmetries (\cref{sec:algorithmic_id_ms}).
	\item[\ding{104}] Apply abstract harmonic analysis to exploit the symmetric structure in the robot's joint space (\cref{sec:joint_space_basis_harmonic_motions}). This technique enables the decomposition of joint space dynamics into a superposition of simpler, independent, lower-dimensional dynamical systems, a concept we term as \gls{dha}.
\end{itemize}
%
To promote the use of the theory and tools developed in this work, we present the repository \textsc{\href{https://github.com/Danfoa/MorphoSymm}{MorphoSymm}},\endnote{Open access repository: \href{https://github.com/Danfoa/MorphoSymm}{github.com/Danfoa/MorphoSymm}\label{note:morphosymm}
} which facilitates the use of data augmentation, \gls{dha}, and the construction of equivariant/invariant neural networks, using \textsc{\href{https://github.com/QUVA-Lab/escnn}{escnn}} \citep{cesa2021program}, for a growing \href{https://github.com/Danfoa/MorphoSymm?tab=readme-ov-file#library-of-symmetric-dynamical-systems}{library} of symmetric robots.

This paper extends our previous work \citep{ordonez-discretesymm}, by providing a more comprehensive analysis of morphological symmetries with an extended study on the conditions for their existence in rigid body dynamics, and by introducing the use of abstract harmonic analysis to leverage the symmetric structure of the robot's joint space dynamics.
%
\subsection*{Outline}
We begin by introducing group and representation theory from a robotics perspective in \cref{sec:background}, followed by a discussion on symmetries in Lagrangian mechanics in \cref{sec:symmetries_in_lagrangian_mechanics}. The concept of morphological symmetries is presented in \cref{sec:morphological_symmetries}, while \cref{sec:morphological_symmetries_rigid_body} delves into their application in rigid body dynamics, including the conditions for symmetry existence and the identification of the symmetry group. There, we also introduce the use of harmonic analysis on the system's joint space. \Cref{sec:applications} provides an overview of morphological symmetries applications, such as data augmentation, equivariant/invariant function approximation, and Dynamics' Harmonics Analysis. Finally, \cref{sec:experiments} presents our experiments and results, with concluding remarks given in \cref{sec:conclusion}.
%
\section{Background}
\label{sec:background}
%
Here, we provide an introduction to group and representation theory, the mathematical frameworks essential for studying symmetries. We aim to familiarize readers with the key concepts and notation that will be used throughout this article. As these fields might be unfamiliar in the context of robotics, we suggest an initial read-through for a general understanding, and revisiting each concept as they are referenced in the following sections. For a more comprehensive introduction, we recommend the works of \citet{weiler2023EquivariantAndCoordinateIndependentCNNs} for a machine learning perspective, and \citet{selig2005geometric_fundamentals_robotics,lanczos2020variational_principles_mechanics} for insights into robotics.
\subsubsection*{Symmetry Groups}
In essence, a symmetry of an object, whether it be a vector space, a manifold, or a robot, refers to an invertible transformation that preserves a significant property of the object, such as the vector space metric, the curvature of the manifold, or the energy of the robot.

Group theory studies sets of symmetry transformations as abstract mathematical constructs independent of the objects with which these symmetries are associated. This abstraction proves beneficial in our work, as it facilitates the analysis of the symmetries inherent in the laws of physics, various robotic systems, and sensor data measurements, all as distinct manifestations of the same symmetry group.

In its abstract form, a symmetry group can be expressed as a set of invertible symmetry transformations, denoted as $\G=\{e,\, \g_{1},\, \ginv[1],\, \g_{2},\, \dots\}$. This set is closed under the operations of \textit{composition} $\Gcomp: \G \times \G \mapsto \G$, and \textit{inversion} $(\cdot)^{\text{-}1}: \G \mapsto \G$. This implies that for any $\g_{1},\g_{2} \in \G$, the composition $\g_{1} \Gcomp \g_{2}$ is also a member of $\G$. Similarly, for any $\g \in \G$, the inverse $\g^{\text{-}1}$ belongs to $\G$ and satisfies $\g^{\text{-}1}  \Gcomp g = e$, where $e$ is the identity (or trivial) transformation. Groups are characterized by their structure, defined by their order $|\G|$, which represents the number of unique elements in the group, and the composition rules of these elements. For example, consider the simplest group, the reflection group $\CyclicGroup[2] = \{e, \g_{s} \;| \g_{s}^2 = e\}$, composed of the trivial transformation and a reflection transformation $\g_{s}$, characterized by $\g_{s}^2 := \g_{s}\Gcomp\g_{s} = e$.

\begin{wrapfigure}{r}{0.20\textwidth}
	\centering
	\resizebox{\linewidth}{!}{\tikzset{every picture/.style={line width=0.75pt}} 

\begin{tikzpicture}[x=0.75pt,y=0.75pt,yscale=-1,xscale=1]

\draw  [draw opacity=0][fill={rgb, 255:red, 135; green, 225; blue, 250 }  ,fill opacity=0.37 ] (4.67,161.38) .. controls (4.67,153.71) and (10.88,147.5) .. (18.54,147.5) -- (116.63,147.5) .. controls (124.29,147.5) and (130.5,153.71) .. (130.5,161.38) -- (130.5,161.38) .. controls (130.5,169.04) and (124.29,175.25) .. (116.63,175.25) -- (18.54,175.25) .. controls (10.88,175.25) and (4.67,169.04) .. (4.67,161.38) -- cycle ;
\draw  [draw opacity=0][fill={rgb, 255:red, 135; green, 225; blue, 250 }  ,fill opacity=0.37 ] (116.63,45.75) .. controls (124.29,45.75) and (130.5,51.96) .. (130.5,59.63) -- (130.5,161.13) .. controls (130.5,168.79) and (124.29,175) .. (116.63,175) -- (116.63,175) .. controls (108.96,175) and (102.75,168.79) .. (102.75,161.13) -- (102.75,59.63) .. controls (102.75,51.96) and (108.96,45.75) .. (116.63,45.75) -- cycle ;
\draw  [draw opacity=0][fill={rgb, 255:red, 135; green, 225; blue, 250 }  ,fill opacity=0.37 ] (164.07,89.7) .. controls (170.47,93.91) and (172.24,102.52) .. (168.03,108.92) -- (128.43,169.06) .. controls (124.22,175.46) and (115.61,177.24) .. (109.21,173.02) -- (109.21,173.02) .. controls (102.81,168.81) and (101.04,160.2) .. (105.25,153.8) -- (144.85,93.66) .. controls (149.06,87.26) and (157.67,85.49) .. (164.07,89.7) -- cycle ;

\draw (21,59) node  [color={rgb, 255:red, 74; green, 74; blue, 74 }  ,opacity=1 ]  {$g_{s} \Gcomp g_{t}$};
\draw (71,9) node  [font=\small,color={rgb, 255:red, 74; green, 74; blue, 74 }  ,opacity=1 ]  {$g_{s} \Gcomp g_{t} \Gcomp g_{f}$};
\draw (20,159.5) node  [color={rgb, 255:red, 74; green, 74; blue, 74 }  ,opacity=1 ]  {$g_{s}$};
\draw (158.56,9) node  [color={rgb, 255:red, 74; green, 74; blue, 74 }  ,opacity=1 ]  {$g_{t} \Gcomp g_{f}$};
\draw (119,59) node  [color={rgb, 255:red, 74; green, 74; blue, 74 }  ,opacity=1 ]  {$g_{t}$};
\draw (71,99) node  [color={rgb, 255:red, 74; green, 74; blue, 74 }  ,opacity=1 ]  {$g_{s} \Gcomp g_{f}$};
\draw (159,99) node  [color={rgb, 255:red, 74; green, 74; blue, 74 }  ,opacity=1 ]  {$g_{f}$};
\draw (118,159.5) node  [color={rgb, 255:red, 74; green, 74; blue, 74 }  ,opacity=1 ]  {$e$};
\draw [color={rgb, 255:red, 208; green, 2; blue, 27 }  ,draw opacity=1 ][line width=1.5]    (106,9) -- (133.06,9) ;
\draw [color={rgb, 255:red, 35; green, 142; blue, 151 }  ,draw opacity=1 ][line width=1.5]    (34,46) -- (59,21) ;
\draw [color={rgb, 255:red, 208; green, 2; blue, 27 }  ,draw opacity=1 ][line width=1.5]    (45.5,59) -- (107.5,59) ;
\draw [color={rgb, 255:red, 35; green, 142; blue, 151 }  ,draw opacity=1 ][line width=1.5]    (129.28,46) -- (148.27,22) ;
\draw [color={rgb, 255:red, 245; green, 166; blue, 35 }  ,draw opacity=1 ][line width=1.5]    (20.87,72) -- (20.13,146.5) ;
\draw [color={rgb, 255:red, 208; green, 2; blue, 27 }  ,draw opacity=1 ][line width=1.5]    (31.5,159.5) -- (109,159.5) ;
\draw [color={rgb, 255:red, 35; green, 142; blue, 151 }  ,draw opacity=1 ][line width=1.5]    (126.13,147.5) -- (150.19,112) ;
\draw [color={rgb, 255:red, 35; green, 142; blue, 151 }  ,draw opacity=1 ][line width=1.5]    (30.96,146.5) -- (60.04,112) ;
\draw [color={rgb, 255:red, 208; green, 2; blue, 27 }  ,draw opacity=1 ][line width=1.5]    (96.5,99) -- (115,99)(123,99) -- (146.5,99) ;
\draw [color={rgb, 255:red, 245; green, 166; blue, 35 }  ,draw opacity=1 ][line width=1.5]    (71,21) -- (71,53.4)(71,61.4) -- (71,86) ;
\draw [color={rgb, 255:red, 245; green, 166; blue, 35 }  ,draw opacity=1 ][line width=1.5]    (158.62,22) -- (158.94,86) ;
\draw [color={rgb, 255:red, 245; green, 166; blue, 35 }  ,draw opacity=1 ][line width=1.5]    (118.87,72) -- (118.12,147.5) ;

\end{tikzpicture}}
	\caption{
		\small
		Structure of the group
		$\G = \CyclicGroup[2] \times \CyclicGroup[2] \times \CyclicGroup[2]$
		\label{fig:G_C2xC2xC2}
	}
	\vspace*{-0.35cm}
\end{wrapfigure}
\subsubsection*{Group structure and subgroups}
%
To comprehend the structure of a (symmetry) group, it is often beneficial to break it down into simpler symmetry groups and analyze how these smaller groups interact to form the larger structure.
For example, the symmetry group $\G$ of the Mini Cheetah robot in \cref{fig:teaser}, is of order $|\G|=8$, and can be defined as the direct product of three distinct reflection groups, $\G = \CyclicGroup[2] \times \CyclicGroup[2] \times \CyclicGroup[2]$ as depicted in \cref{fig:G_C2xC2xC2}.
A subset of a group of symmetries, $\G' \subset \G$, is a symmetry subgroup if the subset is closed under composition and inversion, which is denoted as $\G'<\G$. In our example above, the group $\G$ contains three distinct subgroups of two elements each: $\{e, \g_{s}\}, \{e, \g_{t}\}$, and $\{e, \g_{f}\}$, depicted in~\cref{fig:G_C2xC2xC2}. Although these three reflections describe distinct transformations, these groups are structurally equivalent or \highlight{isomorphic} (iso$\sim$``same'' and morphic$\sim$``shape/structure'') to a reflection group. This means that, there exists a bijective map between the elements of the two groups, and this map preserves the group's structure. We denote the isomorphism property as $\{e, \g_{s}\} \cong \{e, \g_{t}\} \cong \{e, \g_{f}\} \cong \CyclicGroup[2]$.
\subsubsection*{Group actions}
We are mostly interested in the action of a symmetry transformation on specific objects, such as the robot's state and sensor data measurements. Therefore, we need to define how specific objects are transformed by an element $\g \in \G$. Formally, the action of a symmetry transformation $\g \in \G$ on a set $\sS$ is defined as a map $\Glact: \G \times \sS \mapsto \sS$, taking an element of the group and set and returning another set element, i.e., $\g \Glact s \in \sS$ if $s\in \sS$. This map is \textit{associative} under group composition, $(\g_1 \Gcomp \g_2) \Glact s = \g_1 \Glact (\g_2 \Glact s)$, and respects the identity transformation $e \Glact s = s$.
\subsection*{Group representations}
To define a symmetry transformation not on a set but on a vector space $\vsX \subseteq \R^n$, we utilize a \highlight{group representation}. By definition, a group representation on the space $\vsX$ is a map $\rep[\vsX]{}:\G\mapsto\GLGroup(\vsX)$ that assigns invertible linear maps to group elements. When choosing a basis set for the vector space, this linear maps will be represented as matrices, enabling us to model the group composition operator as matrix-matrix multiplication, i.e., $\rep[\vsX]{\g_1 \Gcomp \g_2} = \rep[\vsX]{\g_1} \rep[\vsX]{\g_2}$, symmetry inversion by matrix inversion, $\rep[\vsX]{\g^{\text{-}1}} = \rep[\vsX]{\g}^{\text{-}1}$, and the action of a symmetry on a point $\vx \in \vsX$ to be expressed as a matrix-vector multiplication, i.e., $\g \Glact \vx := \rep[\vsX]{\g} \vx \in \vsX$. Note that $\GLGroup(\vsX)$ is the group of invertible linear maps on $\vsX$. Whenever a vector space possesses a group representation, we refer to it as a \highlight{symmetric space}.

\subsubsection*{$\G$-invariant and $\G$-equivariant maps}
A map between two symmetric vector spaces $f:\vsX \rightarrow \vsY$ often falls within two categories: group invariant or group equivariant. The map $f$ is considered $\G$-invariant if its output remains unchanged regardless of the transformation applied to the input. Instead, a map is considered $\G$-equivariant when the result of applying a transformation to the input, followed by computing the function, is equivalent to first computing the function and then applying the transformation to the output. Formally, these conditions can be expressed as follows:

\begin{equation}
	\small
	\ubcolor[awesomeorange]{
		\vy = f(\rep[\vsX]{\g} \vx)
	}{\text{$\G$-invariant}}
	\;\; \text{and} \;\;
	\ubcolor[awesomeblue]{
		\rep[\vsY]{\g} \vy = f(\rep[\vsX]{\g} \vx)
	}{\text{$\G$-equivariant}}
	\stforall \g \in \G, \vx \in \vsX
	.
	\label{eq:equivariance-invariance-constraints}
\end{equation}%
%

\subsubsection*{Linear maps, change of basis and the conjugate action}
The action of a symmetry transformation $\g$ can be viewed as a point transformation, mapping each point $\vx \in \vsX$ to its symmetric counterpart $\g \Glact \vx$, or as a change of basis in $\vsX$. Specifically, if $\sS_\vsX = \{\bar{\vx}_0, \cdots, \bar{\vx}_n\}$ is a basis set for the vector space, the action of $\g$ on the basis set yields another valid basis set $\g \Glact \sS_\vsX := \{\g \Glact \bar{\vx}_0, \cdots, \g \Glact  \bar{\vx}_n\}$.

The action of a group element on a linear map's matrix representation, which redefines the map in the symmetry-transformed basis, is referred to as the \highlight{conjugate action}. Let $\mA:\vsX \mapsto \vsX$ be a linear map on the symmetric vector space $\vsX$. The map's matrix representation depends on the chosen basis for $\vsX$. Therefore, under a change of basis by a given $\g$, the map's matrix representation is redefined as:

\begin{equation}
	\small
	\g \Gconj \mA  = \rep[\vsX]{\g} \mA \rep[\vsX]{\g}^{\text{-}1}.
	\label{eq:conjugate_action}
\end{equation}%
%

\subsubsection*{The group of rigid transformations in Euclidean space}
In the context of robotics, a crucial group of symmetries is the group of Euclidean isometries (or rigid transformations) in $\dimEvolution$-dimensions. This group is usually denoted as the matrix Euclidean group $\EG[\dimEvolution] := \OG[\dimEvolution] \ltimes \mathbb{T}_{\dimEvolution}$ (typically, $\dimEvolution$ refers to a $3$-dimensional space, but we present some simplified examples in two dimensions $\dimEvolution=2$). Here, $\OG[\dimEvolution]$ represents the orthogonal group, which includes all rotations and reflections, and $\mathbb{T}_{\dimEvolution}$ represents the translation group in $\dimEvolution$-dimensional space.
To define a morphological symmetry, we study this group in is abstract (non-matrix) form. Specifically, let $\G[E] \cong \EG[\dimEvolution]$ denote a group \highlight{isomorphic} to $\EG[\dimEvolution]$, where transformations are abstracted from their usual $\dimEvolution$-dimensional space of action. Throughout this work, we define distinct representations for these symmetry transformations. However, when we use these symmetries as transformations of points, vectors, and spatial vectors in $\dimEvolution+1$-dimensions, we default to the standard homogenous matrix representation:

\begin{equation}
	\begin{aligned}
		\rep[\R^{d+1}]{\g}
		 & :=
		\mEG_\g
		=
		\begin{bsmallmatrix}
			\mSO_\g & \pos_\g \\
			\bm{0} & 1
		\end{bsmallmatrix} \in \EG,
		\\
		\rep[\R^{d}]{\g}
		 & :=
		\mSO_\g \in \OG[\dimEvolution] \quad \stforall \g \in \G[E],
	\end{aligned}
	\label{eq:group_representations_rigid_transformations}
\end{equation}%
where $\rep[\R^{d+1}]{}$ represents each $g\in \G[E]$ as an homogeneous transformation matrix $\mEG_\g\in \EG$. This isometry is comprised of a rotation/reflection matrix $\mSO_\g$, a member of the orthogonal group $\OG[\dimEvolution]$, and a translation vector $\pos_\g \in \R^\dimEvolution$. The representation $\rep[\R^{d}]{}$, defaults to assigning the rotation/reflection matrix $\mSO_\g$ to each $\g \in \G[E]$. When discussing rotation, reflection, or translation actions in $\dimEvolution$-dimensions, we typically use standard robotics notation $\mSE_\g$ and $\mSO_\g$ instead of $ \rep[\R^{d+1}]{\g}$ and $ \rep[\R^{d}]{\g}$, respectively.
\subsubsection*{Inequivalent representations of Euclidean isometries}
The motivation behind the distinction between $\G[E]$ and the Euclidean group $\EG$, is to establish distinct, inequivalent representations of Euclidean isometries. In other words, to define alternative ways for applying rotations, reflections, and translations to the robot's state. Specifically, as detailed in \cref{sec:morphological_symmetries,sec:morphological_symmetries_rigid_body}, we define a morphological symmetry by representing each $\g \in \G[E]$ as a transformation of the robot's joint space configuration.

In this context, two representations of the same group, defined on the same space $\rho^a_{\vsX},\rho^b_{\vsX}:\G\mapsto \GLGroup(\vsX)$, are considered equivalent, denoted as $\rho^a_{\vsX}(\g) \sim \rho^b_{\vsX}(\g)$, if they can be related by a change of basis in $\vsX$. That is, if there exist a linear map $\mT:\vsX \mapsto \vsX$, such that:

\begin{equation}
	\small
	\rho^a_{\vsX}(\g) \sim \rho^b_{\vsX}(\g) \quad \text{if} \quad \rho^a_{\vsX}(\g) = \mT \rho^b_{\vsX}(\g) \mT^{\text{-}1} \stforall \g \in \G.
	\label{eq:equivalent_representations}
\end{equation}%
\begin{figure*}[h!]
	\centering
	\includegraphics[width=\linewidth]{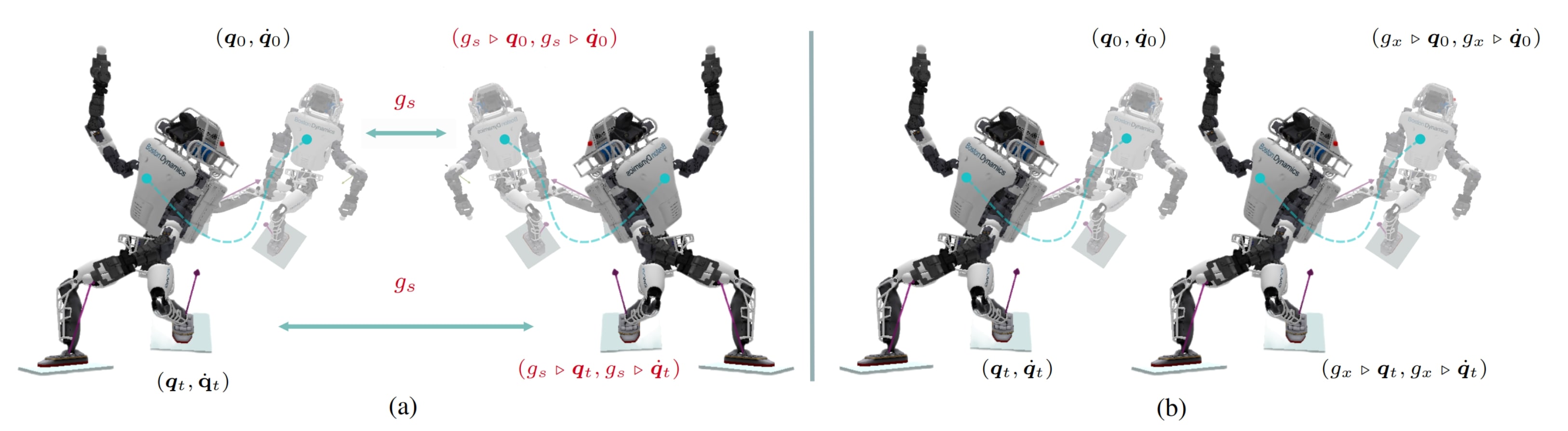}
	\vspace*{-.7cm}
	\caption{(a) The temporal evolution of the robot Boston Dynamics' Atlas robot, when subjected to a horizontal spatial reflection, results in an unreachable symmetric state $\isometryState[\g_s]$, consequence of the physical inability to apply reflections to rigid bodies \citep[2.5]{selig2005geometric_fundamentals_robotics}. Despite this, the dynamics of the original state $\state$ and its virtual counterpart are equivalent up to a reflection of space, maintaining their temporal evolution related by the symmetry $\g_s$. (b) Similarly, the temporal evolution of the original state and a horizontally translated state $\isometryState[\g_x]$ remain equivariant, with $\g_x$ being a feasible symmetry transformation.
		\vspace*{-1em}
	}
	\label{fig:atlas_dynamics_equivariance}
\end{figure*}

\subsubsection*{Decomposition of group representations}
Given two group representations, $\rep[\vsX]{}$ and $\rep[\vsY]{}$, acting on distinct vector spaces $\vsX$ and $\vsY$ respectively, we can construct a new group representation on the space $\vsX \times \vsY$. This construction, known as the direct sum, effectively combines the two representations into a block-diagonal form. Formally, the direct sum of two representations is defined as follows:
\begin{equation}
	\small
	\rep[\vsX \times \vsY]{\g} = \rep[\vsX]{\g} \oplus \rep[\vsY]{\g} :=
	\begin{bsmallmatrix}
		\rep[\vsX]{\g} & \bm{0} \\
		\bm{0} & \rep[\vsY]{\g}
	\end{bsmallmatrix} \stforall \g \in \G.
	\label{eq:direct_sum_representations}
\end{equation}%
\noindent
In \cref{sec:applications,sec:experiments}, we utilize the direct sum to build group representations on spaces derived from different sensor data measurements. Conversely, in \cref{sec:joint_space_basis_harmonic_motions}, we decompose the representation acting on the robot's joint space into a block-diagonal form. This decomposition, known as an \textit{isotypic decomposition} (a key result from abstract harmonic analysis), will provide numerical benefits for the analysis and the simulation of the system's dynamics.
\subsubsection*{Harmonic analysis and the isotypic decomposition}
Our use of harmonic analysis in modeling robotic systems reduces to leveraging the theoretical and numerical advantages of a specific basis of a symmetric vector space $\vsX$. This basis, known as the \highlight{isotypic basis}, has the property that it exposes the decomposition of $\vsX$ into \highlight{orthogonal} lower-dimensional subspaces, along with the decompostion of $\rep[\vsX]{}$ into a direct sum of representations on each subspace:
\begin{equation}
	\begin{aligned}
		\vsX         & = \vsX_1 \oplus^\perp \vsX_2 \oplus^\perp \cdots \oplus^\perp \vsX_\isoCompNum = {\Oplus^\perp}_{\isoCompIdx=1}^{\isoCompNum} \vsX_\isoCompIdx,
		\\
		\rep[\vsX]{} & \sim \rep[\vsX_1]{} \oplus \rep[\vsX_2]{} \oplus \cdots \oplus \rep[\vsX_\isoCompNum]{} = \Oplus_{\isoCompIdx=1}^{\isoCompNum} \rep[\vsX_\isoCompIdx]{}.
		\label{eq:isotypic_decomposition}
	\end{aligned}
\end{equation}%
\noindent
The number of isotypic subspaces is determined by the $\isoCompNum$ unique irreducible representations (\textsl{irreps}) $\{\irrep[\isoCompIdx]{}\}_{\isoCompIdx=1}^{\isoCompNum}$ of the group $\G$. Each isotypic subspace's representation $\rep[\vsX_\isoCompIdx]{}$ is constructed as the direct sum of multiple copies of the same type of irreducible representations, i.e., $\rep[\vsX_\isoCompIdx]{} \sim \irrep[\isoCompIdx]{} \oplus \dots \oplus \irrep[\isoCompIdx]{}$.

These \textsl{irreps} are the fundamental building blocks of any group representation of $\G$. Each \textsl{irrep},
$
	\irrep[\isoCompIdx]{}: \G \rightarrow \GLGroup({\bar{\vsX}_{\isoCompIdx}})
$, describes a unique symmetry pattern, characterized by a subset of symmetry transformations within the group, and the associated space ${\bar{\vsX}_{\isoCompIdx}}$ is the smallest finite-dimensional space that can express the \textsl{irrep} symmetry pattern. For instance, as our groups $\G$ will be subgroups of the Euclidean group, an example  \textsl{irrep} is that of a reflection symmetry ($\irrep[r]{}$), which require a $1$-dimensional line to act on ($\bar{\vsX}_{r} \sim \R$), or \textsl{irreps} describing rotations by an angle $\nicefrac{2\pi}{a}$, $\irrep[\nicefrac{2\pi}{a}]{}$, which require a $2$-dimensional space ($\bar{\vsX}_{\nicefrac{2\pi}{a}} \sim \R^2 \sim \C$) to act on.

This implies that as $\rep[\vsX_\isoCompIdx]{} \sim \irrep[\isoCompIdx]{} \oplus \dots \oplus \irrep[\isoCompIdx]{}$, the associated isotypic subspace is equivalent to $\vsX_{\isoCompIdx} \sim \bar{\vsX}_{\isoCompIdx} \oplus \dots \oplus \bar{\vsX}_{\isoCompIdx}$. Consequently, $\vsX_{\isoCompIdx}$ will only feature the subset of symmetry transformations associated with the \textsl{irrep} $\irrep[\isoCompIdx]{}$. This property is what ensures the orthogonality between isotypic subspaces (see Shur's lemma \citep[Prop 1.5]{Knapp1986}), a property we exploit in \cref{sec:joint_space_basis_harmonic_motions} to decompose the robot's dynamics.

Lastly, we highlight the isotypic decomposition is achieved simply by finding the appropriate change of basis $\mT: \vsX \mapsto \vsX$ that exposes the block-diagonal structure descrived in \cref{eq:isotypic_decomposition}. For further details, refer to \citet[Thm-2.5]{golubitsky2012singularities_groups_bifurcation} or \citet{ordonez2024dynamics}.
%
\section{Symmetries in Lagrangian mechanics} \label{sec:symmetries_in_lagrangian_mechanics}
In the context of dynamical systems, a symmetry is interpreted as a transformation that associates distinct system states with identical or equivalent dynamics. Essentially, this means that if two states are related by a symmetry transformation, the temporal evolution of one mirrors the symmetry-transformed temporal evolution of the other (see \cref{fig:atlas_dynamics_equivariance}a and animation \ref{animation:mini_cheetah-dynamic_symmetric_temporal_evolution}).

To formalize this property in robotics, we shall study symmetries from the lens of Lagrangian mechanics, a modeling framework applicable to a wide range of robot types, including rigid body, continuum, soft, and modular robots. In this framework, the robot's state is determined by the position and velocities of its $\nq$ \gls{dof}, defined as the independent generalized position coordinates $\q \in \confSpace \subseteq \R^{\nq}$ and velocity coordinates $\dq \in \tangConfSpace \subseteq \R^{\nq}$. Where $\confSpace$ denotes the constrained configuration space (a smooth manifold) \citep[chpt 1.5]{lanczos2020variational_principles_mechanics}, and $\tangConfSpace$ represents the configuration tangent space at $\q$ (a local tangent plane). Thus, the system's state is numerically represented as a point in the \highlight{phase space} $(\q, \dq) \in \confBundle := \confSpace \times \tangConfSpace \subseteq \R^{2\nq}$. Furthermore, the system's dynamics (i.e., the temporal evolution of points in $\confBundle$) are governed by the \gls{eom}. These are derived from the Euler-Lagrange differential equations, according to the principle of least action~\citep[II.11]{lanczos2020variational_principles_mechanics}:

\begin{equation}
	\small
	\frac{d}{dt} \frac{\partial \Lagrangian(\q, \dq)}{\partial \dq} - \frac{\partial \Lagrangian(\q, \dq)}{\partial \q} = \bm{0}
	\iff
	\ubcolor[awesomeblue]{
		M(\q)\ddq}{\text{inertial}}
	=
	\ubcolor[awesomeorange]{
		\genForces(\q,\dq)}{\text{moving}},
	\label{eq:eom_lagrangian_mechanics}
\end{equation}%
which, in their general form, can be stated as an equivalence between inertial and moving forces in the space of generalized coordinates. In this space, $M: \confSpace \mapsto \R^{\nq \times \nq}$ represents the generalized mass matrix function, and $\genForces: \confBundle \mapsto \R^{\nq}$ denotes the generalized force field function, determining the effective force result of control actions, contacts, internal constraints such as joint limits, gravitational forces, and external disturbances. These \gls{eom} are entirely derived from the Lagrangian scalar function $\Lagrangian: \confBundle \mapsto \R_+$, which measures the excess of kinetic energy $\kinE:  \confBundle \mapsto \R_+$ over the energy associated with mechanical work function\endnote{
	The mechanical work function $\workFn: \confBundle \mapsto \R_+$ encompasses the system's potential energy (determined by all conservative forces acting on the system) and the work done by all non-conservative forces, such as friction, control actions, contact forces, external disturbances.
	\label{note:work_fn}
}
$\workFn:  \confBundle \mapsto \R_+$, that is $\Lagrangian(\q, \dq) := \kinE(\q,\dq) - \workFn(\q, \dq)$ \citep{lanczos2020variational_principles_mechanics}.

In this context, a symmetry $\g$ is defined as an \highlight{energy preserving} transformation relating equivalent states in distinct regions of the phase space \citep{wieber2006holonomy}. This property is formalized as the \highlight{invariance} of the system's Lagrangian under the symmetry transformation, formally expressed as:

\begin{definition}[(Symmetric robotic system)]
	\label{def:symmetry_of_a_robotic_system}
	Consider a robot with generalized position coordinates $\q \in \confSpace \subseteq \R^{\nq}$, velocity coordinates $\dq \in \tangConfSpace \subseteq \R^{\nq}$, and Lagrangian $\Lagrangian: \confBundle \mapsto \R_+$. The system is deemed to possess a symmetry group $\G$ if its Lagrangian is $\G$-invariant. That is if:
	\begin{equation}
		\small
		\Lagrangian(\q, \dq) = \Lagrangian(\gq, \gdq) \quad \stforall \g \in \G, (\q,\dq) \in \confBundle.
		\label{eq:lagrangian_g_invariance}
	\end{equation}%
\end{definition}
\noindent
The symmetric state $(\gq, \gdq)$ is derived through a linear transformation given by $(\rep[\confSpace]{\g} \q, \rep[\confSpace]{\g} \dq)$. Here, $\rep[\confSpace]{\g}$ is the group representation in the configuration space, and $\rep[\confSpace]{\g} = \rep[\tangConfSpace]{\g}$, given that $\gdq := \rep[\tangConfSpace]{\g} \dq := \sfrac{d(\gq)}{dt} = \rep[\confSpace]{\g} \dq$.
States $(\q, \dq)$ and $(\gq, \gdq)$, related by a symmetry $\g$, as in \cref{eq:lagrangian_g_invariance}, will henceforth be denoted as \highlight{symmetric states}.
\subsubsection*{$\G$-equivariant equations of motion}

\begin{figure*}[t]
	\small
	\begin{equation}
		\begin{aligned}
			\frac{d}{dt}\frac{\partial\Lagrangian(\q,\dq)}{\partial\dq} - \frac{\partial\Lagrangian(\q,\dq)}{\partial\q}
			 & = \bm{0} =
			\frac{d}{dt}\frac{\partial\Lagrangian(\gq,\gdq)}{\partial(\gdq)} - \frac{\partial\Lagrangian(\gq,\gdq)}{\partial(\gq)}
			\\
			\frac{d}{dt}\frac{\partial\Lagrangian(\q,\dq)}{\partial\dq} - \frac{\partial\Lagrangian(\q,\dq)}{\partial\q}
			 & = \bm{0} =
			\left[
				\frac{d}{dt}\frac{\partial\Lagrangian(\gq,\gdq)}{\partial\dq}
				\right]
			\frac{\partial\dq}{\partial(\gdq)} -
			\left[
				\frac{\partial\Lagrangian(\gq,\gdq)}{\partial\q}
				\right] \frac{\partial\q}{\partial(\gq)}
			\\
			\frac{d}{dt}\frac{\partial\Lagrangian(\q,\dq)}{\partial\dq} - \frac{\partial\Lagrangian(\q,\dq)}{\partial\q}
			 & = \bm{0} =
			\left[
				\frac{d}{dt}\frac{\partial\Lagrangian(\gq,\gdq)}{\partial\dq}
				-
				\frac{\partial\Lagrangian(\gq,\gdq)}{\partial\q}
				\right] \rep[\confSpace]{\g}^{-1}   \quad
			\bigg| {\color{awesomeblue}
			\frac{\partial(\gdq)}{\partial\dq} = \frac{\partial(\gq)}{\partial\q} = \rep[\confSpace]{\g}
			}
			\\
			g \Glact
			\left[
				M(\q) \ddq - \genForces(\q,\dq)
				\right]
			 & = \bm{0} =
			M(\gq) \g \Glact \ddq - \genForces(\gq,\gdq) = \bm{0}
			\label{eq:eom_G_equivariance}
		\end{aligned}
	\end{equation}%
	\vspace*{-.8cm}
\end{figure*}

From a practical perspective, the importance of studying symmetry stems from the fact that these transformations relate the dynamics of a state $\state$ with that of the symmetric state $\isometryState$. This suggests that modeling and controlling our robotic system in the vicinity of a single state suffices to model and control the robot in the vicinity \textit{of all} symmetric states.

This geometric property is characterized analytically by the $\G$-equivariance of the system's \gls{eom}, obtained from deriving the \gls{eom} in the original and symmetry transformed coordinates, for any $\g \in \G$, as depicted in \cref{eq:eom_G_equivariance} and in \citet{wheeler2014covariance_eom}.

\noindent
The $\G$-invariant Lagrangian and the $\G$-equivariant \gls{eom} are equivalent statements that define the conditions under which a symmetry exists. Specifically, \cref{eq:eom_G_equivariance} defines a symmetry as a transformation that relates the instantaneous inertial and moving forces impacting symmetric states. This property provides a pathway for characterizing and identifying properties of symmetric systems.

\begin{proposition}[(Symmetric robots exhibit $\G$-equivariant mass matrix and control policies)]
	Reframing \cref{eq:eom_G_equivariance} as an equality constraint of inertial and moving forces between symmetric states, we deduce that for a symmetry to exist, both the generalized mass matrix $M$ and force field $\genForces$ need to be $\G$-equivariant, i.e.,

	\begin{equation}
		\small
		\begin{aligned}
			g \Gconj M(\q)              & = M(\gq),  \quad
			\text{and}                                            \\
			g \Glact \genForces(\q,\dq) & = \genForces(\gq,\gdq).
		\end{aligned}
		\label{eq:mass_and_forces_g_equivariance}
	\end{equation}%
	\vspace*{-1em}
	\label{prop:mass_and_forces_g_equivariance}
\end{proposition}
\noindent
The equivariance of $M$ implies that the inertial properties at the configuration $\gq$ are equivalent (up to a symmetry-induced change of basis) to those of the configuration $\q$ (refer to \cref{eq:conjugate_action}). Similarly, the equivariance of $\genForces$ requires that both the system's passive dynamics and control policy are $\G$-equivariant functions. As discussed in \cref{sec:applications_equiv_fn_approx}, optimal control policies for symmetric robotic systems are inherently equivariant.

To build an intuitive understanding, consider the Atlas robot shown in \cref{fig:atlas_dynamics_equivariance}a. Here, the symmetry $\g_{s}$, representing a reflection of space, relates two symmetric system states $(\q[0], \dq[0])$ and $(\g_{s} \Glact \q[0], \g_{s} \Glact \dq[0])$. As per \cref{eq:lagrangian_g_invariance}, this symmetry arises from their shared kinetic and potential energies, as well as the equal work done by instantaneous contact forces. Consequently, by \cref{eq:eom_G_equivariance}, the instantaneous dynamics of both states are linked by the symmetry transformation. This means that the system's motion trajectory originating from $(\q[0], \dq[0])$ will mirror the motion trajectory starting from $(\g_{s} \Glact \q[0], \g_{s} \Glact \dq[0])$, after applying the transformation $\g_{s}$. Moreover, by \cref{eq:mass_and_forces_g_equivariance}, the symmetric relationship between the motion trajectories is maintained as long as the forces acting on both systems, including contacts and control actions, remain related by the symmetry transformation. A similar analysis holds when considering the translation symmetry group $\G = \mathbb{T}_x$ for the Atlas robot (\cref{fig:atlas_dynamics_equivariance}b), where each $\g\in \mathbb{T}_x$ represents a horizontal translation of the robot and the environment.
\subsubsection*{Feasible and unfeasible symmetries}
In our analysis, we distinguish between \highlight{feasible} symmetries, which yield another reachable state within the constrained phase space $\isometryState \in \confBundle$, for any given state $\state$ (e.g., $\g\in \mathbb{T}_x$ in \cref{fig:atlas_dynamics_equivariance}b), and \highlight{infeasible} symmetries, which lead to states outside of the constrained phase space. Infeasible symmetries can occur when $\g$ involves a reflection of bodies (see \cref{fig:atlas_dynamics_equivariance}a), or when the resultant symmetric state violates some state constraint, such as joints position/velocity limits.

\subsubsection*{Floating base robotic systems} \label{sec:symmetries_floating_base}
In our analysis of locomoting and fixed-base robotic systems, we assume, without loss of generality, that our robot is a floating-base dynamical system evolving in $3$-dimensions. This allows us to decompose the state into $\q = (\pos_\base, \bm{\theta}_\base, \qj)$ and $\dq = (\vel_\base, \angvel[\base], \dqj)$, where $\pos_\base \in \R^3$ and $\bm{\theta} \in \R^3$ correspond to the base's position and orientation, respectively, and $\vel_\base \in \R^3$ and $\angvel[\base] \in \R^3$ represent the base's linear and angular velocities, respectively. Similarly, $\qj \in \confSpaceJS \subseteq \R^{\nj}$ describe the joint space configuration and velocity coordinates of the robot's $\nj=\nq-6$ internal \gls{dof} \citep{ostrowski1996geometric_perspectives}.

The $6$-dimensional (minimal coordinate) representation of $(\pos_\base, \bm{\theta}_\base)$ and $(\vel_\base, \angvel[\base])$ is advantageous for state interpretability and the derivation of the \gls{eom} (\cref{eq:eom_G_equivariance}). However, for computational efficiency, algebraic simplicity, singularities and the non-commutative nature of rotations in $3$-dimensions, rigid body configurations in space are represented in practice in homogenous matrix form (maximal coordinates). Specifically, the base's position configuration is represented by $\baseSE := \scalebox{0.9}{$\begin{bsmallmatrix}\mSO_\base & \pos_\base \\ \bm{0} & 1\end{bsmallmatrix}$} \in \SE$, where $\mSO_\base \in \SO[\dimEvolution]$ is the rotation matrix representing the base's orientation. Similarly, the base's velocity configuration is given by $\basese := \scalebox{0.9}{$\begin{bsmallmatrix} \mso[{\base}] & \vel_\base \\ \bm{0} & 1\end{bsmallmatrix}$} \in \se_\dimEvolution$, where $\mso[{\base}] \in \so_{\dimEvolution}$ is the skew symmetric matrix representation of the angular velocity $\angvel[\base]$ \citep{sola2021micro}. With a mild abuse of notation, we will hereafter represent the system's state as $\q = (\pos_\base, \bm{\theta}_\base, \qj) \simeq (\baseSE, \qj)$, and  $\dq = (\vel_\base, \angvel[\base], \dqj) \simeq (\basese, \dqj)$. This step is required to define the action of euclidean isometries on $(\pos_\base, \bm{\theta}_\base)$ and $(\vel_\base, \angvel[\base])$ in matrix form (see \cref{eq:group_representations_rigid_transformations}).

\begin{figure*}[t!]
	\centering
	\includegraphics[width=\linewidth]{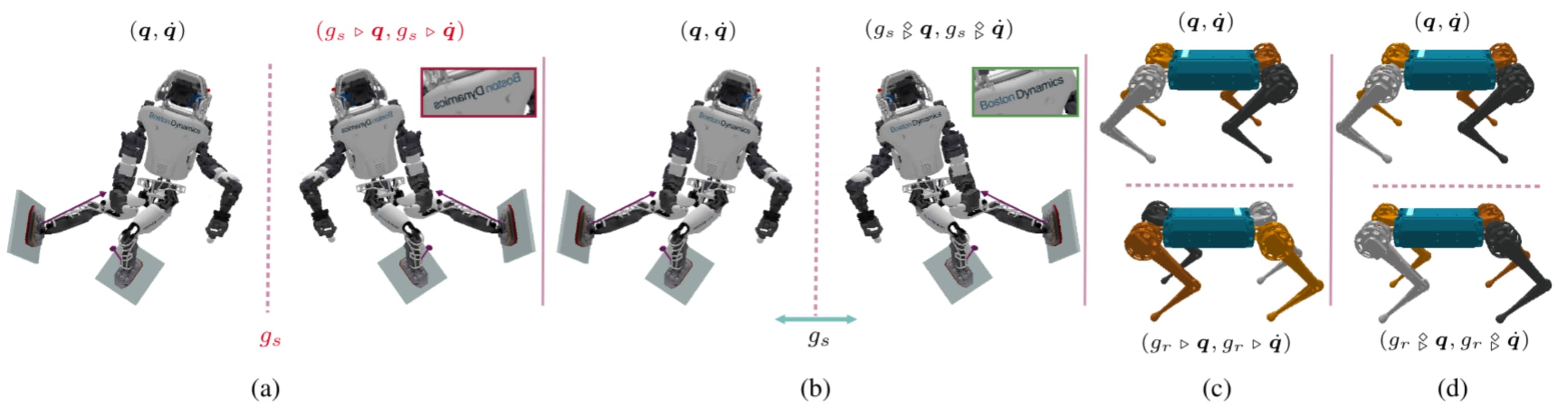}
	\vspace*{-1.8em}
	\caption{(a) Atlas robot undergoing an unfeasible horizontal spatial reflection $\g_s$, leading to an unreachable symmetric state $\isometryState[\g_s]$.
		(b) The same robot experiencing a feasible horizontal spatial reflection $\g_s$ through a morphological symmetry, resulting in a reachable equivalent state $\morphState[\g_s]$.
		(c) Mini Cheetah robot subjected to a feasible $180^\circ$ rotation $\g_r$, yielding the equivalent state $\isometryState[\g_r]$. (d) The same robot undergoing a feasible $180^\circ$ rotation $\g_r$ through a morphological symmetry, leading to the reachable equivalent state $\morphState[\g_r]$.
		\vspace*{-1.2em}
	}
	\label{fig:morphological_symmetries_example}
\end{figure*}

\section{Morphological Symmetries} \label{sec:morphological_symmetries}
Morphological symmetries are fundamentally associated with the capability of specific robotic systems to emulate Euclidean isometries—such as spatial rotations, reflections, or translations—through feasible state transformations. This implies that the robot can reach distinct state configurations, all dynamically equivalent to the state result of applying Euclidean isometries to the robot. A defining feature of these symmetries is that they are inherent properties of the robot, unaffected by time, specific motion tasks, or the operating environment. They stem from morphological or structural similarities resulting from replicated kinematic chains and body parts with symmetric mass distributions.

Before introducing a formal definition, let us consider the Atlas humanoid robot, which possesses the simplest morphological symmetry group, the reflection group $\G = \CyclicGroup[2]$. In \cref{fig:morphological_symmetries_example}a, the robot and its surrounding environment are subjected to an unfeasible reflection transformation, which involves the physically unattainable reflection of rigid bodies, leading to the unreachable symmetric state $\isometryState[\g_s]$. In contrast, \cref{fig:morphological_symmetries_example}b illustrates a scenario where the environment undergoes a similar reflection, yet the robot experiences a feasible state transformation. This transformation adjusts the robot's state to $\morphState[\g_s]$, a reachable and dynamically equivalent state to $\isometryState[\g_s]$. These distinct symmetry transformations of the robot's state represent two inequivalent representations (refer to \cref{sec:applications_equiv_fn_approx}) of the action of the reflectional symmetry $\g_s$, defined as:

\begin{equation}
	\begin{split}
		\isometryState[\g_s] &:=
		\left(\begin{bsmallmatrix}
				\mSE_{\g_s}\baseSE \\ \qj
			\end{bsmallmatrix}
		,
		\begin{bsmallmatrix}
				\mSE_{\g_s}\basese \mSE_{\g_s}^{\text{-}1} \\ \dqj
			\end{bsmallmatrix}
		\right)
		\quad
		\text{and}
		\\
		\morphState[\g_s] &:=
		\left(\begin{bsmallmatrix}
				\mSE_{\g_s}{\baseSE}\mSE_{\g_s}^{\text{-}1} \\ \rep[\confSpaceJS]{\g_s}\qj
			\end{bsmallmatrix}
		,
		\begin{bsmallmatrix}
				\mSE_{\g_s}\basese\mSE_{\g_s}^{\text{-}1} \\ \rep[\tangConfSpaceJS]{\g_s}\dqj
			\end{bsmallmatrix}
		\right).
	\end{split}
	\label{eq:def_morph_op}
\end{equation}%
\noindent
The former state transformation, resulting in $\isometryState[\g_s]$, describes the standard action of a reflection isometry, which involves reflecting the robot's rigid bodies while maintaining the joint space configuration unchanged (or invariant). In contrast, the action of a morphological symmetry, denoted by the operator $\morphOp$, describes a feasible state transformation that results in the reachable state $\morphState[\g_s]$. This transformation includes a reorientation of the base's body $\g \Gconj \baseSE := \mSE_{\g_s}{\baseSE}\mSE_{\g_s}^{\text{-}1} \in \SE[\dimEvolution]$ (refer to \cref{eq:conjugate_action}), coupled with a non-trivial transformation of the joint space (or internal) configuration $\g \Glact \qj := \rep[\confSpaceJS]{\g_s}\qj$.

A similar analysis applies to the Mini Cheetah quadruped robot, which exhibits a morphological symmetry group of order $|\G| = 8$. Focusing on the subgroup $\{e, \g_r\} < \G$, \cref{fig:morphological_symmetries_example}c illustrates the standard action of the isometry $\g_r$, depicting a feasible $180^\circ$ rotation of both the robot and its environment. This feasible transformation leads to the reachable symmetric state $\isometryState[\g_r]$. In contrast, \cref{fig:morphological_symmetries_example}d illustrates the morphological symmetry action of $\g_r$. This alternative transformation results in a distinct symmetric state $\morphState[\g_r]$ that preserves dynamic equivalence to $\isometryState[\g_r]$.

In both examples, the standard action of a Euclidean isometry $\g\in \G[E]
$ alters the robot \highlight{and} its environment, resulting in a system state $\isometryState$ that may or may not be reachable. In contrast, the action of a morphological symmetry defines a \highlight{feasible} robot's state transformation, resulting in a reachable state $\morphState$ that is dynamically equivalent to $\isometryState$. This transformation entails reorienting the robot's base and non-trivial transformation of its joint space configuration. These properties can be formalized for any robotic as follows:
\begin{definition}[(Morphological Symmetry)]
	\label{def:MS}
	Consider a robot with generalized position coordinates $\q \in \confSpace \subseteq \R^{\nq}$ and velocity coordinates $\dq \in \tangConfSpace \subseteq \R^{\nq}$. Let $\Lagrangian: \confBundle \mapsto \R$ denote the system's Lagrangian. An Euclidean isometry $\g \in \G[E]
	$ is deemed a system's morphological symmetry if the action has two inequivalent representations yielding distinct symmetric states $\isometryState$ and $\morphState$, being $\morphState$ a reachable state. That is if:
	\begin{equation}
		\small
		\begin{split}
			\Lagrangian\state
			=
			\Lagrangian\isometryState
			=
			\Lagrangian\morphState
			\quad
			\\
			\stforall
			(\q,\dq), \morphState \in \confBundle,
			\\
			\isometryState \neq \morphState.
		\end{split}
		\label{eq:ms_lagrangian_invariance}
	\end{equation}%
\end{definition}
\noindent
Hereafter, we denote the morphological symmetry group of a robotic system as $\G$. This group comprises all Euclidean isometries that satisfy \cref{def:MS},
that is
$\G \leq \G[E]
$. In a nutshell, the action of a morphological symmetry signifies an alternative pathway for applying specific rotations, reflections, and translations to the robot’s state.
\subsubsection*{Modeling the dynamics of symmetric systems}

A morphological symmetry $\g \in \G$ can be viewed as a point transformation in the robot's configuration space, linking any reachable state $\state$ to its set of symmetric states, featuring \textit{equivalent dynamics}, denoted as $\G\state = \{ \morphState[\g] \st \g \in \G\}$. This relationship is a crucial geometric prior when modeling the system's dynamics, as it requires any optimal analytical or data-driven model to be $\G$-equivariant \citep[Proposition 2]{ordonez2024dynamics}.

In fact, analytical models of rigid body dynamics for systems with morphological symmetries are inherently $\G$-equivariant due to their reliance on the analytical generalized mass matrix function (\cref{eq:mass_and_forces_g_equivariance}). For symmetric robots without analytical or tractable dynamics models, such as continuum, soft, or modular robots, morphological symmetries provide a valuable geometric prior, which in practice could improve the generalization of data-driven models and mitigate the challenges posed by the curse of dimensionality \citep{higgins2022symmetry_deepmind,biettisample}. For more details, refer to \cref{sec:applications}.
\subsubsection*{Control policy constraints}
By \cref{prop:mass_and_forces_g_equivariance}, a robotic system with a symmetry group $\G$ is required to possess a $\G$-equivariant control policy, as non-equivariant control forces break the $\G$-equivariance of the system's \gls{eom} (\cref{eq:eom_G_equivariance}). This property is of special value in the case of morphological symmetries, considering that, under mild assumptions, optimal control policies of symmetric systems are inherently $\G$-equivariant functions \citep{zinkevich2001symmetry_mdp_implications}, as detailed in \cref{sec:applications_reinforcement}.

In essence, this implies that for a bipedal system such as the Atlas robot (\cref{fig:morphological_symmetries_example}b), the optimal control policy is an \highlight{ambidextrous} policy. That is, the optimal action for a state $\state$ is equivalent to the control action for the symmetry-transformed state $\morphState$. This principle extends to systems with larger symmetry groups, such as the Mini Cheetah robot depicted in \cref{fig:teaser} (see animation \ref{animation:mini_cheetah-C2xC2xC2-symmetries_anim_static}).
\subsubsection*{Morphological constraints}
The presence of morphological symmetries in robotic systems, such as the Atlas robot (\cref{fig:morphological_symmetries_example}) and the Mini Cheetah robot (\cref{fig:teaser}), is closely linked to the $\G$-equivariance of the robot's generalized mass matrix (\cref{eq:mass_and_forces_g_equivariance}). This property arises from symmetries in mass distribution and the duplication of bodies and kinematic chains, consequence of \cref{prop:mass_and_forces_g_equivariance}. For example, the Atlas robot can mimic a spatial reflection due to its sagittal symmetry and duplicated limbs, while the Mini Cheetah, with its replicated legs and symmetric cuboid torso, can emulate three spatial reflections.

To understand these morphological constraints, we are required to delve into a specific type of a robot architecture. In the following section we focus on rigid body dynamics, however, note that the analysis presented can be easily adapted to continuum, soft, and modular robots.
\section{Morphological symmetries in rigid body systems} \label{sec:morphological_symmetries_rigid_body}
%

In this section, our objective is to characterize the implications of morphological symmetries in rigid body systems. These systems consist of $\nb$ interconnected rigid bodies evolving in $\EG$. Given that symmetries are transformations that preserve energy (\cref{def:symmetry_of_a_robotic_system}), we aim to discern the symmetry constraints on the kinematic and dynamics parameters. Concretely, we analyze the conditions of $\G$-invariance of the system's kinetic energy $\kinE\state=\kinE\morphState$.

\subsection{Rigid body systems}
The kinetic energy in rigid body dynamics is calculated by summing the energies of all constituent bodies in the system: $\kinE(\vel,\angvel) = \frac{1}{2}\sum_{k}^{\nb} \mass[k]\vel_k^\transpose\vel_k  + \angvel[k]^\transpose \Inertia[k] \angvel[k]$, where $\mass[k] \in \R_+$, $\Inertia[k] \in \R^{\dimEvolution \times \dimEvolution}$, $\vel_k\in \R^{\dimEvolution}$ and $\angvel[k] \in \R^{\dimEvolution}$ represent the mass, rotational inertia at the \gls{com}, linear and velocity of body $k$, respectively. Although this quantity can also be expressed in terms of the system's generalized coordinates and generalized mass matrix $\kinE(\q,\dq) = \frac{1}{2} \dq^{\transpose}\Mass(\q)\dq$, we exploit the decomposition of the state into its floating-base and joint space configurations $\q = (\baseSE, \qj)$. This is done to study the invariance of the kinetic energy of the floating-base body $\kinE_\base$ independently from that of the joint space kinetic energy $\kinE_\confSpaceJS$. These quantities are expressed as:

\begin{equation}
	\small
	\begin{split}
		&\kinE_\base(\baseSE)
		=
		\frac{1}{2} \mass[\base]\vel_\base^\transpose\vel_\base  + \angvel[\base]^\transpose \Inertia[\base] \angvel[\base]
		\quad
		\text{and}\\
		&\kinE_{\confSpaceJS}(\qj,\dqj)
		=
		\frac{1}{2} \dqj^{\transpose}\overline{\Mass}(\qj)\dqj,
	\end{split}
	\label{eq:kinetic_energy_base_joint_space}
\end{equation}%
where
\begin{equation}
	\medmath{
		\overline{\Mass}(\qj)
		{:=}\!\!\!
		\sum_{k}^{\nb-1}\!\! \PosJacob[k]\!(\qj)^\transpose \mass[k] \PosJacob[k]\!(\qj) {+} \OriJacob[k]\!(\qj)^\transpose \Inertia[k]\OriJacob[k]\!(\qj)
	}
	\label{eq:kinetic_energy_base_joint_space_mass_matrix}
\end{equation}%
denotes the joint space generalized mass matrix, constructed from the mass, inertia, and the position and orientation Jacobians $\PosJacob[k]: \confSpace \rightarrow \R^{\dimEvolution \cross \nj}$ and $\OriJacob[k]: \confSpace \rightarrow \R^{\dimEvolution \cross \nj}$ of each body. These Jacobians define the state-dependent mappings from generalized velocities to the k$^{th}$ body's linear $\left(^{\smallbase}\vel_k = \PosJacob[k](\qj)\dqj\right)$ and angular $\left(^{\smallbase}\angvel[k] = \OriJacob[k](\qj)\dqj\right)$ velocity in Euclidean space, relative to the floating-base body's frame \citep{wieber2006holonomy}.

The computation of these Jacobians relies on the system's kinematic parameters, which detail the relative positions and orientations of links and joints, as well as the system's dynamic parameters, which include the mass and inertia of all bodies. Since the kinetic energy is dependent on these parameters, the presence of a morphological symmetry inherently imposes constraints on the kinematic and dynamic parameter space.

\subsection{Constraints on the floating-base's body mass distribution}\label{sec:constraints_on_the_floating}
Consider the conditions under which a morphological symmetry exists for the floating-base body only. In this case, the necessary Lagrangian $\G$-invariance from \cref{eq:ms_lagrangian_invariance} reduces to the equality of kinetic energy between the base body transformed by a Euclidean isometry, denoted as $\g \Glact \baseSE$, and the body transformed by morphological symmetry, represented as $\g \Gconj \baseSE$. This can be expressed as:

\begin{equation}
	\small
	\begin{split}
		\kinE_\base(\g \Glact \baseSE)=\kinE_\base(\g\Gconj\baseSE)
	\end{split}
	\label{eq:base_kinematic_energy_constraint},
\end{equation}%
where the action of the Euclidean isometry represents a potential rotation or rotoreflection of the body $\mR_\g \mR_{\base} \in \OG[\dimEvolution]$, and a morphological symmetry is restricted to a rotation $\mR_\g \mR_{\base}\mR_{\g}^\transpose \in \SO[\dimEvolution]$, due to the feasibility requirement.

\begin{figure}[t!]
	\centering
	\includegraphics[width=\linewidth]{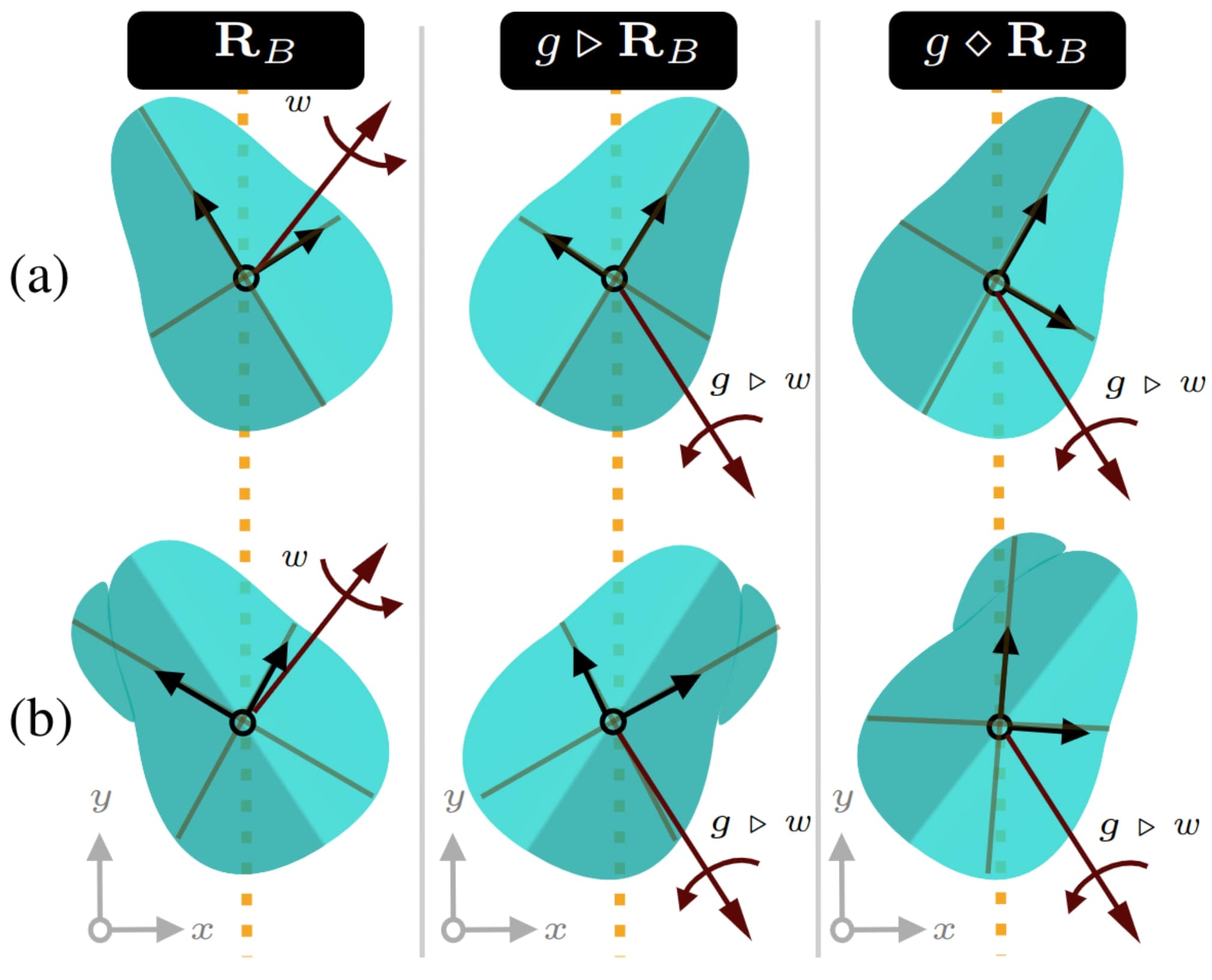}
	\vspace*{-1em}
	\caption{
		(a) Reflectional morphological symmetry is observed in a body with a symmetric mass distribution. For such a body, the unfeasible action of a reflection of the body, $\g \Glact \mR_{\base}$, results in a body configuration that mirrors the inertia and kinetic energy of the  rotated body at $\g \Gconj \mR_{\base}$. (b) However, if the symmetry in mass distribution is disrupted, the states $\g \Glact \mR_{\base}$ and $\g \Gconj \mR_{\base}$ will exhibit different reflected inertias and kinetic energies when subjected to the equal angular velocities. This results in the body's inability to mimic a reflection of space.
		\vspace*{-1em}
	}
	\label{fig:rigid_body_inertia_constraints}
\end{figure}

Since the configurations $\g \Glact \baseSE$ and $\g \Gconj \baseSE$ generally differ, they are constrained to have the same mass distribution to ensure an equivalent dynamics. In other words, this is required to react identically to the application of a moving force (refer to \cref{fig:rigid_body_inertia_constraints}). This requirement is translated to the equality of the reflected inertia tensors:

\begin{equation}
	\small
	\begin{split}
		(\g \Glact \mR_\base)\bar{\mI}_{\base}(\g \Glact \mR_\base)^{\text{-}1}
		&=
		(\g \Gconj \mR_\base) \bar{\mI}_{\base}(\g \Gconj \mR_\base)^{\text{-}1}
		\\
		(\mR_{\g} \mR_\base) \bar{\mI}_{\base} (\mR_\base^\transpose\mR_{\g}^\transpose)
		&=
		(\mR_{\g} \mR_\base \mR_{\g}^\transpose) \bar{\mI}_{\base} (\mR_{\g} \mR_\base^{\transpose}\mR_{\g}^{\transpose}),
	\end{split}
	\label{eq:inertia_symmetry_constraints}
\end{equation}%
where $\Inertia[\base]=\mR_\base \bar{\mI}_{\base} \mR_\base^\transpose \in \R^{\dimEvolution\times\dimEvolution}$ is the reflected body inertia at the configuration $\mR_\base$, and $\bar{\mI}_{\base}$ is the body's diagonal inertia tensor in a frame aligned with the body's principal axes of inertia~\citep{traversaro2016eigenvalue}, which we show in~\cref{fig:rigid_body_inertia_constraints}a.

The ability of a body to adopt various configurations, each sharing identical reflected inertias as defined by~\cref{eq:inertia_symmetry_constraints}, is an inherent characteristic of bodies with symmetrical mass distributions. This property introduces redundancy in the selection of orientation and handedness of a reference frame attached to the body's \gls{com} and aligned with the body's principal axes of inertia. Consequently, this implies the existence of a group of symmetry transformations $\G[B] \leq \G[E]
$ that can alter the body's configuration while preserving the reflected inertia tensor:

\begin{equation}
	\small
	\G[B] = \{g \in
	\G[E]
	\st \mR_\base \bar{\mI}_{\base} \mR_\base^{\transpose} = \mR_\base (\g^{\text{-}1}\Gconj\bar{\mI}_{\base}) \mR_\base^\transpose, \;\forall\; \mR_\base \in \SO[\dimEvolution] \},
	\label{eq:candidates_morphological_symmetries}
\end{equation}%
\noindent
in which the symmetry $\g\in \G[B]$ represents a transformation relative to the body's principal axes of inertia, i.e., $\mR_\base \mR_{\g}$. \Cref{fig:rigid_body_inertia_constraints}a depicts an example of the reflection along one of the bodies principal axes of inertia relating the configurations.
\subsubsection{Identifying candidate morphological symmetries}
\label{sec:identifying_candidate_morphological_symmetries}
The group of symmetries in the mass distribution of the robot's base $\G[B]$ represents the set of in-place rotations and reflections under which the dynamics of this body remain invariant. These are interpreted as the subgroup of symmetries of Newtonian physics preserved by the floating-base's body mass distribution. Therefore, this group represents a set of potential morphological symmetries of the robot, as depicted in~\cref{fig:G_C3}a.

It is important to observe that both resulting body orientations, $\g \Glact \mR_\base = \mR_{\g} \mR_\base \in \OG$ and $\g \Gconj \mR_\base = \mR_{\g} \mR_\base \mR_{\g}^\transpose \in \SO$, share the same reflected inertias. Consequently, when subjected to identical velocities and forces, these two base configurations have equivalent dynamics, as expressed in~\cref{eq:base_kinematic_energy_constraint}. Furthermore, $\g \Gconj \mR_\base$ represents a feasible body configuration even when $\g$ corresponds to an unfeasible isometry, such as a reflection (see \cref{fig:rigid_body_inertia_constraints}).

To understand this fact, let's consider the two bodies illustrated in \cref{fig:rigid_body_inertia_constraints}. In~\cref{fig:rigid_body_inertia_constraints}a, this body exhibits a symmetric mass distribution with respect to its $x$ principal axis of inertia, leading to $\G[B] = \{e, \g \}\simeq \CyclicGroup[2]$. In contrast, in~\cref{fig:rigid_body_inertia_constraints}b, this body lacks any symmetry in its mass distribution, resulting in $\G[B] = \{e\}$. The symmetric mass distribution of body in \cref{fig:rigid_body_inertia_constraints}a enables it to emulate an otherwise unfeasible spatial reflection, $\g \Glact \mR_\base$, through a feasible rotation $\g \Gconj \mR_\base$. It is crucial to note that perturbing the mass distribution disrupts this reflectional symmetry, as evidenced by the differing kinetic energy of the $\g$-transformed body and $\g$-transformed body differ (refer to \cref{fig:rigid_body_inertia_constraints}b).

A practical example can be observed in the Mini Cheetah robot's base body depicted in \cref{fig:teaser}. Its cuboid body showcases three orthogonal reflectional symmetries in the Mini Cheetah's mass distribution, denoted as $\G[B] = \{e, \g_s,\g_t, \g_f \st \g_s^2=\g_t^2=\g_f^2=e\}$. These symmetries result in an arbitrary selection of the reference frame attached to its body, leading to the arbitrary determination of the Mini Cheetah's forward/backward, up/down, and left/right directions.

In both instances, the group $\G[B] \leq
	\G[E]
$, which describes the symmetries in the robot's base mass distribution, suggests potential Euclidean isometries that could result in morphological symmetries of the robotic system. For robotic systems with multiple unique (non-replicated) bodies (e.g., the head of the Atlas robot, see \cref{fig:atlas_dynamics_equivariance}), the group of candidate morphological symmetries is limited to the subgroup of Euclidean isometries that describe the symmetries of mass distribution of all unique bodies. Whether a candidate symmetry is indeed a morphological symmetry depends on (i) the other bodies in the system's kinematic structure and (ii) the robot's morphology admits a joint space transformation that complies with \cref{def:MS}.
\subsection{Constraints on the kinematic structure}
\label{sec:constraints_on_the_kinematic_structure}
%
\begin{figure*}[t!]
	\centering
	\includegraphics[width=\linewidth]{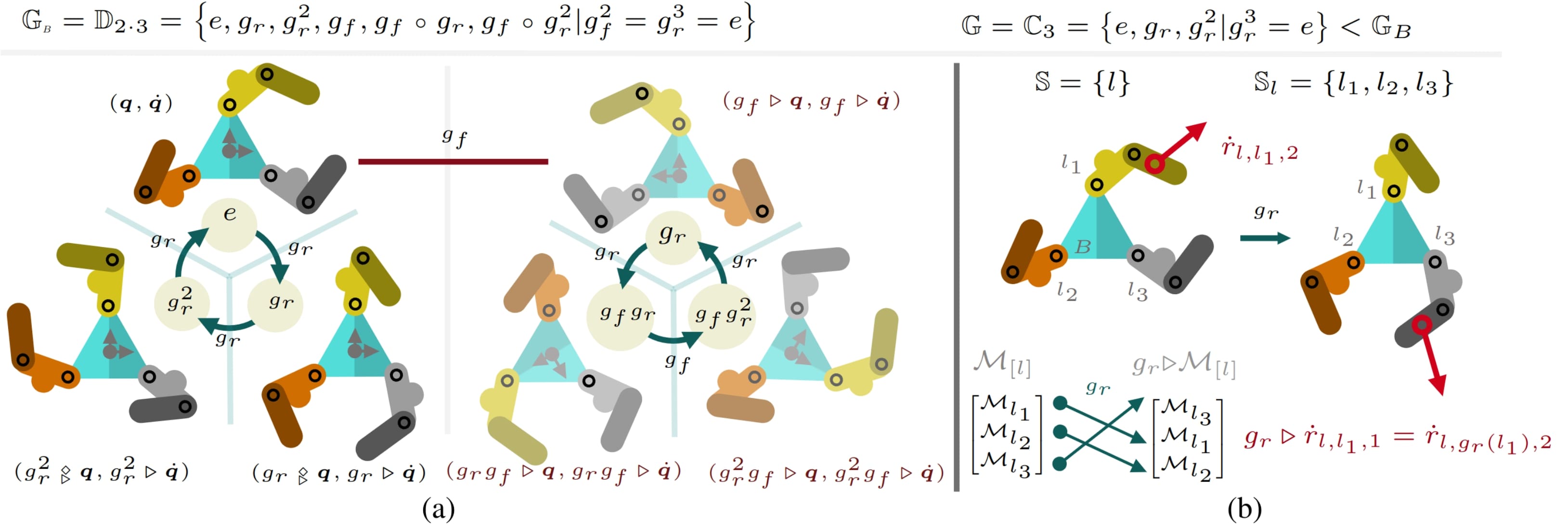}
	\vspace*{-1.5em}
	\caption{
		\small
		(a) Example robotic system, evolving in $\R^2$, featuring a triangular base body which exhibits a symmetric mass distribution characterized by the group of $3$ rotations and $3$ rotoreflections, the  Dihedral group $\G[B] = \DihedralGroup[2\cdot3]$. However, the system's morphological symmetry is limited to the rotational subgroup, $\G = \CyclicGroup[3] < \G[B]$, due to the absence of reflectional symmetry in the limbs' bodies mass distribution, which is necessary for mimicking the reflection isometry $\g_f$ with a feasible joint space transformation. (b) The base body's symmetric mass distribution results in an arbitrary labeling of the robot's limbs ($l_1,l_2$ and $l_3$), enabling the robot to mimic a $120^\circ$ spatial rotation through a joint space transformation that permutes the limbs' configurations.
		\label{fig:G_C3}
		\vspace*{-1em}
	}
\end{figure*}

Modularity in the kinematic structure, specifically the replication of identical or reflected kinematic branches (or substructures), provides the necessary conditions for the existence of joint space action transformations.
In these cases, we consider Euclidean isometries $\g \in \G[B]$ as potential morphological symmetries of the robot. Concretely, given that the configuration of the floating-base body transformed by this isometry differs from the one transformed by the morphological symmetry, (i.e., $\g \Glact \baseSE \neq\g \Gconj \baseSE$ as illustrated in \cref{fig:morphological_symmetries_example,fig:G_C3}), there must exist a joint space action transformation
\begin{equation*}
	(\gqj,\gdqj) = (\rep[\confSpaceJS]{\g} \qj, \rep[\tangConfSpaceJS]{\g} \qj),
\end{equation*}
which ensures that both robot configurations exhibit equal energy states and equivalent dynamics (\cref{eq:ms_lagrangian_invariance}).
\subsubsection*{Modular kinematic structure}
Robotic systems often exhibit a balanced distribution of replicated kinematic branches within their kinematic structure. Examples include the four identical legs of the Mini Cheetah robot and the mirrored arms and legs of the Atlas robot (refer to \cref{fig:teaser} and \cref{fig:atlas_dynamics_equivariance}). This replication introduces symmetries in the labeling (or ordering) of these branches, which in practice enable the interchange of role and configuration of them under the action of a morphological symmetry, to mimic the Euclidean isometry.

As an example, consider the robot in \cref{fig:G_C3}. The symmetric mass distribution of this robot's base body, combined with its replicated limbs, results in arbitrary ordering of its three limbs. This symmetry allows the robot to replicate a $120^\circ$ spatial rotation through a joint space transformation that involves a permutation of its limb configurations and ensures the equivalence of the robot's energy under the action of the morphological symmetry. Similarly, \cref{fig:teaser} shows how the morphological symmetries of the Mini Cheetah robot result in the arbitrary labeling of the \highlight{front/hind} and \highlight{left/right} legs.
\subsubsection{Structure of the joint space action transformation}
\label{sec:structure_of_the_joint_space_action_transformation}
Assume the robot has $\nuch$ unique kinematic branches, represented by the label set $\setKinStructLabels = \{s_1, \cdots, s_\nuch\}$. Each branch $s_i$ possesses $\nchdof(s_i)\in \N$ \gls{dof} and is replicated $\nch(s_i)\in \N$ times within the robot's kinematic structure. The labels for the instances of each branch $s_i$ are denoted as $\setKinStructLabels_i=\{s_{i,1}, \cdots, s_{i,\nch(s_i)}\}$ (see \cref{fig:G_C3}b). As an example, consider robot Atlas, featuring as unique kinematic chains the leg, arm, and head $\setKinStructLabels = \{s_{arm}, s_{leg}, s_{head}\}$. These substructures are replicated $\nch(s_{arm})=\nch(s_{leg}) = 2$ and $\nch(s_{head})=1$ times.

The action of a morphological symmetry in the joint space results in a permutation of the roles of branches with the same type, denoted as $g \Glact s_{i,j}:= s_{i,\g(j)} \in \setKinStructLabels_i$, where $\g(j)$ is the label that $j$ is mapped to under the permutation induced by $\g$. This leads to the decomposition of the joint space configuration space and its associated group representation:

\begin{equation}
	\small
	\begin{split}
		g \Glact s_{i,j} &:= s_{i,\g(j)} \in \setKinStructLabels_i,
		\quad
		\text{and}
		\\
		\rep[\setKinStructLabels_{i}]{\g}
		\begin{bsmallmatrix}
			s_{i,1} \\ s_{i,2} \\ \vdots
		\end{bsmallmatrix}
		& =
		\begin{bsmallmatrix}
			s_{i,\g(1)} \\ s_{i,\g(2)} \\ \vdots
		\end{bsmallmatrix}
		\; \stforall i \in [1,\nuch], j\in[1,\nch(s_i)],
	\end{split}
	\label{eq:permutation_per_substructures}
\end{equation}%
where $\rep[\setKinStructLabels_{s_i}]{\g}$ is the permutation representation acting on the labels of the instances of branch type $s_i$. Following our example with the Atlas robot, the action of  $\g_s$ in \cref{fig:morphological_symmetries_example}b induces a permutation of the left and right arm configurations $\g \Glact s_{arm,1} = s_{arm,2}$ and $\g \Glact s_{arm,2} = s_{arm,1}$.
Given that these permutations do not mix the distinct branch types, we can adopt a basis for the joint space configuration space leading to the decomposition of its associated group representation, i.e.,

\begin{equation}
	\small
	\begin{split}
		\confSpaceJS &:= \confSpaceJS_{[s_{1}]} \times \ldots \times \confSpaceJS_{[\nuch]} \subseteq \R^{\nj},
		\quad \text{and} \\
		\rep[\confSpaceJS]{} &:= \rep[\confSpaceJS_{[s_1]}]{} \oplus \cdots \oplus \rep[\confSpaceJS_{[\nuch]}]{}.
	\end{split}
	\label{eq:joint_space_configuration_decomposition}
\end{equation}%
Each $\confSpaceJS_{[s_i]} := \bigotimes_{j=1}^{\nch(s_i)}\confSpaceJS_{s_i} \subseteq \R^{\nch(s_i)\nchdof(s_i)}$ encapsulates the configuration space of all instances of the kinematic branch type $s_i$ (see \cref{fig:G_C3}b). Conversely, $\confSpaceJS_{s_i} \subseteq \R^{\nchdof(s_i)}$ represents the configuration space of a single instance of type $s_i$. This suggests that the joint space group representation $\rep[\confSpaceJS]{}$ is exclusively constructed from the $\nuch$ representations of each branch type, i.e.,  $[\rep[\confSpaceJS_{s_i}]{}]_{i=1}^{\nuch}$.

Intuitively, the transformation described by $\rep[\confSpaceJS_{s_i}]{}$ implies an independent rotation/reflection of the coordinate frames attached to each joint within the kinematic branch $s_i$. For instance, when the branch $s_i$ is build from $1$-dimensional prismatic or revolute joints, its group representation $\rep[\confSpaceJS_{s_i}]{}$ results in diagonal matrices filled with 1 and -1, describing required reflections on the axes of each \gls{dof}.

Equipped with the above-mentioned formalism, the joint space group representation is defined as:

\begin{equation}
	\small
	\rep[\confSpaceJS]{} :=
	\begin{bsmallmatrix}
		\rep[\confSpaceJS_{[s_1]}]{} & & \\
		& \ddots & \\
		& & \rep[\confSpaceJS_{[\nuch]}]{}
	\end{bsmallmatrix},
	\quad \text{with} \quad
	\rep[\confSpaceJS_{[s_i]}]{} := \rep[\setKinStructLabels_{i}]{} \otimes \rep[\confSpaceJS_{s_i}]{}
	\label{eq:joint_space_configuration_definition}
\end{equation}%
Where $\otimes$ denotes the Kronecker product. This matrix product, in conjunction with the permutation of the branch's configurations $\rep[\setKinStructLabels_{s_i}]{}$, is responsible for ensuring that the group representation $\rep[\confSpaceJS_{[s_i]}]{}$ applies the appropriate rotation/reflection of the joints coordinate frames across all branches of type $s_i$,  as depicted in \cref{fig:G_C3}.
\subsubsection{Algebraic constraints in kinematic parameters}
The approach delineated above provides insights into the internal structure of the joint space group representation, emphasizing the importance of the unique constituent group representations $\rep[\confSpaceJS_{s_i}]{}$ necessary to define the group action in $\confSpace$. Algebraically, the validity of the action of each candidate $\g \in \G[B]$ can be verified by checking whether the transformation yields a kinematic structural symmetry—a prerequisite for the existence of a morphological symmetry. This requirement manifests as an equality between the linear velocity of each body, transformed by the group action $\g \Glact \vel_{s_{i,j},n}$, and the velocity of its permutation counterpart $\vel_{s_{i,\g(j)},n}$ (see \cref{fig:G_C3}b). Here, $\vel_{s_{i,\g(j)},n}$ denotes the linear velocity of body $n$ in the kinematic branch $s_{i,\g(j)}$. This constraint is articulated, for any $n \in [1,\nchdof(s_i)]$, as:

\begin{equation}
	\begin{split}
		\small
		\vel_{s_{i,\g(j)},n}
		& =
		g \Glact \vel_{s_{i,j},n}, 
		\\
		\PosJacob[s_i,\g(j),n](\g \Glact \qj) \g \Glact \dqj
		& =
		g \Glact \PosJacob[s_{i,j},n](\qj),
		\\
		\PosJacob[s_i,\g(j),n](\rep[\confSpaceJS]{\g} \qj) \rep[\tangConfSpaceJS]{\g}
		& =
		\mR_{\g} \PosJacob[s_{i,j},n](\qj).
		\label{eq:kinematic_symmetries_rigid_bodies_b}
	\end{split}
\end{equation}%
It is worth emphasizing that \cref{eq:kinematic_symmetries_rigid_bodies_b} describes the constraints in the kinematic parameters of the robot's kinematic branches. An analog constraint is applied when considering the angular velocities of the bodies.
\subsection{Algorithmic identification of a system's morphological symmetry group}
\label{sec:algorithmic_id_ms}
\definecolor{algorithmcolor}{RGB}{245, 255, 255} 

\begin{figure}
	\centering
	\begin{minipage}{\linewidth}
		\begin{tcolorbox}[colback=algorithmcolor, arc=4mm, boxrule=0pt,frame empty, left=1mm, right=3mm]
			\begin{algorithmic}[1]
				\small 
				\Statex \textbf{\sffamily Identifying Morphological Symmetries}
				\State \sffamily Identify unique bodies in the kinematic structure.
				\State \sffamily Identify the group $\G[B]$ of candidate MS, from the unique bodies (\cref{eq:inertia_symmetry_constraints}).
				\State \sffamily Identify the unique kinematic structures $\setKinStructLabels = \{s_1, \cdots, s_\nuch\}$. Their configurations spaces $\confSpaceJS_{s_i}$, and group representations $\rep[\confSpaceJS_{s_i}]{}$ (\cref{eq:joint_space_configuration_decomposition}).
				\State \sffamily Identify the instances of the kinematic branches $\setKinStructLabels_i=\{s_{i,1}, \cdots, s_{i,\nch(s_i)}\}$ and their associated permutations $\rep[\setKinStructLabels_{s_i}]{}$ (\cref{eq:permutation_per_substructures}).
				\State \sffamily Build the group representation $\rep[\confSpaceJS]{}$ (\cref{eq:joint_space_configuration_definition}).
				\State \sffamily Test each $\g\in \G[B]$ following \cref{eq:kinematic_symmetries_rigid_bodies_b} and \cref{eq:kinetic_energy_base_joint_space}.
			\end{algorithmic}
		\end{tcolorbox}
	\end{minipage}
	\caption{Pseudo-code for identifying the morphological symmetry group $\G$ in rigid body dynamics.}
	\label{fig:pseudocode_identifying_morphological_symmetry_group}
\end{figure}

In previous sub-sections, we outline the necessary conditions for the existence of morphological symmetry in rigid body dynamics. These include specific constraints on the robot's kinematic structure and on the mass distribution within rigid bodies. While this analysis may seem extensive for simpler systems or symmetry groups of low order, it provides a fundamental framework for studying more complex systems typically encountered in robotics. Such systems may have a larger number of \gls{dof} or symmetries, where geometric intuition alone may not be adequate. The pseudo-code in \cref{fig:pseudocode_identifying_morphological_symmetry_group} provides a step-by-step guide for identifying the morphological symmetry group. In the following section we motivate our theoretical analysis by demonstrating the practical benefits of identifying the morphological symmetry group in the context of dynamic motion analysis.

\subsection{Dynamics' harmonic analysis} \label{sec:joint_space_basis_harmonic_motions}
In this section, we exploit the permutation symmetries in the robot's joint space configuration $\confSpaceJS$ using abstract harmonic analysis. Our objective is to elucidate the numerical and modeling benefits of the \highlight{isotypic basis}, an alternative basis set for $\confSpaceJS$ in which the system's dynamics are decomposed as a superposition of orthogonal, lower-dimensional dynamics of symmetry-constrained robot configurations.

Similarly to how the dynamics of a particle in $3$D space are modeled by its dynamics in the $x,y,z \sim \R$ orthogonal subspaces of $\R^3$, the isotypic basis of a robot's joint space configuration space, $\confSpaceJS$, exposes orthogonal subspaces of joint space configurations. This enables the description of any motion trajectory as a superposition of lower-dimensional trajectories per subspace, each governed by its own independent dynamics. We will refer to these orthogonal configuration subspaces as \textit{isotypic subspaces} and as  the robot's \glspl{ncm} of motion (inspired by the analogous term "normal vibrational modes" in molecular dynamics \cite{dresselhaus2007group_theory_applications_to_physics_of_condensed_matter}). The qualifier ”normal” underscores the orthogonality between the directions of motion associated with each \gls{ncm}.
\begin{figure*}[t!]
	\centering
	\includegraphics[width=\linewidth]{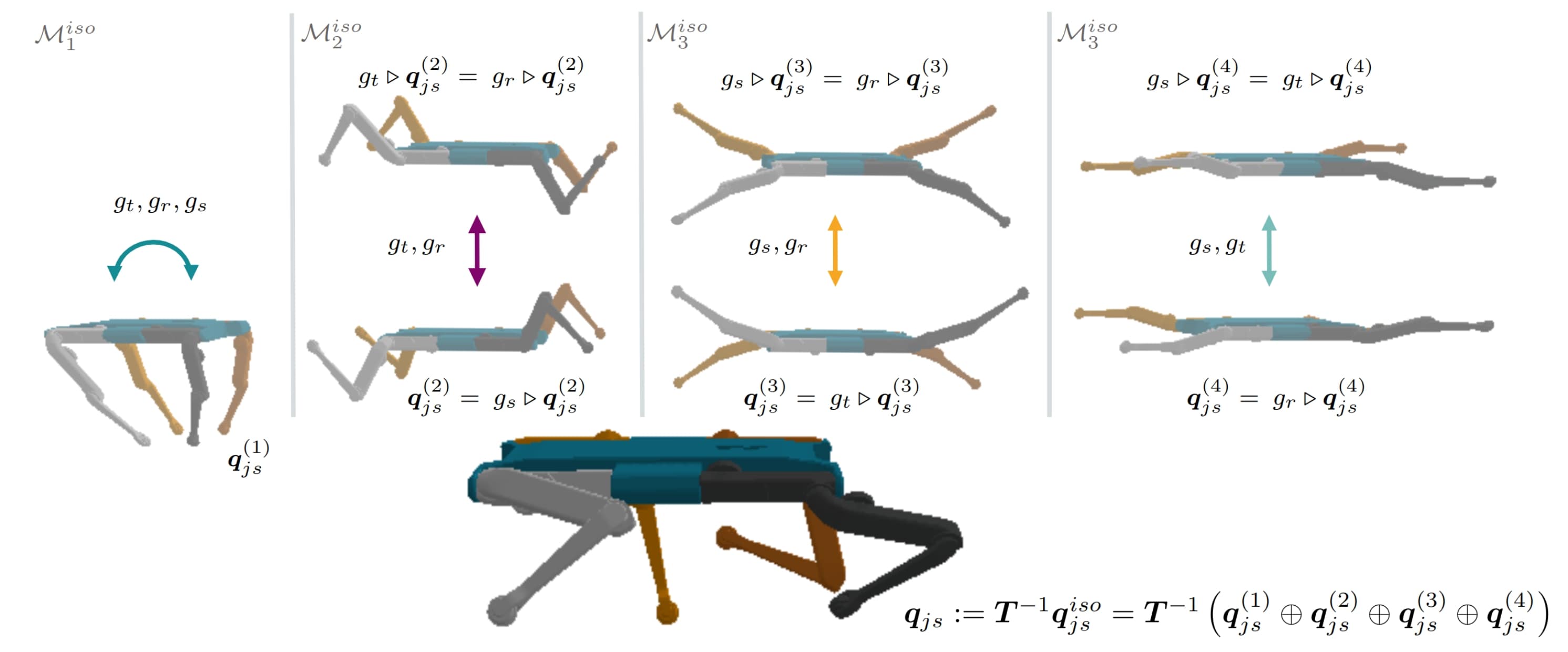}
	\vspace*{-2.0em}
	\caption{
		\small
		Isotypic decomposition of the joint space configuration for the Solo quadruped robot \citep{grimminger2020solo_robot}, considering only the morphological symmetry subgroup $\KleinFourGroup < \G := \KleinFourGroup \times \CyclicGroup[2]$ of order $|\KleinFourGroup|=4$ (see \cref{fig:data_aug_equiv_fn_approx}a). After applying the change of basis $\mT$, the robot's $12$-dimensional joint space configuration space $\confSpaceJS$ is decomposed into four orthogonal isotypic subspaces $[\confSpaceJS^{\iso}_\isoCompIdx]_{\isoCompIdx=1}^{4}$, each of dimension $3$. Consequently, any system configuration $\qj$, or any velocity, acceleration, or force vector in $\tangConfSpaceJS$, can be orthogonally decomposed into projections within each isotypic subspace.
		These subspaces are constrained to feature a unique subset of symmetry transformations from the group $\KleinFourGroup$. Thus, they can be interpreted as subspaces characterizing a range of symmetry-constrained robot's configurations/motions, refereed to as a \acrlong*{ncm} (\gls{ncm}) of motion.
		For instance, the subspace $\confSpaceJS^{\iso}_1$, associated with the symmetries of the trivial subgroup $\{e\}< \KleinFourGroup$, represents the \gls{ncm} of synchronized leg motions, where all legs share the same configuration.
		The remaining isotypic subspaces, $\confSpaceJS^{\iso}_2$, $\confSpaceJS^{\iso}_3$, and $\confSpaceJS^{\iso}_4$, each constrained by symmetry subgroups isomorphic to a reflection group $\CyclicGroup[2]$, represent distinct patterns of synergistic motions. Specifically, $\confSpaceJS^{\iso}_{2}$ corresponds to synchronized left and right legs, $\confSpaceJS^{\iso}_{3}$ to synchronized front and back legs, and $\confSpaceJS^{\iso}_{4}$ to synchronized diagonally opposed legs.
		\vspace{-1em}
		\label{fig:solo_decomposition_harmonic_states}
	}
\end{figure*}

%
\subsubsection*{Block-diagonal structure of the group representation of the joint space configuration space}
%

Recall that a robot with a morphological symmetry group $\G$ posses a symmetric joint space configuration space $\confSpaceJS$ with group representation $\rep[\confSpaceJS]{}: \G \mapsto \GLGroup(\R^{\nj})$. Symmetric vector spaces can be decomposed into a direct sum of orthogonal subspaces, $\confSpaceJS := \confSpaceJS^{\iso}_{1} \oplus^\perp  \cdots \oplus^\perp \confSpaceJS^{\iso}_{\isoCompNum}$, denoted as isotypic subspaces (\cref{eq:isotypic_decomposition}). Each subspace has the charateristic of being invariant under the group action. I.e., $\g \Glact \qj \in \confSpaceJS^{\iso}_{\isoCompIdx}$ for all $\g \in \G$ and $\qj \in \confSpaceJS^{\iso}_{\isoCompIdx}$ (see \cref{fig:solo_decomposition_harmonic_states}).

To expose the orthogonal decomposition, we identify a change of basis $\mT: \confSpaceJS \mapsto \confSpaceJS$ transitioning from a cannonical basis to the isotypic basis, in which $\rep[\confSpaceJS]{}$ decomposes into a direct sum of the $\isoCompNum$ representations (\cref{eq:direct_sum_representations}), each  acting on a specific isotypic subspace:
\begin{equation}
	\small
	\begin{split}
		\rho^{\iso}_{\confSpaceJS}(\g)
		&:=
		\rep[\confSpaceJS^{\iso}_{1}]{} \oplus \cdots \oplus \rep[\confSpaceJS^{\iso}_{\isoCompNum}]{}
		=
		\mT
		\left[
			\rep[\confSpaceJS]{\g}
			\right]
		\mT^{\transpose}.
	\end{split}
	\label{eq:decomp_joint_space_rep}
\end{equation}%
\noindent
Where $\rho^{\iso}_{\confSpaceJS}(\g)$ denotes the group representation in the isotypic basis of $\confSpaceJS$, and each isotypic subspace $\confSpaceJS^{\iso}_{\isoCompIdx}$ describes a space of symmetry-constrained robot configurations (a \gls{ncm}), as depicted in \cref{fig:solo_decomposition_harmonic_states}. The number of subspaces $\isoCompNum$ is determined by the symmetry group $\G$, while each subspace is constrained to feature a distinct subset of the symmetries of $\G$, associated with one of the unique irreducible representations $\irrep[\isoCompIdx]{}$ of the group $\G$ (see \cref{sec:background}). This is because each $\rep[\confSpaceJS^{\iso}_{\isoCompIdx}]{}$ is composed of $\irrepMultiplicity[\isoCompIdx]$ copies of a unique irreducible representation $\irrep[\isoCompIdx]{}$ of the group, i.e., $\rep[\confSpaceJS^{\iso}_{\isoCompIdx}]{} \sim \Oplus_{\isoCompIrrepIdx}^{\irrepMultiplicity[\isoCompIdx]} \irrep[\isoCompIdx]{}$ (\cref{eq:isotypic_decomposition}).

The significance of this decomposition lies in the orthogonality constraints between isotypic subspaces, $\confSpaceJS^{\iso}_{i} \perp \confSpaceJS^{\iso}_{j}$ for all $i\neq j$, due to the different symmetry subsets acting on each subspace. This orthogonal decomposition of the joint space configuration results in equivalent decompositions of the generalized velocities, forces, momentum, and work vectors, effectively enabling the study of the joint space dynamics as the superposition of the dynamics of each lower-dimensional \gls{ncm}, as discussed next.

\subsubsection*{Decomposing motions into a superposition of \glspl{ncm}}

Consider that the isotypic decomposition of the joint space configuration implies that with the change of coordinates $\mT$, we can decompose any robot's joint space state $(\qj, \dqj)$, along with the generalized acceleration and moving forces $(\ddqj, \genForcesJs)$, into their projections to each of the orthogonal lower-dimensional isotypic subspaces:

\begin{equation}
	\begin{split}
		\qjiso &:= \qjiso[(1)] \oplus \cdots \oplus \qjiso[(\isoCompNum)] = \mT \qj,
		\\
		\dqjiso &:= \dqjiso[(1)] \oplus \cdots \oplus \dqjiso[(\isoCompNum)] = \mT \dqj.
		\\
		\ddqjiso &:= \ddqjiso[(1)] \oplus \cdots \oplus \ddqjiso[(\isoCompNum)] = \mT \ddqj,
		\\
		\taujiso &:= \taujiso[(1)] \oplus \cdots \oplus \taujiso[(\isoCompNum)] = \mT \genForcesJs.
	\end{split}
	\label{eq:joint_space_configuration_decomp_into_iso_comp}
\end{equation}%
Where $\qjiso[(\isoCompIdx)], \dqjiso[(\isoCompIdx)], \ddqjiso[(\isoCompIdx)], \taujiso[(\isoCompIdx)] \in \R^{n_{k,j}}$ denote the projections of the joint space generalized positions, velocities, accelerations, and forces onto their repsective $\isoCompIdx^{\text{th}}$ isotypic subspaces of dimension $n_{\isoCompIdx,j}= |\confSpaceJS^{\iso}_{\isoCompIdx}| \leq |\confSpaceJS| = \nj$ (see \cref{fig:solo_decomposition_harmonic_states} and
\href{https://danfoa.github.io/DynamicsHarmonicsAnalysis/media/animations/mini-cheetah_Klein4-concrete_galloping_harmonic_analysis.gif?utm_source=\source}{animation}
\endnote{
	Harmonic analysis of the Mini Cheetah robot:
	Animation online at
	\href{https://danfoa.github.io/DynamicsHarmonicsAnalysis/media/animations/mini-cheetah_Klein4-concrete_galloping_harmonic_analysis.gif?utm_source=\source}{https://bit.ly/MiniCheetah-DHA-Trot-Concrete}
	\label{animation:mini_cheetah-concrete_galloping_harmonic_analysis}
}).

Note that in the isotypic basis, each dimension of $\qjiso \in \R^{\nj}$ no longer describes the position of a single \gls{dof} but rather of a synergy of \glspl{dof}. Likewise, each dimension of $\dqjiso, \ddqjiso, \taujiso$ describes the velocity, acceleration, and force charaterizing the dynamics of the synergy of \glspl{dof}.

Moreover, due to the orthogonality constraint between isotypic subspaces, the generalized moving forces of each subspace are constrained to do zero work along the directions of displacement in other isotypic subspaces. This results in the decomposition of the instantaneous joint space work $\workj := \tauj^\transpose \dqj$ into the sum of the work done in each isotypic subspace:
\begin{equation}
	\label{eq:decom_work_into_iso_comp}
	\small
	\workj = \sum_{\isoCompIdx=1}^{\isoCompNum} \workjiso[(\isoCompIdx)] := \sum_{\isoCompIdx=1}^{\isoCompNum} {\taujiso[(\isoCompIdx)]}^\transpose \dqjiso[(\isoCompIdx)] = (\mT \genForcesJs)^\transpose \mT \dqj = \genForcesJs^\transpose \dqj.
\end{equation}%

A similar decomposition follows for the generalized momentum $\momentumj := \Mass_{\js}(\qj) \dqj \in \R^{\nj}$, given by:

\begin{equation}
	\label{eq:decom_momentum_into_iso_comp}
	\small
	\momentumjiso = \momentumjiso[(1)] \oplus \cdots \oplus \momentumjiso[(\isoCompNum)] = \mT \momentumj = \mT (\Mass_{\js}(\qj) \dqj).
\end{equation}%
Note that the decomposition of the generalized momentum follows because this vector-valued function also transforms according to $\rep[\confSpaceJS]{}$, i.e., $\g \Glact \momentumj = \rep[\confSpaceJS]{\g} \momentumj = \Mass_{\js}(\gqj) \gdqj$ for all $\g \in \G$. Here, $\Mass_{\js}(\cdot)$ denotes the joint space generalized mass matrix.

To get an intuitive perspective in the aforementioned decomposition consider the dynamics of a rigid body $A$ in $3$D space acted on upon by a controlled force vector $\vf = [f_{\hat{x}}, f_{\hat{y}}, f_{\hat{z}}] \in \R^3$. The dynamics of such a system are commonly modeled and controlled independently in the orthogonal lower-dimensional directions $\hat{x}, \hat{y}, \hat{z} \sim \R \subset \R^3$ of $3$D space. The motivation for this decomposition lies in the orthogonality constraints of these directions of motion ensuring that each force component $f_{\hat{x}}, f_{\hat{y}}, f_{\hat{z}}$ does zero work on the directions of displacement orthogonal to it, rendering the dynamics along these directions of motion independent. Analogously, \cref{eq:joint_space_configuration_decomp_into_iso_comp,eq:decom_work_into_iso_comp,eq:decom_momentum_into_iso_comp} describe the decomposition of a robot's joint space state dynamics evolving in $\R^\nj$ into a superposition of $\isoCompNum$ orthogonal dynamics, each evolving in subspaces of lower dimension $\R^{n_{\isoCompIdx,j}} \subset \R^{\nj}$, for $\isoCompIdx \in [1, \isoCompNum]$. Similarly as in the rigid body example, this decomposition can be leveraged in analytic and data driven dynamics modelling \citep{ordonez2024dynamics} and in model-based optimal control and estimation \citep{amatucci2024accelerating,budhiraja2019dynamics}.

\section{Applications of morphological symmetries}\label{sec:applications}
\begin{figure*}[t!]
	\centering
	\resizebox{\linewidth}{!}{
		\input{images/K4_solo_data_augmentation.tex}
	}
	\vspace*{-1.5em}
	\caption[Symmetries of the quadruped robot Solo]{
		(a) Symmetries of the quadruped robot Solo \citep{grimminger2020solo_robot} and its sensor data measurements (see
		\href{https://danfoa.github.io/MorphoSymm/static/animations/solo-Klein4-symmetries_anim_static.gif?utm_source=\source}{animation}
		\endnote{Morphological Symmetry group of the Solo robot $\G=\KleinFourGroup$:
			See animation online at \href{https://danfoa.github.io/MorphoSymm/static/animations/solo-Klein4-symmetries_anim_static.gif?utm_source=\source}{https://bit.ly/Solo-MorphoSymm-Klein4}
			\label{animation:solo-Klein4-symmetries_anim_static}
		}). Here we consider the robot's morphological symmetry subgroup $\KleinFourGroup < \G := \KleinFourGroup \times \CyclicGroup[2]$ of order $|\KleinFourGroup|=4$, which consists of two perpendicular reflections $(\g_s, \g_t)$ and a $180^\circ$ rotation ($\g_r$) of space. These symmetry transformations influence the robot's state $\state$, resulting in the rotation/reflection of the state's associated proprioceptive sensor data measurements (contact forces $\vf \in \R^3$, centroidal linear $\linMomentum\in \R^3$ and angular $\angMomentum\in \R^3$ momenta), as well as exteroceptive measurements (contact points and contacts surface orientation $\mSE_c\in \SE[3]$).
		(b) Diagram depicting a $\KleinFourGroup$-equivariant neural network that processes the state and sensor data measurements of the Solo robot, $\nnIn$, and outputs the binary foot contact states for each leg, $\nnOut$. The network's $\G$-equivariance is achieved by constraining each linear layer's weight matrix $^i\mW \in \R^{n\times m}$ to the space of $\G$-equivariant matrices, such that ${}^i\mW = \g \Gconj{}^i\mW$ for all $\g\in \KleinFourGroup$. This constraint reduces the layer's trainable parameters to the vector of coefficients ${}^i\vc \in \R^{p}$, which mixes the $p<mn$ basis vectors of the layer's space of $\G$-equivariant matrices.
		\vspace*{-1em}
	}
	\label{fig:data_aug_equiv_fn_approx}
\end{figure*}
In this section, we present morphological symmetries as a physics informed geometric prior to be exploited in all data-driven applications of modeling, control, and estimation in robotics. We characterize how these symmetries can significantly enhance the performance of machine learning models and mitigate the challenges associated with data collection in robotics. In \cref{sec:data_augmentation}, we present morphological symmetries as a technique for augmenting datasets comprised of the robots' proprioceptive and exteroceptive sensor measurements. Alternatively, in \cref{sec:applications_equiv_fn_approx}, we show how to leverage morphological symmetries as constraints in learning and optimization processes, such as the optimization of the parameters of a neural network.
Lastly, in \cref{sec:dynamic_harmonic_analysis}, we introduce the implications of \gls{dha} and the isotypic basis for analytical methods in control and estimation.

\subsubsection*{Overview}
Data-driven applications of supervised, unsupervised, and reinforcement learning rely on the approximation a target function $f \in \functionSpace: \vsX \mapsto \vsY$ using a model $\nn \in \functionSpace$ parameterized by learnable parameters $\nnParams$, and a dataset $\sD = \{(\vx,\vy), \cdots\}$. Here, $\functionSpace$ is the hypothesis function space, encompassing all functions mapping from the input space $\vsX$ to the output space $\vsY$.

In the field of robotics, both the input and output spaces usually comprise a combination of state and sensor data measurements. When the robotic system possesses a morphological symmetry group $\G$, these symmetries are imprinted in both the state and sensor data measurements. Consequently, both $\vsX$ and $\vsY$ typically become symmetric vector spaces, and the target function $f$ transitions into a $\G$-invariant or $\G$-equivariant map, as outlined in \cref{eq:equivariance-invariance-constraints}. We can exploit this bias either through the use of data augmentation of by enforcing equivariance/invariance constraints on the machine learning model.
\subsection{Data augmentation} \label{sec:data_augmentation}
%
As discussed earlier, a morphological symmetry refers to a robotic system's ability to replicate a Euclidean isometry through a feasible state transformation. This isometry can be interpreted as a transformation of both the system's state and the robot's operational environment. Considering that any rotation, reflection, or translation of the world results in an equally plausible operational environment, it follows that any proprioceptive and exteroceptive sensor data measurement, captured by the system at a given state $\state$, can be related to a measurement taken by the system at the transformed state $\morphState$ within the transformed operational environment as depicted in~\cref{fig:data_aug_equiv_fn_approx}a.

Consider a dataset $\sD = \{(\vx_0,\vy_0), \cdots, (\vx_{|\sD|},\vy_{|\sD|})\}$ derived from the robot's operation in either a simulated or real-world environment. Each data point $(\vx,\vy)$ may comprise a combination of state $\state$ measurements and associated proprioceptive/exteroceptive sensor data measurements, such as contact forces, tactile sensing, height maps, depth maps, joint torques, etc (see \cref{fig:data_aug_equiv_fn_approx}a).

By leveraging the robot's morphological symmetry group $\G$, we can synthetically generate $|\G|$ samples per data point. This is achieved by considering the orbit of symmetric measurements $\G(\vx,\vy) = \{(\g \Glact \vx, \g \Glact \vy) \stforall \g \in \G\}$, effectively increasing the dataset size to $|\G||\sD|$, as depicted in \cref{fig:data_aug_equiv_fn_approx}a. Therefore, this method serves as a form of data augmentation, assuming the generated sample belongs to the same data distribution as the original dataset. In other words, if the probability of encountering the data point $(\vx,\vy)$ during the robot's normal operation should be the same as the probability of encountering $(\g \Glact \vx, \g \Glact \vy)$. For example, for every minute of recorded data from the Mini Cheetah robot with $|\G|=8$, we can augment the dataset with an additional 7 minutes of recordings by applying the symmetries to the data (see animation \ref{animation:mini_cheetah-C2xC2xC2-symmetries_anim_static}).

To augment the data, we need to define the action of a symmetry on the input/output spaces $\vsX$ and $\vsY$. This involves identifying the group representations $\rep[\vsX]{g} \in \GLGroup(\vsX)$ and $\rep[\vsY]{g} \in \GLGroup(\vsY)$, which allows us to apply the group actions $\g \Glact \vx:= \rep[\vsX]{g} \vx$ and $\g \Glact \vy:= \rep[\vsY]{g} \vy$. Since $\vsX$ and $\vsY$ are comprised of proprioceptive/exteroceptive sensor data measurements, these group representations are constructed from direct sums/products of the group representations we identified in \cref{sec:background,sec:morphological_symmetries,sec:morphological_symmetries_rigid_body}.

\subsubsection*{Exteroceptive measurements}
These measurements can be represented as individual points, and spatial vectors, or a combination thereof. They may encompass contact points, contact forces, terrain or surface normals, depth images, and the positions and orientations of bodies in space. To transform these measurements, we employ the group representations $\rep[\R^{\dimEvolution + 1}]{}$ and $\rep[\R^{\dimEvolution}]{}$ (or direct sums/products of these representations). These represent the rotation/reflection or translation associated with each $\g \in \G$, as detailed in \cref{eq:group_representations_rigid_transformations}.


\subsubsection*{Proprioceptive measurements}
Proprioceptive measurements can often be represented as geometric quantities in $\R^{\dimEvolution}$. Examples include contact points, end-effector positions/velocities. To transform these we use $\rep[\R^{\dimEvolution + 1}]{}$ and $\rep[\R^{\dimEvolution}]{}$, as above. Alternatively, proprioceptive measurements can be points in the system's joint space $\confSpaceJS$ and $\tangConfSpaceJS$ spaces, such as joint positions/velocities/accelerations/torques. To transform these measurements, we employ the group representations $\rep[\tangConfSpaceJS]{}$, $\rep[\confSpace]{}$ (characterized in \cref{sec:morphological_symmetries_rigid_body}). When the measurements are associated with one of the robot's $\nuch$ unique kinematic branches, such as leg contact states in a locomoting system (see \cref{fig:data_aug_equiv_fn_approx}) or the end-effector position of lim, we utilize the permutation representations $[\rep[\setKinStructLabels_i]{}]_{i=1}^{\nuch}$.

The above-mentioned representations allow us to apply an appropriate reordering of measurements, according to the permutations of the branches (see \cref{sec:structure_of_the_joint_space_action_transformation}).
For instance, consider the task of augmenting the vector composed of the three body velocities of the last body of each limb of the robot depicted in \cref{fig:G_C3}b. This vector, denoted as $\vx = [\vel^\transpose_{l,l_1,2}, \vel^\transpose_{l,l_2,2}, \vel^\transpose_{l,l_3,2}]^\transpose \in \R^{9}$, can be transformed using the group representation for vectors in $\R^{\dimEvolution}$ (i.e., $\rep[\R^{\dimEvolution}]{}$) and the limb's permutation representation $\rep[\setKinStructLabels_{l}]{}$. The resulting  representation $\rep[\vsX]{}:= \rep[\setKinStructLabels_{l}]{} \otimes \rep[\R^{\dimEvolution}]{}$ is obtained in a manner analogous to the process outlined in \cref{eq:joint_space_configuration_definition}.
\subsection{Equivariant/Invariant function approximation}
\label{sec:applications_equiv_fn_approx}
In robotics, the target function $f \in \functionSpace: \vsX \mapsto \vsY$ we aim to approximate is often a $\G$-invariant or $\G$-equivariant map. This constraint narrows the hypothesis space $\functionSpace$ to either the space of $\G$-invariant functions $\functionSpace_{\G}^{inv}$ or $\G$-equivariant functions $\functionSpace_{\G}^{eq}$ \citep{weiler2023EquivariantAndCoordinateIndependentCNNs}. As introduced above, these are defined as:
\begin{equation*}
	\begin{split}
		\functionSpace_{\G}^{inv} &= \{f \in \functionSpace \st \vy = f(\g \Glact \vx), \forall \g\in \G\},
		\\
		\functionSpace_{\G}^{eq} &= \{f \in \functionSpace \st \g \Glact \vy = f(\g \Glact \vx), \forall \g\in \G\}.
	\end{split}
\end{equation*}
%
Therefore, an alternative to data augmentation when approximating $\G$-invariant/equivariant functions is to constrain the function space of the machine learning model. This can be achieved by ensuring that $\nn \in \functionSpace_{\G}^{inv}$ or $\nn \in \functionSpace_{\G}^{eq}$. While this constraint reduces the model's expressivity by limiting the types of functions it can represent \citep{bronstein2021geometric}, it also enhances the model's generalization capabilities beyond the training distribution and improves sample efficiency. Below, we include examples of $\G$-equivariant and $\G$-invariant functions in robotic systems with morphological symmetry group $\G$.

\subsubsection*{$\G$-equivariant functions}
Examples of $\G$-equivariant functions in robotics include the forward and inverse dynamics, as outlined in \cref{eq:eom_G_equivariance}, and the generalized mass matrix function, as per \cref{eq:mass_and_forces_g_equivariance}. Other examples include the robot's positional and rotational Jacobians, collectively represented as a map $\Jacob: \confSpace \times \confBundle \mapsto \se_\dimEvolution$ introduced in \cref{eq:kinematic_symmetries_rigid_bodies_b} (\cite{wieber2006holonomy}). The inverse and forward kinematics functions, defined as the maps $f_{ik}: \EG \mapsto \confSpace$ and $f_{fk}: \confSpace \mapsto \EG$, respectively, also fall into this category. Additionally, practical non-analytical functions, such as contact location detection (see \cref{fig:data_aug_equiv_fn_approx}), can also be considered $\G$-equivariant.

\subsubsection*{$\G$-Invariant Functions}
Examples of $\G$-invariant functions in robotics include the robot's kinetic energy $\kinE: \confSpace \times \confBundle \mapsto \R$ and potential energy $\potE: \confSpace \mapsto \R$, as outlined in \cref{def:symmetry_of_a_robotic_system}. Additionally, practical non-analytical functions such as the detection of external disturbance and emergency stop scenarios also fit into this category.


\subsubsection*{Optimal control and reinforcement learning}
\label{sec:applications_reinforcement}
In optimal control and reinforcement learning settings, the control policy $\pi: \vsS \mapsto \vsA$ is constrained to be $\G$-equivariant functions when both of the \gls{mdp} state $\vsS$ and action $\vsA$ spaces are symmetric spaces, and the reward function $r: \vsS \times \vsA \mapsto \R$ is $\G$-invariant \citep{zinkevich2001symmetry_mdp_implications,wang2022robot_learning_equivariant_models}. This is particularly pertinent for robots with morphological symmetries, as the \gls{mdp} state space is typically defined as $\vsS := \confSpace \times \confBundle \times \vsX_d$ and $\vsA$ is usually a subspace of $\confBundle$. Here, $\vsX_d$ represents a space of relevant sensor data measurements for control, such as perception sensor measurements and contact states. Therefore, both $\vsS$ and $\vsA$ are typically symmetric spaces.

Most reward functions are $\G$-invariant in practice, as these functions are often defined in terms of distances in Euclidean space and/or configuration space, such as, distance to target point/configuration \citep{wang2022so2_equivariant_rl,brehmer2023geometric2}.
These measurements remain invariant under Euclidean isometries and morphological symmetries. Moreover, the $\G$-invariance of the reward functions translates to the $\G$-invariance of the value $V_\pi: \vsS \mapsto \R$, and action-value function $Q_{\pi}: \vsS \times \vsA \mapsto \R$
\citep{zinkevich2001symmetry_mdp_implications}. These are the fundamental functions upon which optimal control and reinforcement learning algorithms are built \citep{sutton2018reinforcement,zinkevich2001symmetry_mdp_implications}. Several works have provided empirical evidence of the benefits of symmetry exploitation in approximating these functions \citep{brehmer2024edgi2,rezaei2022continuousMDPhomomorphisms,wang2022robot_learning_equivariant_models,ordonez2022adaptable,weissenbacher2022koopman_symmetries,mondal2022eqr_rl,van2020mdp,mittal2024symmetry,su2024leveraging,wang2022approximately}.
\subsection{Dynamics' Harmonics Analysis}\label{sec:dynamic_harmonic_analysis}
%
\begin{figure*}[t!]
	\centering
	\begin{subfigure}[b]{0.32\textwidth}
		\centering
		\includegraphics[width=\textwidth]{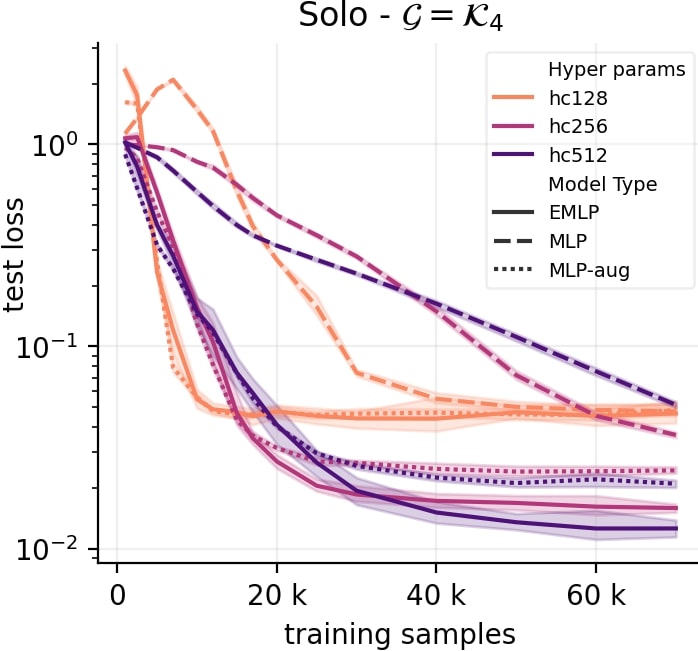}
		\label{fig:solo_results_k4_diff_sizes}
	\end{subfigure}
	\begin{subfigure}[b]{0.32\textwidth}
		\centering
		\includegraphics[width=\textwidth]{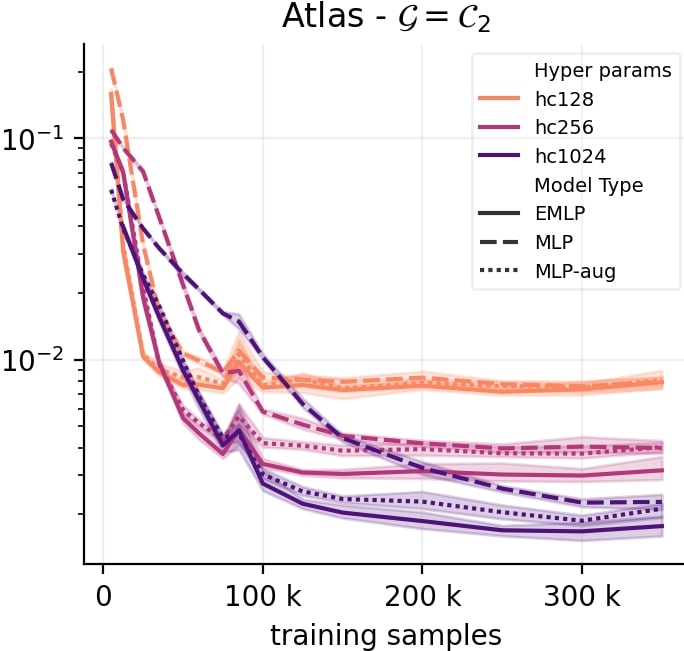}
		\label{fig:atlas_results_c2_diff_sizes}
	\end{subfigure}
	\begin{subfigure}[b]{0.32\textwidth}
		\centering
		\includegraphics[width=\textwidth]{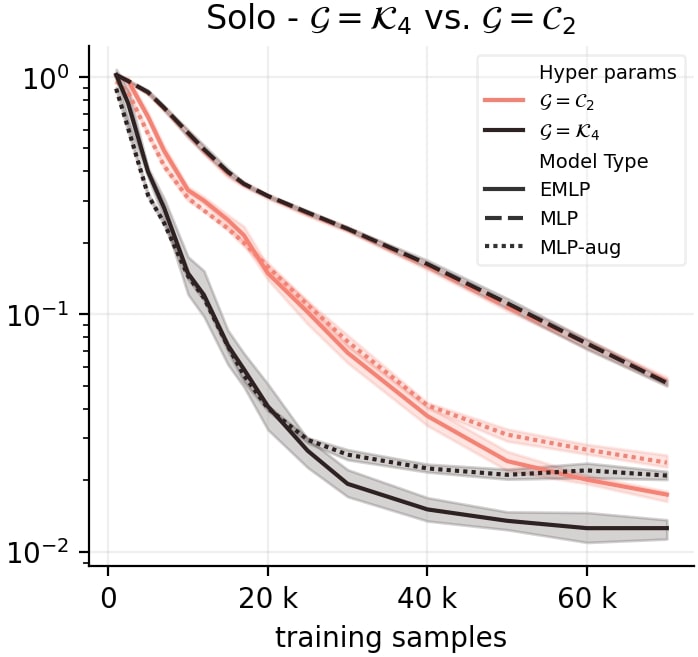}
		\label{fig:solo_results_k4_c2_comparison_512}
	\end{subfigure}
	\vspace*{-1.3em}
	\caption{
	Centroidal estimation results comparing \acrshort{mlp}, \acrshort{mlp-aug}, and \acrshort{emlp} models. Left and Middle: Test set sample efficiency of model variants with different capacities (number of neurons \textit{hc} in hidden layers) for the Solo and Atlas robots. Right: Sample efficiency for the Solo robot with models having $hc=512$, when exploiting $\G=\KleinFourGroup$ (sagittal and traversal symmetries) and $\G=\CyclicGroup[2] ={e, \g[s]} \subset \KleinFourGroup$ (only sagittal symmetry). The plots depict the average and standard deviation across 10 seeds.
	\vspace*{-1em}
	}
	\label{fig:com_estimation_results}
\end{figure*}
As detailed in \cref{sec:joint_space_basis_harmonic_motions}, the joint space dynamics of a robotic system with morphological symmetries can be decomposed into a superposition of $\isoCompNum$ lower-dimensional dynamics evolving in a \gls{ncm}. This decomposition is exposed after applying a change of basis $\mT$ to $\confSpaceJS$, mapping from a canonical basis to the isotypic basis, as referenced in \cref{eq:decomp_joint_space_rep} and depicted in \cref{fig:solo_decomposition_harmonic_states}.

This decomposition of the dynamics enables the translation of the problems of modeling and control of robotic systems with $\nj$ \gls{dof}, to the modeling and control of $\isoCompNum$ subsystems with $n_{\isoCompIdx,j} \leq \nj$ \gls{dof} each. This structure can be leveraged to mitigate the state dimension cubic computational complexity, $O(\nj^3)$, of trajectory optimization and optimal estimation algorithms \citep{amatucci2024accelerating,alessandri2003ddpestimation}.

Moreover, in the isotypic basis of $\confSpaceJS$ each dimension describes the amplitude of a specific synergy of motion of multiple \gls{dof}. This basis can be leveraged also for the control action sapce, for which the change of a dimension of the action space corresponds to the modulation of the dynamics of the corresponding synergy of motion. This synergistic  decomposition and control aligns with theories in biomechanics and motor control \citep{scholz1999uncontrolled,feldman2009equilibrium,ijspeert2008central}, which sustain that the control of a high-\gls{dof} system is not achieved by individually controlling each \gls{dof}. Instead, it is controlled as the superposition of several lower-dimensional synergistic motions, often referred to in robotics as dynamic motion primitives (DMPs) \citep{schaal2003control}. In this context, the dimensions of each of the robot's \gls{ncm} can be interpreted as \highlight{symmetric} DMPs. This basis is derived analytically from the knowledge of the system's morphological symmetries, providing a systematic way to identify these lower-dimensional synergistic motions for arbitrary symmetric systems.
%
\section{Experiments}\label{sec:experiments}
%
In this section, we empirically demonstrate the benefits of leveraging morphological symmetries in data-driven methods. We examine the impact of symmetry exploitation on the generalization and sample efficiency of neural network-based machine learning models. Specifically, we focus on two cases: data augmentation and equivariance/invariance constraints. To do so, we conducted synthetic and real-world experiments involving the approximation of several $\G$-invariant and $\G$-equivariant functions. These are regression and classification problems in supervised learning.

We conduct a synthetic experiment that involves a regression problem, with the goal of approximating the $\G$-equivariant function that computes the system's centroidal momentum, as detailed in \cref{subsec:com_momentum_estimation}. In addition, we present a real-world classification problem in \cref{subsec:contact_detection}, which utilizes a dataset of the Mini Cheetah quadruped robot's locomotion across various terrains and using different gaits \citep{lin2021deep_contact_estimation}. Specifically, our focus is on approximating the $\G$-equivariant static-friction-regime contact detection function, a crucial component in leg odometry and state estimation.

To approximate these functions, we trained a conventional neural network model $\nn \in \functionSpace$ on a dataset $\sD=\{(\vx_0,\vy_0),\cdots,(\vx_{|\sD|},\vy_{|\sD|})\}$, which was partitioned into training, validation, and testing sets (70\%, 15\%, and 15\% of $|\sD|$, respectively). We compared this model against another trained on an augmented training dataset of size $0.7|\G||\sD|$. Additionally, we compared against a $\G$-equivariant model $\nn[\nnParams_{eq}] \in \functionSpace_{\G}^{eq}$ (\cref{sec:applications_equiv_fn_approx}). Both of these models leverage the robot's morphological symmetry group (\cref{sec:applications}).

Finally, to motivate our finding in \cref{sec:dynamic_harmonic_analysis}, we present in \cref{subsec:exp_harmonyc_analysis_mini_cheetah} how \gls{dha} can be used to characterize the lower-dimensional synergistic nature of the dynamics of legged locomotion in quadruped robots.

\subsection{Centroidal momenta estimation $\G$-equivariant regression}
\label{subsec:com_momentum_estimation}
%
%
\begin{figure*}[t!]
	\centering
	\begin{subfigure}[b]{0.325\textwidth}
		\centering
		\includegraphics[width=\textwidth]{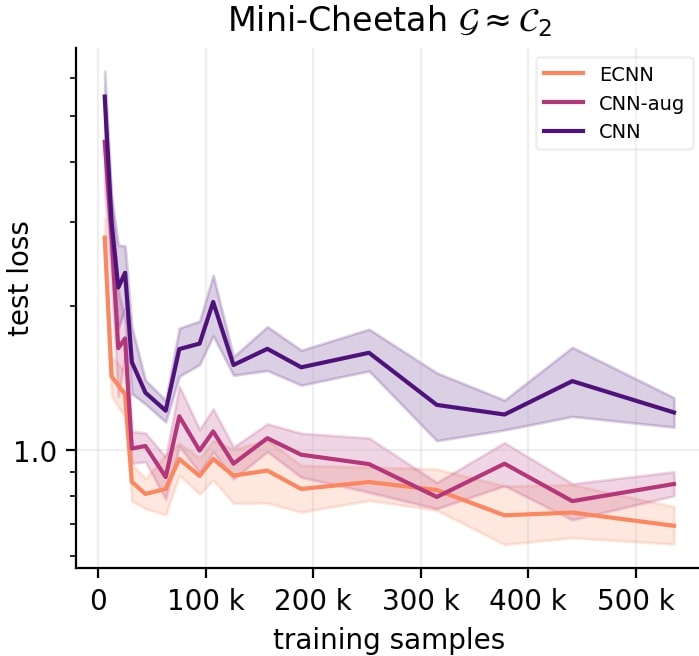}
	\end{subfigure}
	\begin{subfigure}[b]{0.325\textwidth}
		\centering
		\includegraphics[width=\textwidth]{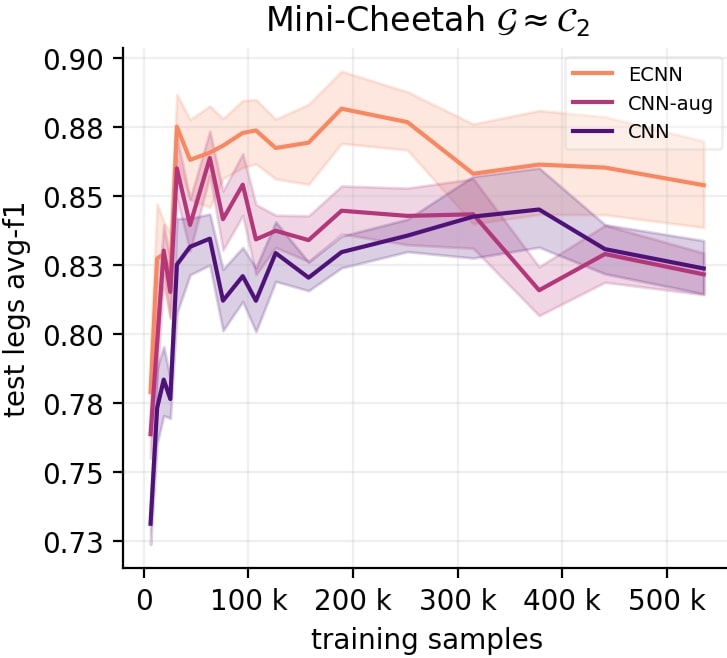}
	\end{subfigure}
	\hfill
	\begin{subfigure}[b]{0.325\textwidth}
		\centering
		\includegraphics[width=\textwidth] {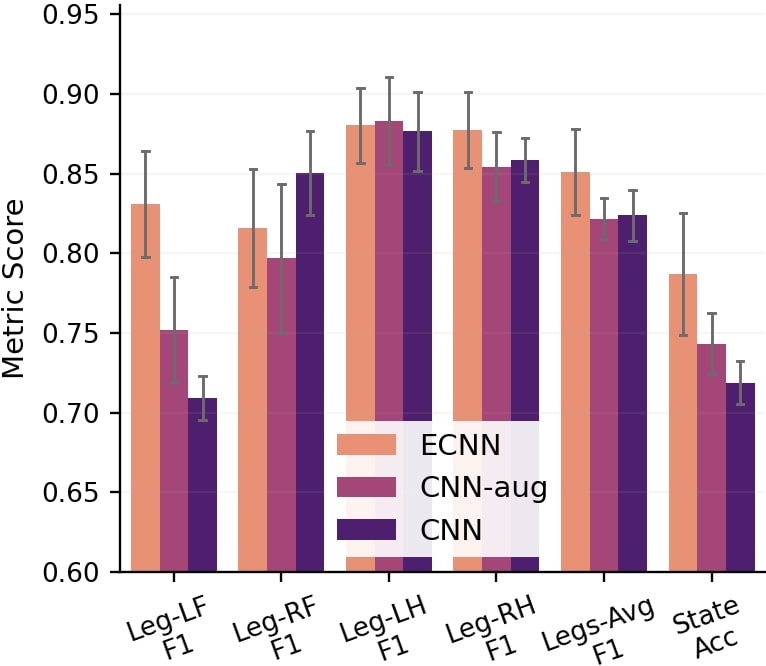}
	\end{subfigure}
	\caption{Static-Friction-Regime contact detection results comparing \gls{cnn}, \gls{cnn-aug}, and \gls{ecnn}. Left: Sample efficiency in log-log scale. Middle: Average legs F1-score. Right: Classification metrics on test set performance of models trained with the entire training set. The selected metrics include contact-state ($\nnOut\in \R^{16}$) accuracy (Acc) and f1-score (F1) for each leg binary contact state. Due to the sagittal symmetry of the robot, the left front (LF) and right front (RF) legs are expected to be symmetric, as well as the left hind (LH) and right hind (RH) legs. F1-score is presented considering the class imbalanced nature of the dataset. The reported values represent the average and standard deviation across $8$ different seeds.
		\vspace*{-1em}
	}
	\label{fig:contact_results}
\end{figure*}

We approximated the robot's linear $\linMomentum \in \R^3$ and angular $\angMomentum\in \R^3$ momentum as a function of the system's state $\state$. This involved computing the momentum of each of the robot's $\nb$ bodies and aggregating these relative to the state-dependent \gls{com} location. As introduced above, the centroidal momentum is a highly non-linear and $\G$-equivariant function reliant on the robot's kinematic and dynamic parameters with morphological symmetries (refer to \cref{sec:morphological_symmetries_rigid_body} and \cref{fig:data_aug_equiv_fn_approx}a):

\begin{equation}
	\small
	\begin{split}
		\g \Glact \momentum
		=
		\CMM(\g \morphOp \q) \g \Glact\dq
		\\
		\st \g \Glact \momentum := \rep[\vsH]{\g} \momentum
		=
		\begin{bsmallmatrix}
			\mR_\g & \bm{0} \\
			\bm{0} & |\mR_\g|\mR_\g
		\end{bsmallmatrix}
		\begin{bsmallmatrix}
			\linMomentum \\
			\angMomentum
		\end{bsmallmatrix},
		\forall \g \in \G,
	\end{split}
	\label{eq:com_momentum_pin}
\end{equation}%
where $\momentum \in \vsH \subseteq \R^6$ denotes the stacked linear and angular momentum components, and $\CMM$ is the state-dependent \gls{cmm} function introduced by \citet{orin2013centroidal_momentum_matrix_cmm}. The group representation $\rep[\vsH]{}$ outlines the corresponding rotation or reflection of these momentum components. Importantly, the action on the flat angular momentum, $|\mR_\g|\mR_\g, $\endnote{The action of a rotation/reflection on a flat pseudovector (or axial vector) $\angMomentum$ is defined as $\g \Glact \angMomentum := |\mR_\g|\mR_\g \angMomentum$. This is equivalent to the symmetry action described in \cref{eq:def_morph_op}, in which axial vectors is represented as a member of the special Orthogonal group lie algebra $\so_\dimEvolution$, that is, as a skew-symmetric representation of the axial vector. In this form the group action is defined as $\g \Glact [\angMomentum]_\times := \mR_\g \Gconj [\angMomentum]_\times = \mR_\g [\angMomentum]_\times \mR_\g^{\transpose}$
	\label{note:action_pseudovector}
}
signifies the necessary change of sign for axial or pseudovector quantities in the event of a reflection \citep{quigley1973pseudovectors}, as depicted in \cref{fig:data_aug_equiv_fn_approx}a.

To approximate this function, we utilized a standard \gls{mlp} model $\nn \in \functionSpace: \confSpace \times \confBundle \mapsto \vsH$, trained using a synthetic dataset $\sD=\{(\state,\vh),\cdots\}$, generated with \cref{eq:com_momentum_pin} and \textsc{Pinocchio} \citep{carpentier2019pinocchio}. We compared with the model trained using the augmented dataset \gls{mlp-aug}, and a $\G$-equivariant version of the model $\nn[\nnParams_{eq}] \in \functionSpace_{\G}^{eq}$ (refer to  \cref{sec:applications_equiv_fn_approx}), denoted as \gls{emlp}.
We tested two robots: Atlas, a $32$-\gls{dof} humanoid robot with $\G = \CyclicGroup[2]$ (see \cref{fig:atlas_dynamics_equivariance}), and Solo, a $12$-\gls{dof} quadruped robot with $\G=\KleinFourGroup$ (see \cref{fig:data_aug_equiv_fn_approx}).

In \cref{fig:com_estimation_results}-left/middle, we compare the performance of several \gls{mlp} models with varying numbers of trainable parameters, against their counterparts using data augmentation \gls{mlp-aug} and $\G$-equivariance constraints \gls{emlp}. Across both robots and all model capacities, \gls{emlp} and \gls{mlp}-Aug outperform \gls{mlp} in terms of sample efficiency (better generalization with fewer data) and robustness to overfitting when training data is limited. Among the \gls{emlp} and \gls{mlp-aug} variants, the models with the lowest number of trainable parameters exhibit similar sample complexity and performance, but as the model's capacity increases, the \gls{emlp} model reaches a superior sample efficiency and generalization. In addition, \cref{fig:com_estimation_results}-right shows a comparison for the Solo robot, evaluating the performance of the model variants when exploiting either the entire symmetry group ($\KleinFourGroup$) or a subgroup of the true symmetry group ($\CyclicGroup[2] \subset \KleinFourGroup$). The results indicate that sample efficiency and generalization capacity increase with the number of \highlight{true} symmetries of the data exploited.
%
\subsection{Static-friction-regime contact detection $\G$-equivariant classification}
\label{subsec:contact_detection}
\begin{figure*}[t!]
	\centering
	\includegraphics[width=.93\textwidth]{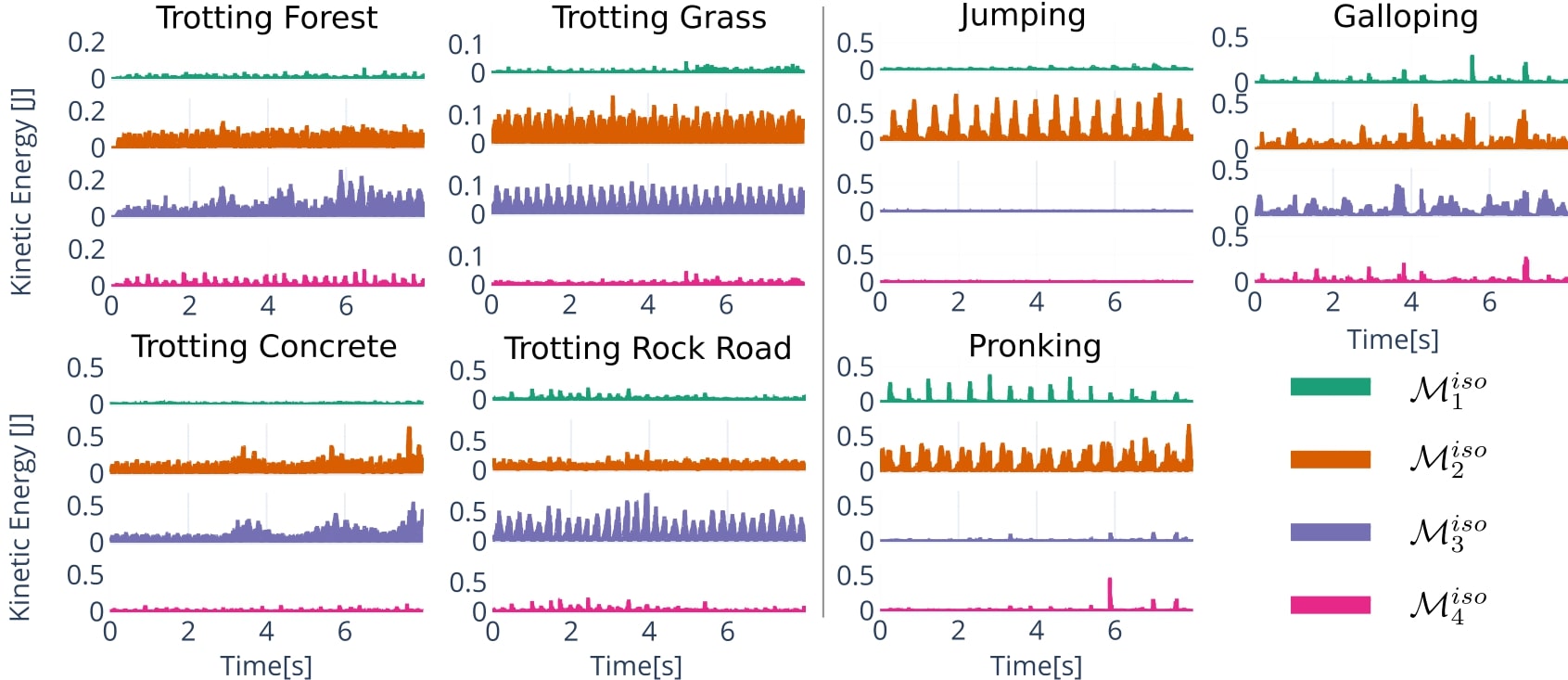}
	\caption{
		Decomposition of the joint space dynamics of legged locomotion of the Mini Cheetah robot as a superposition of lower dimensional \gls{ncm}. The dynamics of locomotion for different gait types and terrains are characterized by the projection of trajectories of motion onto the isotypic spaces $\confSpaceJS^{\iso}_{1},\confSpaceJS^{\iso}_{2},\confSpaceJS^{\iso}_{3},\confSpaceJS^{\iso}_{4} \subseteq \R^3$. The relevance of each isotypic subspace, per gait type and terrain, is quantified by computing the joint space kinetic energy of the projected motion in the isotypic subspace. \textbf{Left:} The trotting gait dynamics on \href{https://danfoa.github.io/DynamicsHarmonicsAnalysis/media/animations/mini-cheetah_Klein4-forest_harmonic_analysis.gif?utm_source=\source}{forest}, \href{https://danfoa.github.io/DynamicsHarmonicsAnalysis/media/animations/mini-cheetah_Klein4-grass_harmonic_analysis.gif?utm_source=\source}{grass}, \href{https://danfoa.github.io/DynamicsHarmonicsAnalysis/media/animations/mini-cheetah_Klein4-concrete_difficult_slippery_harmonic_analysis.gif?utm_source=\source}{concrete}, \href{https://danfoa.github.io/DynamicsHarmonicsAnalysis/media/animations/mini-cheetah_Klein4-sidewalk_harmonic_analysis.gif?utm_source=\source}{road with rocks} terrains primarily evolve in the $\confSpaceJS^{\iso}_{2} \times \confSpaceJS^{\iso}_{3} \in \R^6$ subspace, as indicated by the kinetic energy distribution. \textbf{Right:} The \href{https://danfoa.github.io/DynamicsHarmonicsAnalysis/media/animations/mini-cheetah_Klein4-air_jumping_gait_harmonic_analysis.gif?utm_source=\source}{jumping}, \href{https://danfoa.github.io/DynamicsHarmonicsAnalysis/media/animations/mini-cheetah_Klein4-concrete_pronking_harmonic_analysis.gif?utm_source=\source}{pronking},  gaits on asphalt terrain mainly evolve in $\confSpaceJS^{\iso}_{2}$ and $\confSpaceJS^{\iso}_{1} \times \confSpaceJS^{\iso}_{2} \in \R^6$ subspaces, respectively, while the \href{https://danfoa.github.io/DynamicsHarmonicsAnalysis/media/animations/mini-cheetah_Klein4-concrete_galloping_harmonic_analysis.gif?utm_source=\source}{galloping} gait dynamics periodically evolve in all isotypic subspaces.
	}
	\vspace*{-4mm}
	\label{fig:decomposition_mini_cheetha_trajecories}
\end{figure*}
We aim to approximate a function detecting the static-friction-regime binary contact state of the four legs of the Mini Cheetah quadruped robot, with symmetry group $\G=\KleinFourGroup\times\CyclicGroup[2]$. This function's inputs are the history of the state and proprioceptive sensor data measurements and outputs are binary variables per leg indicating whether the robot's leg is in a static-friction-regime state or in a non-contact or slipping state. These categorical estimated measurements are fundamental in state estimation pipelines dependent on leg odometry \citep[e.g.,][]{camurri2017probabilistic,lin2021deep_contact_estimation} and legged locomotion control \citep[e.g.,][]{mastalli2020planning,lee2020learning,corberes2024perceptive}.

The dataset $\sD = \{([\vx_{t}]_{t=0}^{-150}, \vy_{t}), \cdots \}$, first introduced by \citet{lin2021deep_contact_estimation}, comprises samples containing a categorical variable $\vy \in \R^{16}$. This variable represents the ground truth system's contact state among the $16$ different combinations of each of the $4$ legs' possible binary contact states. Each sample also includes a history of the past $150$ frames of a vector of proprioceptive sensor data measurements $\vx = [\qj, \dqj, ^{\base}\acc_{\base}, ^{\base}\angvel[\base], ^{\base}\pos_{feet}, ^{\base}\vel_{feet}] \in \vsX \subseteq \R^{54}$. Here, $\qj \in \confSpaceJS \subseteq \R^{12}$ and $\dqj \in \tangConfSpaceJS \subseteq \R^{12}$ denote the robot's joint space position and velocity generalized coordinates, $^{\base}\acc_\base$ and $^{\base}\angvel[\base] \in \R^{3}$ represent the robot's base linear acceleration and angular velocity, and $^{\base}\pos_{feet} = [^{\base}\pos_{RF},^{\base}\pos_{LF},^{\base}\pos_{RH},^{\base}\pos_{LH}] \in \vsX_{feet} \subseteq \R^{12}$ and $^{\base}\vel_{feet} = [^{\base}\vel_{RF},^{\base}\vel_{LF},^{\base}\vel_{RH},^{\base}\vel_{LH}] \in \vsX_{feet} \subseteq \R^{12}$ are the positions and velocities of each of the four legs' feet, respectively.

The ground truth contact state is estimated offline using a non-causal algorithm, i.e., a function dependent on both past and future proprioceptive measurements. The experiment's objective is to train a model to predict the contact state using only past proprioceptive data. Our unconstrained model $\nn \in \functionSpace: \vsX \mapsto \vsY$ is the original \gls{cnn} architecture from \citet{lin2021deep_contact_estimation}. We compare this model against the \gls{cnn-aug} and the \gls{ecnn}, denoted $\nn[\nnParams_{eq}] \in \functionSpace_{\G}^{eq}$.

\subsubsection*{Determining the input and output group representations $\rep[\vsX]{}$ and $\rep[\vsY]{}$}

To define the group action on the input space, we proceeded as detailed in \cref{sec:constraints_on_the_kinematic_structure} and identified the labels $\setKinStructLabels_{leg} = \{RF, LF, RH, LH\}$ of the four instances of the unique kinematic branches along with the label's permutation representation $\rep[\setKinStructLabels_{leg}]{}$. These labels describe the change of roles of the legs under a symmetry transformation (see \cref{fig:teaser}). Since the position and velocity measurements in $^{\base}\pos_{feet}$ and $^{\base}\vel_{feet}$ are associated with each leg, we rotated/reflected these measurements and permuted the robot's legs to define the symmetry transformations, i.e., $\rep[\vsX_{feet}]{} := \rep[\setKinStructLabels_{leg}]{} \otimes \rep[\R^3]{}$. This enabled us to define the input space group representation as $\rep[\vsX]{} := \rep[\confSpaceJS]{} \oplus \rep[\tangConfSpaceJS]{} \oplus \rep[\R^3]{} \oplus \rep[\R^3]{} \oplus \rep[\vsX_{feet}]{} \oplus \rep[\vsX_{feet}]{}$.

The sampling efficiency and average leg contact state classification results are depicted in \cref{fig:contact_results}-left-middle. The equivariant model \gls{ecnn} has superior generalization performance and robustness to dataset biases compared to the unconstrained models. Following \gls{ecnn}, \gls{cnn-aug} exhibits better performance than the original \gls{cnn}. In \cref{fig:contact_results}-right, we report the classification metrics of the test set when using the entire training data. The \gls{ecnn} model outperforms both \gls{cnn-aug} and \gls{cnn} in contact state classification and average leg contact detection. Notably, exploiting symmetries held mitigate suboptimal asymmetries in the models, preventing them from favoring the classification of one leg over others (observe legs LF and RF in \cref{fig:contact_results}-right).
\subsection{Dynamics harmonic analysis applied to the locomotion of a quadruped robot}
\label{subsec:exp_harmonyc_analysis_mini_cheetah}

In this experiment, we decompose the Mini Cheetah robot's joint space into $\isoCompNum=4$ \glspl{ncm} to study and characterize various locomotion gaits, across different terrains (refer to \cref{sec:joint_space_basis_harmonic_motions}).

Specifically, we took real-world joint space motion trajectories $[(\q[js,t], \dq[js,t])]_{t\in \Time}$, collected over a certain time span $\Time$ from the dataset in \cref{subsec:contact_detection}, and projected them onto the robot's isotypic subspaces. This projection is accomplished by applying a change of basis $\mT$ to $\confSpaceJS$, as outlined in \cref{eq:joint_space_configuration_decomp_into_iso_comp}. The result are the lower-dimensional trajectories $[(\q[js,t]^{(\isoCompIdx)}, \dq[js,t]^{(\isoCompIdx)})]_{t\in \Time}$ of each of the $\isoCompIdx \in [1,\isoCompNum]$ isotypic subspaces, representing the projections of the state trajectory into each \gls{ncm}. For instance, see an example of the decomposition of a motion trajectory into four lower-dimensional trajectories in animation \ref{animation:mini_cheetah-concrete_galloping_harmonic_analysis}.

To quantify which \glspl{ncm} are relevant in characterizing the different locomotion gaits, we take each of these lower-dimensional trajectories and independently compute their joint space kinetic energy. This value is a proxy indicating how much of the original trajectories generalized momentum and forces is associated with each \gls{ncm} (see \cref{eq:decom_momentum_into_iso_comp,eq:decom_work_into_iso_comp}). In \cref{fig:decomposition_mini_cheetha_trajecories}, we show how this decomposition can be used to identify the lower-dimensional nature of the trotting, jumping, pronking, and galloping gaits of the Mini Cheetah robot. Each of these gaits primarily evolves in a $3$ or $6$-dimensional subspace of the joint space composed of one or two \glspl{ncm}. Furthermore, the less relevant subspaces for each gait appear to be exited only temporarily during disturbance rejection. Interestingly, even when the terrain changes, the distribution of kinetic energy between isotypic spaces remains unaffected, as depicted in \cref{fig:decomposition_mini_cheetha_trajecories}-left for a trotting gait. This indicates that to control these dynamic behaviors, the bulk of the control policy optimization is expected to be performed in the relevant isotypic subspaces.

\section{Conclusions}
\label{sec:conclusion}

This work presents a framework based on group and representation theory for leveraging morphological symmetries in robotics. These symmetries, inherent in a robot's body structure, correspond to symmetries in mass distribution and the duplication of kinematic branches.
We illustrated how these symmetries extend to the robot's state space and both proprioceptive and exteroceptive measurements, resulting in the equivariance of the robot's equations of motion and optimal control policies. Furthermore, we revealed the potential of these symmetries to enhance both data-driven and analytical approaches for modeling, control, and estimation in robotics.

In the context of data-driven methods, we showed how morphological symmetries can be leveraged to boost a machine learning model's sample efficiency and generalization capabilities.
This underscores their value in addressing the robustness and reliability requirements that are critical for deploying data-driven models in robotics applications, where the challenges of data collection are particularly pronounced. The enhancement of the model can be realized either through data augmentation or by enforcing equivariance/invariance constraints on the models. We provided empirical evidence of these benefits through two experiments--a synthetic regression and a real-world classification problems-- involving both a bipedal and a quadrupedal robot.

In the context of analytical methods, we leveraged morphological symmetries to simplify the modeling and control of symmetric robotic systems. Specifically, we used abstract harmonic analysis to identify the block-diagonal structure of a robot's joint space mass matrix, leading to the decomposition of the robot's dynamics into a superposition the independent dynamics of several lower-dimensional synergistic symmetry-constrained robot motions.
This method, inspired by the use of harmonic analysis in particle physics, is introduced in robotics by the term \acrfull{dha}. We have demonstrated the relevance of this decomposition in the context of legged locomotion, by identifying the low-dimensional synergistic nature of the dynamics of locomotion of the Mini Cheetah robot, providing insights into the lower-dimensional spaces relevant for the motion control per each gait type.

\subsection*{Discussion and future work}
We identified three relevant directions of exploration of the implications of morphological symmetries
\subsubsection*{Optimization}
The use of \Gls{dha} (\cref{sec:joint_space_basis_harmonic_motions}), facilitates the parallel optimization on each of the orthogonal isotypic subspaces of $\confSpaceJS$ in trajectory optimization methods. Notably, this decomposition can be leveraged in the iLQR \citep{li2004iLQR} and DDP \citep{mayne1966DDP} algorithms, which play a crucial role in trajectory optimization for control and estimation \citep{tassa2014control,mastalli-crocoddyl,alessandri2003ddpestimation,kobilarov2015estimation}.

\subsubsection*{Computational Design}
Morphological symmetries provide substantial advantages for both data-driven and analytical methods in modeling, control, and estimation in robotics. The extent of these benefits is proportional to the order and structure of the symmetry group. This underscores the potential use of computational design in optimizing a system's morphology to exploit these implications.

\subsubsection*{Continuum, Soft, and Modular Robots}
Morphological symmetries are a common feature in soft, continuum, and modular robots. These robots often lack analytical solutions for forward/inverse kinematics/dynamics functions, necessitating the use of data-driven methods for function approximation. The presence of morphological symmetries in these robots imposes $\G$-equivariant/invariance constraints on these target functions. This provides a means to integrate the invariances inherent in Newtonian physics into the modeling and control of these robot types.


\section*{Delcaration of conflicting interests}
The author(s) declared no potential conflicts of interest with respect to the research, authorship, and/or publication of this article.

\section*{Funding}
The author(s) disclosed receipt of the following financial support
for the research, authorship, and/or publication of this article:
GRAVATAR project PID2023-151184OB-I00 funded by the Spanish Ministry of Science, Innovation, and Universities MCIU, the State Research Agency AEI/10.13039/501100011033 and by the European Regional Development Fund ERDF; PNRR-MUR Project PE000013-CUP-J53C22003010006. Future Artificial Intelligence Research (FAIR), funded by the European Union, NextGenerationEU.

\section*{ORCID iDs}
Daniel Ordoñez Apraez \orcidlink{0000-0002-9793-2482} \href{https://orcid.org/0000-0002-9793-2482}{https://orcid.org/0000-0002-9793-2482}\\
Giulio Turrisi \orcidlink{0000-0003-3007-3553} \href{https://orcid.org/0000-0003-3007-3553}{https://orcid.org/0000-0003-3007-3553}\\
Vladimir Kostic \orcidlink{0000-0001-8341-1400} \href{https://orcid.org/0000-0001-8341-1400}{https://orcid.org/0000-0001-8341-1400}\\
Francesc Moreno-Noguer \orcidlink{0000-0002-8640-684X} \href{https://orcid.org/0000-0002-8640-684X}{https://orcid.org/0000-0002-8640-684X}\\
Antonio Agudo \orcidlink{0000-0001-6845-4998} \href{https://orcid.org/0000-0001-6845-4998}{https://orcid.org/0000-0001-6845-4998}\\
Massimiliano Pontil \orcidlink{0000-0001-9415-098X} \href{https://orcid.org/0000-0001-9415-098X}{https://orcid.org/0000-0001-9415-098X}\\
Claudio Semini \orcidlink{0000-0002-3034-4686} \href{https://orcid.org/0000-0002-3034-4686}{https://orcid.org/0000-0002-3034-4686}\\
Carlos Mastalli \orcidlink{0000-0002-0725-4279} \href{https://orcid.org/0000-0002-0725-4279}{https://orcid.org/0000-0002-0725-4279}\\

\theendnotes

\bibliographystyle{SageH}
\bibliography{references.bib}

\begin{thebibliography}{59}
\providecommand{\natexlab}[1]{#1}
\providecommand{\url}[1]{\texttt{#1}}
\providecommand{\urlprefix}{URL }
\expandafter\ifx\csname urlstyle\endcsname\relax
  \providecommand{\doi}[1]{DOI:\discretionary{}{}{}#1}\else
  \providecommand{\doi}{DOI:\discretionary{}{}{}\begingroup \urlstyle{rm}\Url}\fi

\bibitem[{Alessandri et~al.(2003)Alessandri, Baglietto and Battistelli}]{alessandri2003ddpestimation}
Alessandri A, Baglietto M and Battistelli G (2003) Receding-horizon estimation for discrete-time linear systems.
\newblock \emph{IEEE Transactions on Automatic Control} 48.

\bibitem[{Amatucci et~al.(2024)Amatucci, Turrisi, Bratta, Barasuol and Semini}]{amatucci2024accelerating}
Amatucci L, Turrisi G, Bratta A, Barasuol V and Semini C (2024) Accelerating model predictive control for legged robots through distributed optimization.
\newblock \emph{2024 IEEE/RSJ International Conference on Intelligent Robots and Systems (IROS)} .

\bibitem[{Bietti et~al.(2024)Bietti, Venturi and Bruna}]{biettisample}
Bietti A, Venturi L and Bruna J (2024) On the sample complexity of learning under invariance and geometric stability.
\newblock In: \emph{Proceedings of the 35th International Conference on Neural Information Processing Systems}, NIPS '21. Red Hook, NY, USA: Curran Associates Inc.
\newblock ISBN 9781713845393.

\bibitem[{Brehmer et~al.(2024)Brehmer, Bose, de~Haan and Cohen}]{brehmer2024edgi2}
Brehmer J, Bose J, de~Haan P and Cohen T (2024) Edgi: equivariant diffusion for planning with embodied agents.
\newblock In: \emph{Proceedings of the 37th International Conference on Neural Information Processing Systems}, NIPS '23'. pp. 1--17.

\bibitem[{Brehmer et~al.(2023)Brehmer, de~Haan, Behrends and Cohen}]{brehmer2023geometric2}
Brehmer J, de~Haan P, Behrends S and Cohen T (2023) Geometric algebra transformer.
\newblock In: \emph{Advances in Neural Information Processing Systems}, volume~37. pp. 1--10.
\newblock \urlprefix\url{https://arxiv.org/abs/2305.18415}.

\bibitem[{Bronstein et~al.(2021)Bronstein, Bruna, Cohen and Veli{\v{c}}kovi{\'c}}]{bronstein2021geometric}
Bronstein MM, Bruna J, Cohen T and Veli{\v{c}}kovi{\'c} P (2021) Geometric deep learning: Grids, groups, graphs, geodesics, and gauges.
\newblock \emph{arXiv preprint arXiv:2104.13478} .

\bibitem[{Budhiraja et~al.(2019)Budhiraja, Carpentier and Mansard}]{budhiraja2019dynamics}
Budhiraja R, Carpentier J and Mansard N (2019) Dynamics consensus between centroidal and whole-body models for locomotion of legged robots.
\newblock In: \emph{2019 International Conference on Robotics and Automation (ICRA)}. IEEE, pp. 6727--6733.

\bibitem[{Camurri et~al.(2017)Camurri, Fallon, Bazeille, Radulescu, Barasuol, Caldwell and Semini}]{camurri2017probabilistic}
Camurri M, Fallon M, Bazeille S, Radulescu A, Barasuol V, Caldwell DG and Semini C (2017) Probabilistic contact estimation and impact detection for state estimation of quadruped robots.
\newblock \emph{IEEE Robotics and Automation Letters} 2(2): 1023--1030.

\bibitem[{Carpentier et~al.(2019)Carpentier, Saurel, Buondonno, Mirabel, Lamiraux, Stasse and Mansard}]{carpentier2019pinocchio}
Carpentier J, Saurel G, Buondonno G, Mirabel J, Lamiraux F, Stasse O and Mansard N (2019) The pinocchio c++ library: A fast and flexible implementation of rigid body dynamics algorithms and their analytical derivatives.
\newblock In: \emph{2019 IEEE/SICE International Symposium on System Integration (SII)}. IEEE, pp. 614--619.

\bibitem[{Cesa et~al.(2021)Cesa, Lang and Weiler}]{cesa2021program}
Cesa G, Lang L and Weiler M (2021) A program to build e (n)-equivariant steerable cnns.
\newblock In: \emph{International Conference on Learning Representations}.

\bibitem[{Chirikjian and Kyatkin(2000)}]{chirikjian2000engineering}
Chirikjian GS and Kyatkin AB (2000) \emph{Engineering applications of noncommutative harmonic analysis: with emphasis on rotation and motion groups}.
\newblock CRC press.

\bibitem[{Corbères et~al.(2024)Corbères, Mastalli, Merkt, Havoutis, Fallon, Mansard, Flayols, Vijayakumar and Tonneau}]{corberes2024perceptive}
Corbères T, Mastalli C, Merkt W, Havoutis I, Fallon M, Mansard N, Flayols T, Vijayakumar S and Tonneau S (2024) Perceptive locomotion through whole-body mpc and optimal region selection.

\bibitem[{Cornwell(1997)}]{cornwell1997group}
Cornwell JF (1997) \emph{Group theory in physics: An introduction}.
\newblock Academic press.

\bibitem[{de~Haan et~al.(2024)de~Haan, Cohen and Brehmer}]{de2023euclidean}
de~Haan P, Cohen T and Brehmer J (2024) Euclidean, projective, conformal: Choosing a geometric algebra for equivariant transformers.
\newblock In: Dasgupta S, Mandt S and Li Y (eds.) \emph{Proceedings of The 27th International Conference on Artificial Intelligence and Statistics}, \emph{Proceedings of Machine Learning Research}, volume 238. PMLR, pp. 3088--3096.
\newblock \urlprefix\url{https://proceedings.mlr.press/v238/haan24a.html}.

\bibitem[{Dresselhaus et~al.(2007)Dresselhaus, Dresselhaus and Jorio}]{dresselhaus2007group_theory_applications_to_physics_of_condensed_matter}
Dresselhaus MS, Dresselhaus G and Jorio A (2007) \emph{Group theory: application to the physics of condensed matter}.
\newblock Springer Science \& Business Media.

\bibitem[{Feldman and Levin(2009)}]{feldman2009equilibrium}
Feldman AG and Levin MF (2009) The equilibrium-point hypothesis--past, present and future.
\newblock \emph{Progress in motor control: A multidisciplinary perspective} : 699--726.

\bibitem[{Golubitsky et~al.(2012)Golubitsky, Stewart and Schaeffer}]{golubitsky2012singularities_groups_bifurcation}
Golubitsky M, Stewart I and Schaeffer DG (2012) \emph{Singularities and Groups in Bifurcation Theory: Volume II}, volume~69.
\newblock Springer Science \& Business Media.

\bibitem[{{Grimminger} et~al.(2020){Grimminger}, {Meduri}, {Khadiv}, {Viereck}, {Wüthrich}, {Naveau}, {Berenz}, {Heim}, {Widmaier}, {Flayols}, {Fiene}, {Badri-Spröwitz} and {Righetti}}]{grimminger2020solo_robot}
{Grimminger} F, {Meduri} A, {Khadiv} M, {Viereck} J, {Wüthrich} M, {Naveau} M, {Berenz} V, {Heim} S, {Widmaier} F, {Flayols} T, {Fiene} J, {Badri-Spröwitz} A and {Righetti} L (2020) An open torque-controlled modular robot architecture for legged locomotion research.
\newblock \emph{IEEE Robotics and Automation Letters} 5(2): 3650--3657.
\newblock \doi{10.1109/LRA.2020.2976639}.

\bibitem[{Higgins et~al.(2022)Higgins, Racani{\`e}re and Rezende}]{higgins2022symmetry_deepmind}
Higgins I, Racani{\`e}re S and Rezende D (2022) Symmetry-based representations for artificial and biological general intelligence.
\newblock \emph{Frontiers in Computational Neuroscience} : 28.

\bibitem[{Ijspeert(2008)}]{ijspeert2008central}
Ijspeert AJ (2008) Central pattern generators for locomotion control in animals and robots: a review.
\newblock \emph{Neural networks} 21(4): 642--653.

\bibitem[{Jumper et~al.(2021)Jumper, Evans, Pritzel, Green, Figurnov, Ronneberger, Tunyasuvunakool, Bates, {\v{Z}}{\'\i}dek, Potapenko et~al.}]{jumper2021highly}
Jumper J, Evans R, Pritzel A, Green T, Figurnov M, Ronneberger O, Tunyasuvunakool K, Bates R, {\v{Z}}{\'\i}dek A, Potapenko A et~al. (2021) Highly accurate protein structure prediction with alphafold.
\newblock \emph{Nature} 596(7873): 583--589.

\bibitem[{Katz et~al.(2019)Katz, Di~Carlo and Kim}]{katz2019mini}
Katz B, Di~Carlo J and Kim S (2019) Mini cheetah: A platform for pushing the limits of dynamic quadruped control.
\newblock In: \emph{2019 international conference on robotics and automation (ICRA)}. IEEE, pp. 6295--6301.

\bibitem[{Klein et~al.(2023)Klein, Kr{\"a}mer and No{\'e}}]{klein2023equivariant}
Klein L, Kr{\"a}mer A and No{\'e} F (2023) Equivariant flow matching.
\newblock \emph{arXiv preprint arXiv:2306.15030} .

\bibitem[{Knapp(1986)}]{Knapp1986}
Knapp AW (1986) \emph{Representation Theory of Semisimple Groups, An Overview Based on Examples (PMS-36)}.
\newblock Princeton: Princeton University Press.

\bibitem[{Kobilarov et~al.(2015)Kobilarov, Ta and Dellaert}]{kobilarov2015estimation}
Kobilarov M, Ta DN and Dellaert F (2015) Differential dynamic programming for optimal estimation.
\newblock In: \emph{IEEE International Conference on Robotics and Automation (ICRA)}.

\bibitem[{Lanczos(2020)}]{lanczos2020variational_principles_mechanics}
Lanczos C (2020) \emph{The variational principles of mechanics}.
\newblock University of Toronto press.

\bibitem[{Lee et~al.(2020)Lee, Hwangbo, Wellhausen, Koltun and Hutter}]{lee2020learning}
Lee J, Hwangbo J, Wellhausen L, Koltun V and Hutter M (2020) Learning quadrupedal locomotion over challenging terrain.
\newblock \emph{Science robotics} 5(47): eabc5986.

\bibitem[{Li and Todorov(2004)}]{li2004iLQR}
Li W and Todorov E (2004) Iterative linear quadratic regulator design for nonlinear biological movement systems.
\newblock In: \emph{First International Conference on Informatics in Control, Automation and Robotics}, volume~2. SciTePress.

\bibitem[{Lin et~al.(2021)Lin, Zhang, Yu and Ghaffari}]{lin2021deep_contact_estimation}
Lin TY, Zhang R, Yu J and Ghaffari M (2021) Legged robot state estimation using invariant kalman filtering and learned contact events.
\newblock In: \emph{5th Annual Conference on Robot Learning}.

\bibitem[{Mastalli et~al.(2020{\natexlab{a}})Mastalli, Budhiraja, Merkt, Saurel, Hammoud, Naveau, Carpentier, Righetti, Vijayakumar and Mansard}]{mastalli-crocoddyl}
Mastalli C, Budhiraja R, Merkt W, Saurel G, Hammoud B, Naveau M, Carpentier J, Righetti L, Vijayakumar S and Mansard N (2020{\natexlab{a}}) Crocoddyl: An efficient and versatile framework for multi-contact optimal control.
\newblock In: \emph{IEEE International Conference on Robotics and Automation (ICRA)}.

\bibitem[{Mastalli et~al.(2020{\natexlab{b}})Mastalli, Havoutis, Focchi, Caldwell and Semini}]{mastalli2020planning}
Mastalli C, Havoutis I, Focchi M, Caldwell DG and Semini C (2020{\natexlab{b}}) {Motion Planning for Quadrupedal Locomotion: Coupled Planning, Terrain Mapping, and Whole-Body Control}.
\newblock \emph{IEEE Transactions on Robotics} 36.

\bibitem[{Mayne(1966)}]{mayne1966DDP}
Mayne D (1966) A second-order gradient method for determining optimal trajectories of non-linear discrete-time systems.
\newblock \emph{International Journal of Control} 3.

\bibitem[{Mittal et~al.(2024)Mittal, Rudin, Klemm, Allshire and Hutter}]{mittal2024symmetry}
Mittal M, Rudin N, Klemm V, Allshire A and Hutter M (2024) Symmetry considerations for learning task symmetric robot policies.
\newblock \emph{arXiv preprint arXiv:2403.04359} .

\bibitem[{Mondal et~al.(2022)Mondal, Jain, Siddiqi and Ravanbakhsh}]{mondal2022eqr_rl}
Mondal AK, Jain V, Siddiqi K and Ravanbakhsh S (2022) Eqr: Equivariant representations for data-efficient reinforcement learning.
\newblock In: \emph{International Conference on Machine Learning}. PMLR, pp. 15908--15926.

\bibitem[{No{\'e} et~al.(2020)No{\'e}, Tkatchenko, M{\"u}ller and Clementi}]{noe2020machine}
No{\'e} F, Tkatchenko A, M{\"u}ller KR and Clementi C (2020) Machine learning for molecular simulation.
\newblock \emph{Annual review of physical chemistry} 71: 361--390.

\bibitem[{Ordo{\~n}ez-Apraez et~al.(2022)Ordo{\~n}ez-Apraez, Agudo, Moreno-Noguer and Martin}]{ordonez2022adaptable}
Ordo{\~n}ez-Apraez D, Agudo A, Moreno-Noguer F and Martin M (2022) An adaptable approach to learn realistic legged locomotion without examples.
\newblock In: \emph{2022 International Conference on Robotics and Automation (ICRA)}. IEEE, pp. 4671--4678.

\bibitem[{Ordo{\~n}ez-Apraez et~al.(2024)Ordo{\~n}ez-Apraez, Kostic, Turrisi, Novelli, Mastalli, Semini and Pontil}]{ordonez2024dynamics}
Ordo{\~n}ez-Apraez D, Kostic V, Turrisi G, Novelli P, Mastalli C, Semini C and Pontil M (2024) Dynamics harmonic analysis of robotic systems: Application in data-driven koopman modelling.
\newblock In: \emph{6th Annual Learning for Dynamics \& Control Conference}. PMLR, pp. 1318--1329.

\bibitem[{Ordo{\~n}ez-Apraez et~al.(2023)Ordo{\~n}ez-Apraez, Martin, Agudo and Moreno-Noguer}]{ordonez-discretesymm}
Ordo{\~n}ez-Apraez D, Martin M, Agudo A and Moreno-Noguer F (2023) {On discrete symmetries of robotics systems: A group-theoretic and data-driven analysis}.
\newblock In: \emph{Proceedings of the 19th Robotics: Science and Systems Conference (RSS)}. Daegu, Republic of Korea.
\newblock ISBN 978-0-9923747-9-2.
\newblock \urlprefix\url{https://www.roboticsproceedings.org/rss19/index.html}.

\bibitem[{Orin et~al.(2013)Orin, Goswami and Lee}]{orin2013centroidal_momentum_matrix_cmm}
Orin DE, Goswami A and Lee SH (2013) Centroidal dynamics of a humanoid robot.
\newblock \emph{Autonomous robots} 35(2): 161--176.

\bibitem[{Ostrowski and Burdick(1996)}]{ostrowski1996geometric_perspectives}
Ostrowski J and Burdick J (1996) Geometric perspectives on the mechanics and control of robotic locomotion.
\newblock In: \emph{Robotics Research}. Springer, pp. 536--547.

\bibitem[{Quigley(1973)}]{quigley1973pseudovectors}
Quigley RJ (1973) Pseudovectors and reflections.
\newblock \emph{American Journal of Physics} 41(3): 428--430.

\bibitem[{Rezaei-Shoshtari et~al.(2022)Rezaei-Shoshtari, Zhao, Panangaden, Meger and Precup}]{rezaei2022continuousMDPhomomorphisms}
Rezaei-Shoshtari S, Zhao R, Panangaden P, Meger D and Precup D (2022) Continuous mdp homomorphisms and homomorphic policy gradient.
\newblock \emph{Advances in Neural Information Processing Systems} 35: 20189--20204.

\bibitem[{Schaal et~al.(2003)Schaal, Peters, Nakanishi and Ijspeert}]{schaal2003control}
Schaal S, Peters J, Nakanishi J and Ijspeert A (2003) Control, planning, learning, and imitation with dynamic movement primitives.
\newblock In: \emph{Workshop on Bilateral Paradigms on Humans and Humanoids: IEEE International Conference on Intelligent Robots and Systems (IROS 2003)}. pp. 1--21.

\bibitem[{Scholz and Sch{\"o}ner(1999)}]{scholz1999uncontrolled}
Scholz JP and Sch{\"o}ner G (1999) The uncontrolled manifold concept: identifying control variables for a functional task.
\newblock \emph{Experimental brain research} 126: 289--306.

\bibitem[{Selig(2005)}]{selig2005geometric_fundamentals_robotics}
Selig JM (2005) \emph{Geometric fundamentals of robotics}, volume 128.
\newblock Springer.

\bibitem[{Solà et~al.(2021)Solà, Deray and Atchuthan}]{sola2021micro}
Solà J, Deray J and Atchuthan D (2021) {A micro Lie theory for state estimation in robotics}.

\bibitem[{Su et~al.(2024)Su, Huang, Ordo{\~n}ez-Apraez, Li, Li, Liao, Turrisi, Pontil, Semini, Wu et~al.}]{su2024leveraging}
Su Z, Huang X, Ordo{\~n}ez-Apraez D, Li Y, Li Z, Liao Q, Turrisi G, Pontil M, Semini C, Wu Y et~al. (2024) Leveraging symmetry in rl-based legged locomotion control.
\newblock In: \emph{IEEE/RSJ International Conference on Intelligent Robots and Systems (IROS)}.

\bibitem[{Sutton and Barto(2018)}]{sutton2018reinforcement}
Sutton RS and Barto AG (2018) \emph{Reinforcement learning: An introduction}.
\newblock MIT press.

\bibitem[{Tassa et~al.(2014)Tassa, Mansard and Todorov}]{tassa2014control}
Tassa Y, Mansard N and Todorov E (2014) Control-limited differential dynamic programming.
\newblock In: \emph{IEEE International Conference on Robotics and Automation (ICRA)}. IEEE, pp. 1168--1175.

\bibitem[{Traversaro et~al.(2016)Traversaro, Brossette, Escande and Nori}]{traversaro2016eigenvalue}
Traversaro S, Brossette S, Escande A and Nori F (2016) {Identification of fully physical consistent inertial parameters using optimization on manifolds}.
\newblock In: \emph{IEEE/RSJ International Conference on Intelligent Robots and Systems (IROS)}.

\bibitem[{Van~der Pol et~al.(2020)Van~der Pol, Worrall, van Hoof, Oliehoek and Welling}]{van2020mdp}
Van~der Pol E, Worrall D, van Hoof H, Oliehoek F and Welling M (2020) Mdp homomorphic networks: Group symmetries in reinforcement learning.
\newblock \emph{Advances in Neural Information Processing Systems} 33: 4199--4210.

\bibitem[{Wang et~al.(2023)Wang, Jia, Zhu, Walters and Platt}]{wang2022robot_learning_equivariant_models}
Wang D, Jia M, Zhu X, Walters R and Platt R (2023) On-robot learning with equivariant models.
\newblock In: Liu K, Kulic D and Ichnowski J (eds.) \emph{Proceedings of The 6th Conference on Robot Learning}, \emph{Proceedings of Machine Learning Research}, volume 205. PMLR, pp. 1345--1354.
\newblock \urlprefix\url{https://proceedings.mlr.press/v205/wang23c.html}.

\bibitem[{Wang et~al.(2022{\natexlab{a}})Wang, Walters and Platt}]{wang2022so2_equivariant_rl}
Wang D, Walters R and Platt R (2022{\natexlab{a}}) {$\mathrm{SO}(2)$}-equivariant reinforcement learning.
\newblock In: \emph{International Conference on Learning Representations}.
\newblock \urlprefix\url{https://openreview.net/forum?id=7F9cOhdvfk_}.

\bibitem[{Wang et~al.(2022{\natexlab{b}})Wang, Walters and Yu}]{wang2022approximately}
Wang R, Walters R and Yu R (2022{\natexlab{b}}) Approximately equivariant networks for imperfectly symmetric dynamics.
\newblock In: Chaudhuri K, Jegelka S, Song L, Szepesvari C, Niu G and Sabato S (eds.) \emph{Proceedings of the 39th International Conference on Machine Learning}, \emph{Proceedings of Machine Learning Research}, volume 162. PMLR, pp. 23078--23091.
\newblock \urlprefix\url{https://proceedings.mlr.press/v162/wang22aa.html}.

\bibitem[{Weiler et~al.(2023)Weiler, Forré, Verlinde and Welling}]{weiler2023EquivariantAndCoordinateIndependentCNNs}
Weiler M, Forré P, Verlinde E and Welling M (2023) \emph{Equivariant and Coordinate Independent Convolutional Networks}.
\newblock Open access book.
\newblock \urlprefix\url{https://maurice-weiler.gitlab.io/cnn_book/EquivariantAndCoordinateIndependentCNNs.pdf}.

\bibitem[{Weissenbacher et~al.(2022)Weissenbacher, Sinha, Garg and Yoshinobu}]{weissenbacher2022koopman_symmetries}
Weissenbacher M, Sinha S, Garg A and Yoshinobu K (2022) Koopman q-learning: Offline reinforcement learning via symmetries of dynamics.
\newblock In: \emph{International Conference on Machine Learning}. PMLR, pp. 23645--23667.

\bibitem[{Wheeler(2014)}]{wheeler2014covariance_eom}
Wheeler JT (2014) General coordinate covariance of the euler lagrange equations.
\newblock \urlprefix\url{http://www.physics.usu.edu/Wheeler/ClassicalMechanics/CMCoordinateinvarianceofEulerLagrange.pdf}.
\newblock Classical Mechanics class notes.

\bibitem[{Wieber(2006)}]{wieber2006holonomy}
Wieber PB (2006) Holonomy and nonholonomy in the dynamics of articulated motion.
\newblock In: \emph{Fast motions in biomechanics and robotics}. Springer, pp. 411--425.

\bibitem[{Zinkevich and Balch(2001)}]{zinkevich2001symmetry_mdp_implications}
Zinkevich M and Balch T (2001) Symmetry in markov decision processes and its implications for single agent and multi agent learning.
\newblock In: \emph{In Proceedings of the 18th International Conference on Machine Learning}. Citeseer.

\end{thebibliography}


\section*{Appendix}

\section*{Notation}
\begin{tabularx}{\textwidth}{p{0.6in} p{2.4in}}
	\multicolumn{2}{c}{\textbf{Numbers and Arrays}}                                                                                                                                \\
	$x$                                 & A scalar, or scalar function $x(\cdot) $                                                                                                 \\
	${\vx}$                             & A vector, or vector-valued function $\vx(\cdot)$                                                                                         \\
	$\vx_1 \oplus \vx_2$                & Direct sum (stacking) of vectors, such that $\vx_1 \oplus \vx_2 := \begin{bsmallmatrix} \vx_1 \\ \vx_2 \end{bsmallmatrix}$               \\
	$\mK$                               & A matrix                                                                                                                                 \\
	$\mA \oplus \mB$                    & Direct sum of matrices, such that $\mA \oplus \mB := \begin{bsmallmatrix} \mA & \mO \\ \mO & \mB \end{bsmallmatrix}$                     \\
	$\mI$                               & Identity matrix                                                                                                                          \\
	\multicolumn{2}{c}{}                                                                                                                                                           \\ 
	\multicolumn{2}{c}{\textbf{Sets, Vector Spaces, and Function spaces}}                                                                                                          \\
	$\vsX, \vsZ, \vsH, \vsF$            & A vector/Hilbert space                                                                                                                   \\
	$\R,\C$                             & The set of real and complex numbers                                                                                                      \\
	$\vsX \oplus \vsY$                  & Direct sum of vector spaces $\vsX$ and $\vsY$ such that if $\vx \in \vsX$ and $\vy \in \vsY$, then $\vx \oplus \vy \in \vsX \oplus \vsY$ \\
	\multicolumn{2}{c}{}                                                                                                                                                           \\ 
	\multicolumn{2}{c}{\textbf{Group and representation theory}}                                                                                                                   \\
	$\G$                                & A symmetry group                                                                                                                         \\
	$\g, \g_1, \g_a$                    & A symmetry group element                                                                                                                 \\
	$\g \Glact \vx$                     & The (left) group action of $\g$ on $\vx$ defined by $\g \Glact \vx := \rep[\vsX]{\g}\vx$, for a chosen basis $\sI_{\vsX}$                \\
	$\rep[\vsX]{} $                     & A representation of the group $\G$ on the vector space $\vsX$, defined for a chosen basis $\sI_{\vsX}$                                   \\
	$\rep[\vsX]{\g}$                    & Representation of the group element $\g$ on the vector space $\vsX$, defined for a chosen basis $\sI_{\vsX}$                             \\
	$\rep[\vsX]{} \oplus \rep[\vsY]{} $ & Direct sum of group representations,
	such that $\rep[\vsX]{\g} \oplus \rep[\vsY]{\g} := \begin{bsmallmatrix} \rep[\vsX]{\g} &  \\  & \rep[\vsY]{\g} \end{bsmallmatrix}$                                             \\
	$\G \vx$                            & The group orbit of $\vx$, defined as $\G\vx := \{ \g \Glact \vx \;|\; \g \in \G\}$                                                       \\
	$\G[a] \times \G[b]$                & Direct product of groups $\G[a]$ and $\G[b]$                                                                                             \\
	$\GLGroup[\vsX]$                    & General Linear group on the vector space $\vsX$                                                                                          \\
	$\CyclicGroup[n]$                   & Cyclic group of order $n$                                                                                                                \\
	$\KleinFourGroup$                   & Klein four-group                                                                                                                         \\
\end{tabularx}

\printglossaries

\end{document}